\documentclass[10pt,twocolumn,letterpaper]{article}

\usepackage[pagenumbers]{cvpr} 

\newcommand{\Paragraph}[1]{\noindent\textbf{#1}}
\usepackage{xspace}
\usepackage{amsmath, amsfonts} 
\usepackage{algorithmic}
\usepackage{graphicx}
\usepackage{array}
\usepackage{textcomp}
\usepackage{stfloats}
\usepackage{url}
\usepackage{verbatim}
\usepackage{graphicx}
\usepackage{booktabs}  
\usepackage{multirow}
\usepackage{makecell}
\usepackage{marvosym}
\usepackage{xcolor}
\usepackage{colortbl}
\usepackage{marvosym} 
\usepackage[utf8]{inputenc}

\usepackage{bibunits}
\defaultbibliographystyle{ieeenat_fullname}  
\defaultbibliography{main}                   

\usepackage{enumitem}
\usepackage[inkscapeformat=pdf]{svg}
\usepackage{pgfplots}
\usepackage{newtxtext} 
\pgfplotsset{compat=1.18}
\usepgfplotslibrary{polar}
\usepackage[pagebackref,breaklinks,colorlinks,allcolors=cvprblue]{hyperref}
\definecolor{SODgreen}{RGB}{200,230,210}
\definecolor{CoSODpurple}{RGB}{220,200,230}
\definecolor{SISorange}{RGB}{250,220,190}
\definecolor{OurBlue}{RGB}{200,220,240}

\definecolor{cpt_purple}{RGB}{242, 229, 242}
\definecolor{cpt_green}{RGB}{217, 243, 229}
\definecolor{cpt_yellow}{RGB}{251, 240, 222}

\definecolor{cvprblue}{rgb}{0.21,0.49,0.74}
\usepackage[pagebackref,breaklinks,colorlinks,allcolors=cvprblue]{hyperref}

\def\paperID{12754} 
\def\confName{CVPR}
\def\confYear{2026}

\title{Saliency-R1: Incentivizing Unified Saliency Reasoning Capability in MLLM with Confidence-Guided Reinforcement Learning}

\author{
    Long Li$^{1}$\enspace 
    Shuichen Ji$^{1}$\enspace 
    Ziyang Luo$^{1}$\enspace
    Zhihui Li$^{2}$\enspace
    \\
    Dingwen Zhang$^{1}$\enspace
    Junwei Han$^{1}$\enspace
    Nian Liu$^{1}$\footnotemark[2]
    \\
    $^1$ Northwestern Polytechnical University\enspace 
    $^2$ University of Science and Technology of China
}

\begin{document}



 
\twocolumn[{%
\renewcommand\twocolumn[1][]{#1}%
\maketitle
\vspace{-12mm}
\begin{center}
    \centering
    \includegraphics[width=\linewidth]{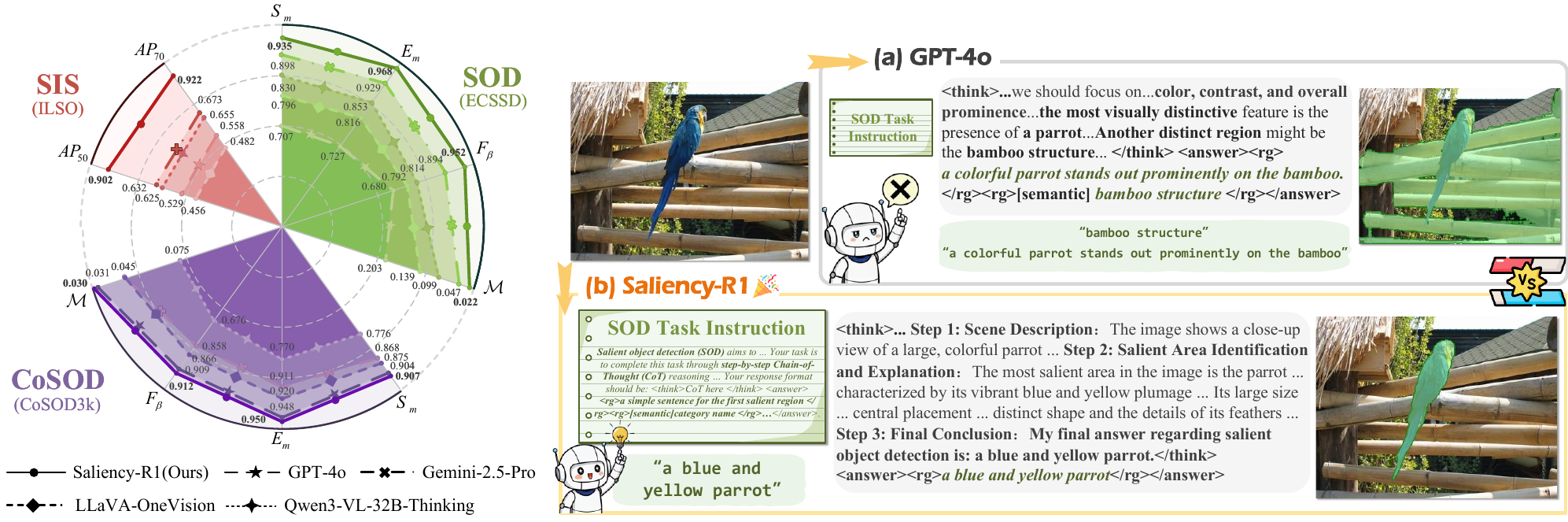}
    \vspace{-6mm} 
    \captionof{figure}{Existing multimodal large language models (MLLMs) exhibit limitations in saliency reasoning. This paper proposes Saliency-R1 to incentivize unified saliency reasoning of MLLM across three representative tasks, \ie, Salient Object Detection (SOD), Salient Instance Segmentation (SIS), and Co-salient Object Detection (CoSOD).}
    \label{fig:overview}
\end{center}%
\vspace{-1mm}
}]

\footnotetext[2]{Corresponding author: liunian228@gmail.com.}

\vspace{-2mm}
\begin{abstract}
Although multimodal large language models (MLLMs) excel in high-level vision-language reasoning, they lack inherent awareness of visual saliency, making it difficult to identify key visual elements. To bridge this gap, we propose Saliency-R1, the first unified MLLM framework that jointly tackles three representative and heterogeneous saliency tasks: Salient Object Detection (SOD), Salient Instance Segmentation (SIS), and Co-salient Object Detection (CoSOD), enhancing the model's capacity for saliency reasoning. We introduce a textual interface with structured tags (\texttt{<rg>}, \texttt{<ins>}) to encode region- and instance-level referring expressions, enabling a single referring segmenter to produce task-appropriate masks. To train the MLLM efficiently, we propose Confidence-Guided Policy Optimization (CGPO), a novel single-sample reinforcement learning algorithm. CGPO improves on GRPO by replacing group-normalized advantages with a per-sample signal based on reward–confidence discrepancy, thereby reducing computational waste, mitigating signal dilution, and lowering training overhead. Our model exceeds or matches the performance of robust open/closed-source MLLMs and specialized state-of-the-art methods across all three tasks, demonstrating the efficacy of our framework in saliency reasoning.
\end{abstract}    
\vspace{-4mm}
\section{Introduction}
\label{sec:intro}

Multimodal large language models (MLLMs) \cite{hurst2024gpt, comanici2025gemini, yang2025qwen3, xai2025grok, zhu2025internvl3, li2024llava, V2025glm4.1} integrate visual perception with linguistic reasoning, demonstrating remarkable capabilities in complex reasoning tasks. They excel in generalizing across various vision-language tasks, \eg, visual question answering \cite{fang2025guided, weng2025unsupervised}, image captioning \cite{sarto2025image, abdulgalil2025next}, and referring expression comprehension \cite{xuan2024pink, chen2025revisiting}, which \textit{rely on extensive world knowledge}. However, in the realm of saliency detection, tasks that \textit{rely on scene-specific visual context} to identify visually prominent elements, MLLMs often underperform \cite{dahou2025vision}, as illustrated in Figure~\ref{fig:overview}. This shortcoming indicates that current MLLMs lack inherent \textit{saliency awareness}: they can comprehend \textit{``what is visually described"} but struggle to identify \textit{``what is visually prominent"} in a scene.


To bridge this gap, we aim to enhance the reasoning capabilities of MLLM in the context of saliency detection.
This field encompasses various tasks, with the most representative being Salient Object Detection (SOD), Salient Instance Segmentation (SIS), and Co-salient Object Detection (CoSOD). SOD \cite{liu2018picanet, wei2020label, zhuge2022salient, liu2021part, liu2021visual} segments all visually prominent regions in a single image (\eg, ``a kitten” and ``a stuffed animal”); SIS \cite{fan2019s4net, wu2021regularized, pei2022transformer} distinguishes individual salient instances within a single image (\eg, ``a red cup on the left” and ``a blue cup on the right”); CoSOD \cite{zheng2023gconet+, li2025advanced, li2024conda, wang2025visual} identifies common salient objects that belong to the same semantic category across a group of images (\eg, ``rubber duck”);  
The collection of these three tasks represents the core paradigms of modern saliency detection, covering single-image and image-group inputs, intra-image and inter-image modeling, and output granularity that ranges from region-level segmentation (unions) to instance-level disambiguation.
Thus, to enable MLLMs to perform comprehensive saliency detection, we integrate these three representative tasks into a unified MLLM-based framework.

Since all three saliency tasks require segmentation outputs, we need to convert the MLLM’s reasoning output into segmentation masks. Recent MLLM-based segmentation methods \cite{lai2024lisa, wang2024segllm, wei2024hyperseg, chen2024sam4mllm, liu2025segmentation} address this by creating structured prompt representations that connect language reasoning with mask generation, typically inputting these into a promptable segmentation model (\eg, SAM \cite{kirillov2023segment}). These prompt representations can be categorized into two paradigms. The first uses learnable special tokens (\eg, \texttt{<SEG>} \cite{lai2024lisa, wang2024segllm}) to activate the SAM mask decoder. The second relies on geometric primitives, \eg, bounding boxes, points \cite{chen2024sam4mllm}, or attention-derived keypoints \cite{liu2025segmentation}, as prompts for the SAM-style model to generate masks. Unfortunately, these designs are originally devised for single object segmentation and fail to capture the distinct output granularities needed for the three tasks: SOD (multi-category region unions), SIS (disambiguated instances), and CoSOD (single-category regions across a group of images).

To overcome this limitation, we propose a unified textual referring interface that employs text-based region-level and instance-level referring expressions, combined with a referring segmentation model to generate task-specific segmentation masks.
This approach enables a single MLLM to effectively tackle all three tasks without requiring architectural modifications. Furthermore, our approach not only unifies these tasks but also maximizes the MLLM's text reasoning capabilities, facilitating easier training and application across diverse scenarios.

As for the training of saliency reasoning in the MLLM, we employ a two-stage training pipeline: Supervised Fine-Tuning (SFT) followed by Reinforcement Learning (RL) \cite{guo2025deepseek}. Although GRPO \cite{shao2024deepseekmath} is widely adopted in RL training, it suffers from three significant limitations: (1) it lacks confidence-aware learning, leading to wasted computation on uninformative responses; (2) reward and penalty signals are diluted due to group-mean baselining advantage calculations; and (3) it incurs excessive overhead from multi-sample generation and KL regularization. To address these issues, we propose Confidence-Guided Policy Optimization (CGPO), a novel RL algorithm that incorporates model reasoning confidence based on Bayesian calibration theory \cite{dawid1982well}.  CGPO leverages the reward and confidence discrepancy as a per-sample advantage calculation, which not only minimizes wasted computation on uninformative responses but also prevents signal dilution by bypassing group-mean calculation. Moreover, this group-independent advantage calculation permits streamlining the algorithm process to single-sampling, substantially reducing overhead. 

In summary, our main contributions are as follows:
\begin{itemize}
\item We are the first to incentivize the unified saliency reasoning capability for MLLM, using a unified textual interface to comprehensively tackle three representative and heterogeneous tasks \ie, SOD, SIS, and CoSOD.


\item We present CGPO, a lightweight single-sample RL algorithm that innovatively utilizes reward-confidence discrepancy as a per-sample advantage signal, reducing computational waste, mitigating signal dilution, and minimizing overhead.

\item Our model matches or exceeds the performance of closed- and open-source MLLMs and most specialized SOTA methods across all three tasks.

\end{itemize}

\section{Related Work}
\vspace{-1mm}
\subsection{Saliency Detection}
\vspace{-1mm}


\Paragraph{SOD.} SOD aims to segment visually salient regions in an image. Early CNN-based methods \cite{liu2018picanet} integrated global and local contexts, and label decoupling~\cite{wei2020label} refined boundaries. Later, part–whole modeling~\cite{zhuge2022salient,liu2021part} improved object completeness, while Transformer-based models \cite{liu2021visual} enabled unified global reasoning and detail refinement~\cite{yun2022selfreformer}.



\Paragraph{SIS.} SIS extends SOD to category-agnostic, instance-level segmentation~\cite{fan2019s4net,pei2020salient}. Early multi-stage methods~\cite{li2017instance} suffered from error accumulation; end-to-end models \cite{fan2019s4net,wu2021regularized} improved robustness. Recent Transformer-based approaches \cite{pei2022transformer} use query-driven, anchor-free detection, eliminating hand-crafted steps like NMS~\cite{wang2020solo}.



\Paragraph{CoSOD.}
CoSOD identifies common salient objects across multiple related images~\cite{tang2022rethinking,fan2021rethinking}, relying on cross-image consensus. Recent methods achieve this via strategies including cross-image attention~\cite{zhang2021CADC}, democratic prototyping~\cite{yu2022democracy}, group consistency~\cite{zheng2023gconet+}, region mining~\cite{li2025advanced}, and foundation model adaptation with consensus prompts~\cite{wang2025visual}.


Despite progress in task-specific models, saliency reasoning remains fragmented. We unify SOD, CoSOD, and SIS by incentivizing unified saliency reasoning in MLLM with confidence-guided reinforcement learning.

\vspace{-1mm}
\subsection{MLLM-based Segmentation}
\vspace{-1mm}


Recent works~\cite{lai2024lisa,wang2024segllm,wei2024hyperseg,chen2024sam4mllm,liu2025segmentation} adapt MLLMs for segmentation by linking language reasoning with mask generation via structured interfaces.  
LISA~\cite{lai2024lisa} and SegLLM~\cite{wang2024segllm} introduce a \texttt{<SEG>} token whose hidden state prompts a frozen mask decoder (\eg, SAM~\cite{kirillov2023segment}).  
HyperSeg~\cite{wei2024hyperseg} and SAM4MLLM~\cite{chen2024sam4mllm} guide SAM in segmentation by transforming semantic information into geometric prompts, while LENS~\cite{liu2025segmentation} enhances segmentation by utilizing keypoints extracted from attention maps.

However, these methods overlook the varied output granularities across SOD, SIS, and CoSOD. We introduce a structured textual interface that encodes task-aware granularity, enabling one segmenter to seamlessly handle all three tasks.

\vspace{-1mm}
\subsection{Advancements in GRPO}
\vspace{-1mm}


Recent advances have refined GRPO~\cite{shao2024deepseekmath} to enhance the stability and scalability of reinforcement learning. 
DAPO~\cite{yu2025dapo} decouples clipping bounds and applies dynamic sampling to enhance exploration and filter uninformative batches.
DrGRPO~\cite{liu2025understanding} provides an unbiased objective by removing response-length and group-variance normalization, improving token efficiency. 
GSPO~\cite{zheng2025gspo} instead defines sequence-level importance ratios for more stable training.

Unlike them, our CGPO uses model confidence for per-sample advantage estimation, enabling the model to optimize informative responses. This also reduces multi-sample generation, minimizes response interference, and significantly lowers computational burden.
\begin{figure*}[t]
\centering
\includegraphics[width=1.0\linewidth]{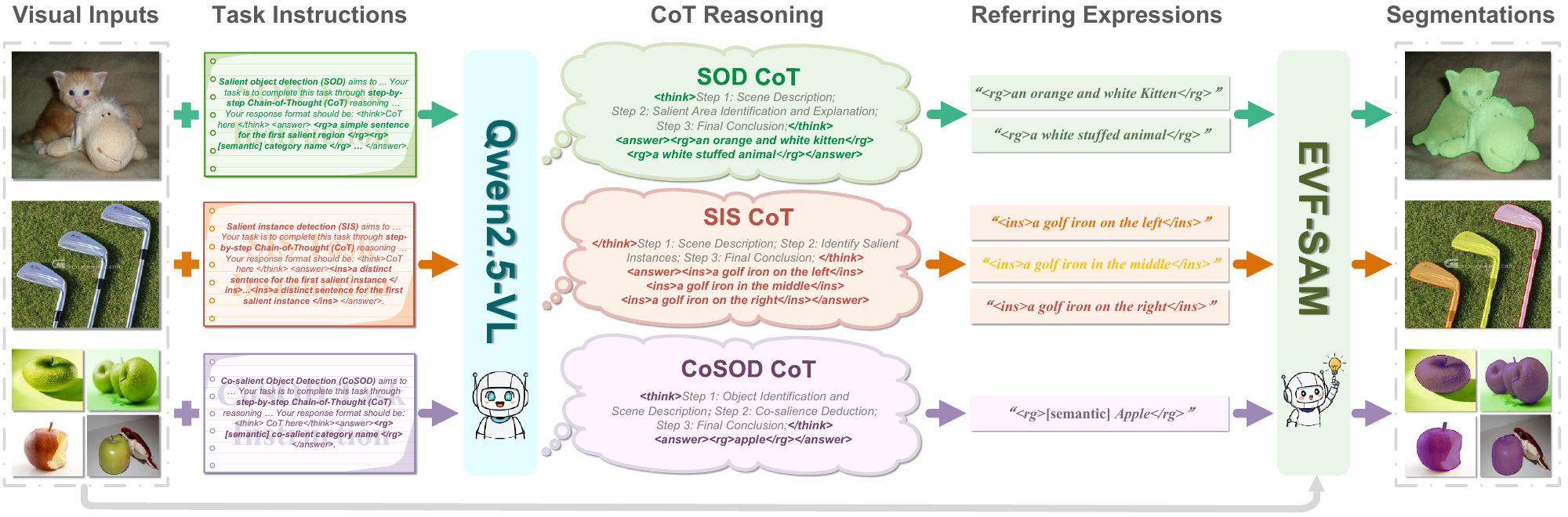}
\vspace{-4mm}
\captionof{figure}{\textbf{Overview of the Saliency-R1 framework.} Given visual inputs and task-specific instructions, the MLLM (QWen2.5-vl) generates structured CoT reasoning, where referring expressions can be parsed and processed by a referring segmentation model (EVF-SAM) to produce task-specific masks for SOD, SIS, and CoSOD.}
\label{fig:framework}
\vspace{-4mm}
\end{figure*}

\begin{figure}
\centering
\includegraphics[width=1.0\linewidth]{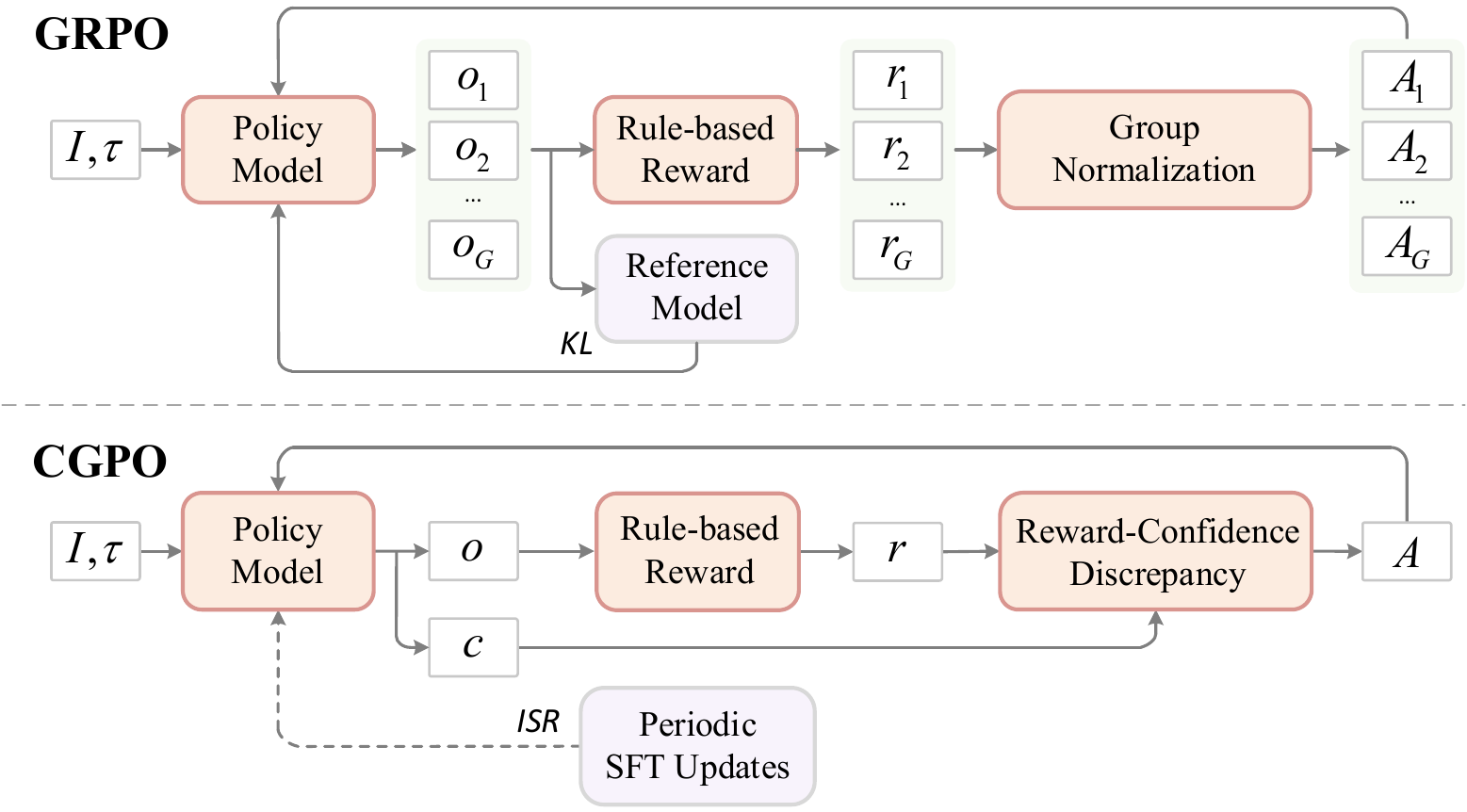}
\vspace{-5mm}
\caption{\textbf{Comparison between GRPO and our proposed CGPO.} GRPO uses multiple responses, group-normalized advantages, and KL regularization with a reference model. In contrast, CGPO employs a single response, calculates advantage with reward-confidence discrepancy, and replaces KL with ISR (Interleaved SFT Regularization).}
\vspace{-5mm}
\label{fig:comp_grpo_cgpo}
\end{figure}

\section{Proposed Method}
\subsection{Framework Construction}
\label{subsec:framework}

\Paragraph{Overview.}
As shown in Figure~\ref{fig:framework}, given a visual input $\boldsymbol{I}$ and a task instruction $\tau \in \{\tau^{SOD}, \tau^{SIS}, \tau^{CoSOD}\}$, the MLLM (Qwen2.5-VL~\cite{bai2025qwen2}) generates a structured Chain-of-Thought (CoT) response, from which region- or instance-level referring expressions are parsed to guide a referring segmenter $\mathcal{S}$ (EVF-SAM~\cite{zhang2024evf}) in producing segmentation masks with task-specific granularity across all three tasks.

\Paragraph{Input structure.}
The visual input $\boldsymbol{I}$ and instruction $\tau$ are task-specific:
\begin{itemize} 
\setlength{\itemsep}{0pt}  
    \setlength{\parskip}{0pt}  
    \setlength{\parsep}{0pt}  
    \item \textbf{SOD/SIS}: $\boldsymbol{I} = \{x^{\text{SOD}}\}/\{x^{SIS}\}$ (single image);  
    $\tau^{\text{SOD}}$ prompts step-by-step reasoning to identify all visually prominent regions, while $\tau^{\text{SIS}}$ requires enumerating all distinct salient instances with disambiguating attributes (\eg, location, color).
    \item \textbf{CoSOD}: $\boldsymbol{I} = \{x_k^{\text{CoSOD}}\}_{k=1}^K$ (a group of $K$ images);  $\tau^{\text{CoSOD}}$ guides the model to deduce the single shared semantic category across the group.  
\end{itemize} 

\Paragraph{Canonical CoT format.}
The MLLM output follows:
\begin{equation}
\texttt{<think>}\,\mathcal{T}\,\texttt{</think>}\,\texttt{<answer>}\,\mathcal{E}\,\texttt{</answer>},
\end{equation}
where $\mathcal{T}$ is a three-step reasoning process (scene/object identification $\rightarrow$ task analysis $\rightarrow$ conclusion), and $\mathcal{E}$ is the referring expression 
and can be parsed via string matching.

We design $\mathcal{E}$ using two structured textual tags to explicitly encode granularity heterogeneity: (1) \texttt{<rg>$\cdots$</rg>}: denotes a salient region tied to one semantic category. For SOD, where multiple salient regions of the same or different semantic categories may coexist (\eg, “an orange kitten” and “a white stuffed animal”), we use a combination of multiple \texttt{<rg>} blocks to refer to each region separately. For CoSOD, a single \texttt{<rg>} expresses the shared category across all images. (2) \texttt{<ins>$\cdots$</ins>}: denotes an instance-level descriptor enriched with distinguishing attributes, used exclusively in SIS to disambiguate co-occurring salient instances. These tags avoid the limitations of implicit interfaces (e.g., \texttt{<SEG>}) that bind all salient units into a single hidden-state embedding, making multi-target segmentation infeasible without architectural redesign.

Finally, $\mathcal{E}$ for three tasks are formulated as:
\begin{equation}
    \mathcal{E} :=
    \begin{cases}
        \big\{ \texttt{<rg>} d_i \texttt{</rg>} \big\}_{i=1}^{N} & \text{(SOD)}, \vspace{4pt} \\
        \big\{ \texttt{<ins>} \phi_j \texttt{</ins>} \big\}_{j=1}^{M} & \text{(SIS)}, \vspace{4pt} \\
        \texttt{<rg>[semantic] } s \texttt{</rg>} & \text{(CoSOD)}. 
    \end{cases}
\end{equation}
where $d_i$ is a single-category region description (\eg, ``an orange kitten'' or ``\texttt{[semantic] person}") for SOD, $\phi_j$ is an attribute-rich instance descriptor (\eg, ``a red cup on the left''), and $s$ is the single shared semantic category for CoSOD (\eg, ``\texttt{[semantic] rubber duck}"). $N$ is the number of salient regions in SOD, and $M$ is the number of salient instances in SIS.

\Paragraph{Referring segmentation.}
Given $\boldsymbol{I}$ and the parsed referring expressions $\mathcal{E} \in \{\mathcal{E}^{SOD}, \mathcal{E}^{SIS}, \mathcal{E}^{CoSOD}\}$, the referring segmenter $\mathcal{S}$ produces the final mask(s) $\boldsymbol{M}$ as:
\begin{equation}
    \boldsymbol{M} =
    \begin{cases}
        \bigvee_{i=1}^{N} \mathcal{S}\big(x^{\text{SOD}}, \mathcal{E}^{\text{SOD}}_i\big) & \text{(SOD)}, \vspace{4pt} \\ 
        \big\{ \mathcal{S}\big(x^{\text{SIS}}, \mathcal{E}^{\text{SIS}}_j\big) \big\}_{j=1}^{M} & \text{(SIS)}, \vspace{4pt} \\
        \big\{ \mathcal{S}\big(x_k^{\text{CoSOD}}, \mathcal{E}^{\text{CoSOD}}\big) \big\}_{k=1}^K & \text{(CoSOD)}.
    \end{cases}
\end{equation}
where $\bigvee$ denotes pixel-wise logical OR (\ie, union) of binary masks, merging all \texttt{<rg>} regions into a single salient mask for SOD.
This design enables a single segmentor $\mathcal{S}$ (EVF-SAM) to handle all three tasks without architectural modification.

\subsection{CoT Alignment}
\label{subsec:cot_alignment}
We train the MLLM to generate high-quality, task-aligned CoT reasoning in two stages. First, Supervised Fine-Tuning (SFT) initializes the model using a curated dataset of ground-truth CoT sequences.
Second, Reinforcement Learning (RL) refines the policy with a composite reward that jointly assesses reasoning format and segmentation accuracy (see supplementary material). This encourages structured, task-accurate CoT outputs.


\subsubsection{Supervised Fine-Tuning}
We curate the SFT data using an ``output-to-reasoning" strategy. Specifically, we prompt a powerful closed-source MLLM (\ie, Gemini \cite{comanici2025gemini}) with both the input image(s) and the corresponding ground-truth segmentation mask(s), and ask it to explain why the highlighted regions are salient. The resulting CoT explanations are treated as ground-truth reasoning traces for supervised training. This ensures logical consistency between the CoT and the ground-truth masks.

The MLLM is trained via the standard autoregressive cross-entropy loss to maximize the log-likelihood of the ground-truth CoT sequences. Formally, for visual input \(\boldsymbol{I}\), task instruction \(\tau\), and the corresponding ground-truth CoT response \(o^{\mathrm{gt}} = (o^{\mathrm{gt}}_1, \dots, o^{\mathrm{gt}}_{T^{\mathrm{gt}}})\), the SFT objective is:
\begin{equation}
\mathcal{J}_{\text{SFT}}(\theta) = \mathbb{E}_{(\boldsymbol{I}, \tau, o^{\mathrm{gt}})} \left[ \sum_{t=1}^{T^{\mathrm{gt}}} \log \pi_\theta \big( o^{\mathrm{gt}}_t \mid \boldsymbol{I}, \tau, o^{\mathrm{gt}}_{<t} \big) \right].
\label{eq:jsft}
\end{equation}
Here, \(\pi_\theta(\cdot \mid \boldsymbol{I}, \tau, o_{<t})\) denotes the MLLM’s autoregressive policy parameterized by \(\theta\), \ie, the probability distribution over the next token given the visual input, task instruction, and previously generated tokens.

\subsubsection{Reinforcement Learning}
\Paragraph{Preliminary.}  
To contextualize our CGPO, we briefly review GRPO~\cite{shao2024deepseekmath}, a group-based RL algorithm for optimizing autoregressive language policies. As illustrated in Figure~\ref{fig:comp_grpo_cgpo}, for each input pair $(\boldsymbol{I}, \tau)$, GRPO samples $G$ candidate responses $\{o_i\}_{i=1}^G$, computes a verifiable reward $r_i \in [0,1]$ for each, and constructs a group-normalized advantage, 
\begin{equation}
\vspace{-1mm}
A_i = \frac{r_i - \mu_r}{\sigma_r},
\label{eq:grpo_adv}
\vspace{-1mm}
\end{equation}
where $\mu_r$ and $\sigma_r$ are the empirical mean and standard deviation of the $G$ rewards. The policy update uses the importance sampling ratio:
\begin{equation}
\rho_{i,t}(\theta) = \frac{\pi_\theta(o_{i,t} \mid \boldsymbol{I}, \tau, o_{i,<t})}{\pi_{\theta_{\text{old}}}(o_{i,t} \mid \boldsymbol{I}, \tau, o_{i,<t})},
\label{eq:grpo_ratio}
\end{equation}
combined with a PPO-style clipped objective:
\begin{equation}
\vspace{-1mm}
\mathcal{C}_i = \frac{1}{T_i} \sum_{t=1}^{T_i} 
\min\left( 
    \rho_{i,t}(\theta) A_i,\; 
    \mathrm{clip}(\rho_{i,t}(\theta), 1-\varepsilon, 1+\varepsilon) A_i 
\right),
\label{eq:grpo_clip}
\vspace{-1mm}
\end{equation}
and a KL penalty against a fixed reference policy $\pi_{\text{ref}}$:
\begin{equation}
\vspace{-1mm}
\mathcal{J}_i^{\text{GRPO}} = \mathcal{C}_i - \beta D_{\mathrm{KL}}(\pi_\theta \| \pi_{\text{ref}}).
\label{eq:grpo_loss_i}
\vspace{-1mm}
\end{equation}

The overall objective is:
\begin{equation}
\vspace{-1mm}
\mathcal{J}_{\text{GRPO}}(\theta) = \mathbb{E}_{(\boldsymbol{I}, \tau), \{o_i\}} \left[ 
    \frac{1}{G} \sum_{i=1}^{G} \mathcal{J}_i^{\text{GRPO}}
\right].
\label{eq:grpo_obj}
\end{equation}

\Paragraph{Confidence-Guided Policy Optimization.} Despite its widespread use, GRPO faces three key limitations that impede learning efficiency and policy quality. 



\begin{figure}
\centering
\includegraphics[width=1.0\linewidth]{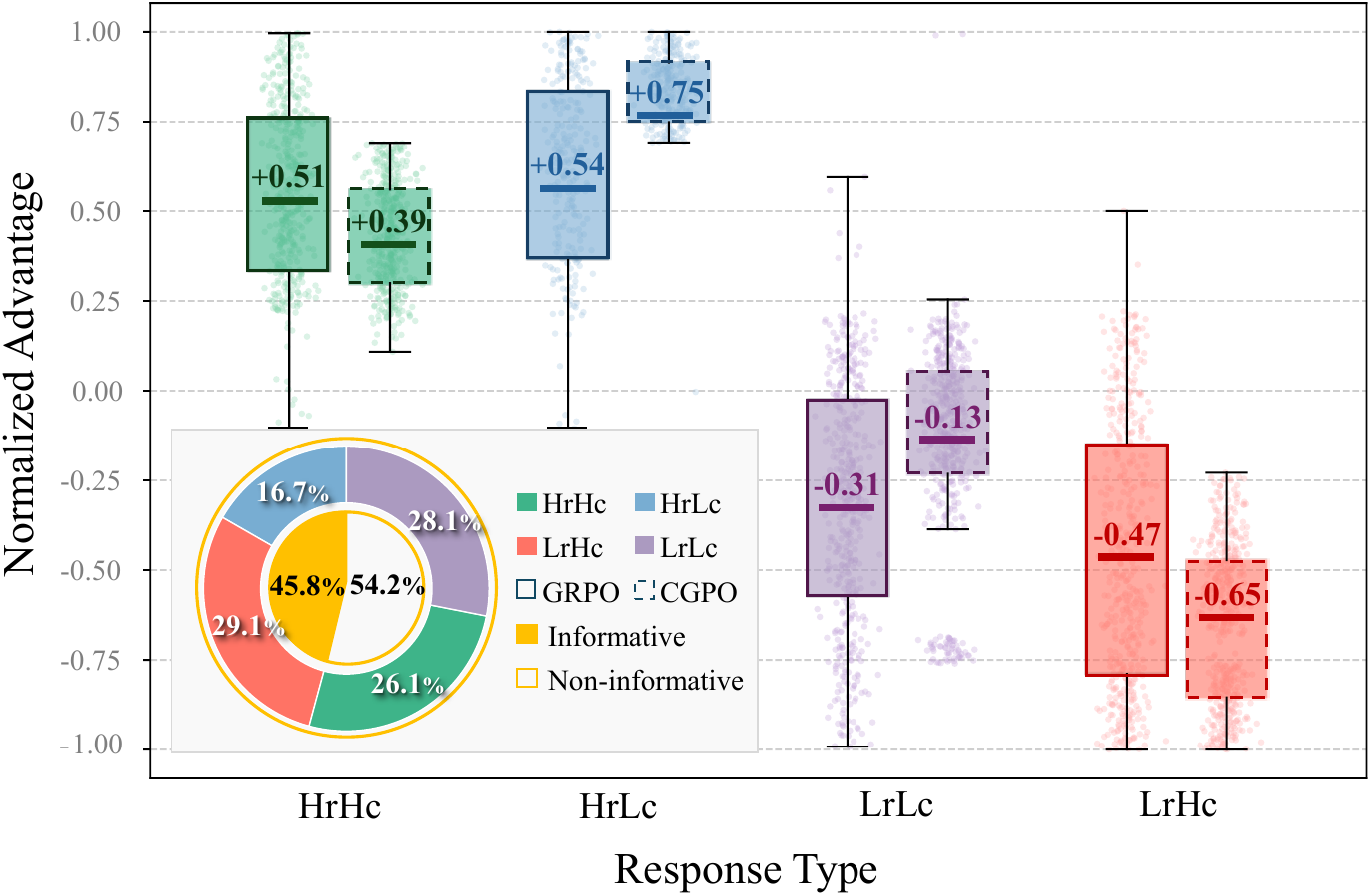}
\vspace{-4mm}
\caption{\textbf{Joint Response-Type Distribution and Per-Sample Advantage Analysis.} We analyze 8,000 responses generated by the SFT-initialized model, categorizing them into four types based on the bottom/top 20\% empirical thresholds of reward and model confidence. Confidence is defined as the mean token-wise generation probability of the CoT output (Eq.~\ref{confidence_calcualtion}). Advantages for GRPO and CGPO are rank-normalized \cite{chen2022rank} to $[-1,\,1]$, respectively, to ensure fair comparison.}
\label{fig:response_type_advantage_analysis}
\vspace{-4mm}
\end{figure}

First, GRPO is confidence-agnostic: it computes advantages solely from group-level reward statistics and ignores the model’s confidence. By discarding confidence, GRPO fails to prioritize informative responses, \ie, HrLc (high-reward \& low-confidence; correct but under-explored) and LrHc (low-reward \& high-confidence; overconfident errors). As shown in Figure~\ref{fig:response_type_advantage_analysis}, these two types constitute only 45.8\% of all responses, while the majority (54.2\%) belong to non-informative categories: HrHc (high-reward \& high-confidence; already well-learned) and LrLc (low-reward \& low-confidence; unreliable). Crucially, these non-informative responses nonetheless receive non-negligible positive or negative advantages, leading to unnecessary parameter updates and wasted computation on samples that contribute little to policy improvement. 

Second, the shared group-mean baseline causes signal dilution: advantages for informative responses are compressed. As shown in Figure~\ref{fig:response_type_advantage_analysis}, HrLc responses receive only modestly positive values, while LrHc responses incur insufficiently negative penalties. This occurs because dominant HrHc responses elevate the group mean, suppressing HrLc signals, and LrLc outliers depress it, softening penalties for LrHc, diluting learning signals and hindering policy improvement. 

Third, GRPO incurs prohibitive computational overhead: it requires $G$ forward passes, $G$ reward evaluations, and storage of $G$ KV caches per input due to group sampling, and further adds cost from the KL divergence penalty against a fixed reference model. 

To this end, we propose Confidence-Guided Policy Optimization (CGPO), as shown in Figure~\ref{fig:comp_grpo_cgpo}, a lightweight single-sample RL algorithm that explicitly accounts for the model’s confidence during policy updates to overcome GRPO’s key limitations. Specifically, we reinterpret advantage estimation through a Bayesian lens~\cite{kendall2017uncertainties}: the mean token-wise confidence 
\begin{equation}
\label{confidence_calcualtion}
\vspace{-1mm}
c = \frac{1}{T} \sum_{t=1}^{T} \pi_\theta(o_t \mid \boldsymbol{I}, \tau, o_{<t})
\vspace{-1mm}
\end{equation}
serves as a proxy for the prior belief that the generated CoT is correct, while the task-specific reward \(r \in [0,1]\) acts as observational evidence of correctness. The ideal learning signal should be designed to align prior belief with observational evidence, a principle known as calibration~\cite{dawid1982well, murphy1977reliability}. Since \(r\) is a soft estimation of correctness~\cite{szegedy2016rethinking}, the natural misalignment measure is the binary cross-entropy:
\begin{equation}
\vspace{-1mm}
\mathcal{L}_{\text{NLL}} = -r \log c - (1 - r) \log(1 - c).
\end{equation}


However, instead of minimizing $\mathcal{L}_{\text{NLL}}$ directly, we extract a policy update direction from its gradient. Since policy gradient methods maximize an objective, we follow the \textit{negative} gradient of $\mathcal{L}_{\text{NLL}}$ with respect to $c$:
\begin{equation}
\vspace{-1mm}
-\nabla_c \mathcal{L}_{\text{NLL}} = \frac{r - c}{c(1 - c)}.
\vspace{-1mm}
\end{equation}

The numerator $(r - c)$ captures the discrepancy between reward (evidence) and confidence (prior), \ie, Bayesian surprise~\cite{itti2009bayesian}. The denominator, however, causes numerical instability near $c \to 0$ or $c \to 1$. Discarding this unstable scaling factor, we adopt the signed error $(r - c)$ as our per-sample advantage:
\begin{equation}
\vspace{-1mm}
A = r - c.
\vspace{-1mm}
\end{equation}


As shown in Figure~\ref{fig:response_type_advantage_analysis}, our design yields advantages for non-informative responses (HrHc and LrLc) that are weaker than GRPO, thereby suppressing unnecessary updates. Meanwhile, by leveraging the per-sample reward–confidence discrepancy as the advantage signal, CGPO alleviates signal dilution and ensures that informative responses (HrLc and LrHc) receive stronger (more positive or more negative) signals. Furthermore, requiring only a single reasoning trace, CGPO removes group sampling and substantially reduces the training overhead. 

In addition to the single-sample advantage design, we simplify policy regularization by replacing GRPO’s static KL divergence penalty with Interleaved SFT Regularization (ISR). Specifically, we alternate between RL updates and SFT steps based on a fixed period \(P\).

The RL update uses the single-sample advantage and follows a PPO-style clipped objective:
\begin{equation}
\begin{split}
\mathcal{J}_{\text{RL}}(\theta) = \mathbb{E}_{(\boldsymbol{I}, \tau), o} \Bigg[ 
    &\frac{1}{T} \sum_{t=1}^{T} \min\Big( 
        \rho_{t}(\theta) \, A, \\
    & \mathrm{clip}\big(\rho_{t}(\theta), 1-\varepsilon, 1+\varepsilon\big) \, A 
    \Big) \Bigg].
\end{split}
\label{eq:cgpo_rl_obj}
\end{equation}

The SFT update uses the objective defined in Eq.~\eqref{eq:jsft}. We alternate between RL and SFT updates every \(P\) steps, yielding the final CGPO objective:
\begin{equation}
\vspace{-1mm}
\mathcal{J}_{\text{CGPO}}(\theta) =
\begin{cases}
\mathcal{J}_{\text{RL}}(\theta), & \text{if } s \bmod P \neq 0, \\
\mathcal{J}_{\text{SFT}}(\theta), & \text{if } s \bmod P = 0,
\end{cases}
\label{eq:cgpo_obj}
\vspace{-1mm}
\end{equation}
where \(s\) is the current training step.

\subsubsection{Reward Definition}
\label{sec:reward}

During RL training, the total reward $r$ combines a correctness term and a format term:
\begin{equation}
\vspace{-1mm}
    r = \lambda r_{\text{corr}} + (1-\lambda) r_{\text{fmt}}, \quad \lambda = 0.5.
\end{equation}
$r_{\text{corr}}$ measures segmentation quality, \ie, S-measure \cite{fan2017structure} for SOD/CoSOD and our proposed Instance-Aligned S-Measure (IASM) for SIS, which simultaneously considers both instance detection and segmentation performance. $r_{\text{fmt}}$ encourages well-structured, tag-consistent CoT outputs.  
The details of the IASM formulation and format scoring rules are provided in the supplementary material.

\section{Experiments}
\subsection{Evaluation Datasets and Metrics}

We evaluate our unified framework on standard benchmarks for all three tasks: SOD with DUT-O~\cite{yang2013saliency}, ECSSD~\cite{yan2013hierarchical}, and PASCAL-S~\cite{li2014secrets}; SIS using ILSO~\cite{li2017instance}, SIS10K~\cite{pei2022transformer}, and SOC~\cite{fan2018salient}; and CoSOD with CoSOD3k~\cite{fan2021rethinking}, CoCA~\cite{zhang2020gradient}, and CoSal2015~\cite{zhang2016detection}. For SOD and CoSOD, we select the following commonly used evaluation metrics: Structure-measure ($S_m$)~\cite{fan2017structure}, max F-measure ($F_{\text{max}}$)~\cite{5206596}, E-measure ($E_m$)~\cite{Fan2018Enhanced}, and Mean Absolute Error ($\mathcal{M}$)~\cite{6751300}. For SIS, we adopt the standard Average Precision (AP) metrics, specifically AP${50}$ and AP${70}$, which are commonly used for evaluating instance-level segmentation tasks.

\begin{table*}[t]
\centering
\footnotesize
\renewcommand{\arraystretch}{1.1}
\renewcommand{\tabcolsep}{3.8mm} 
\caption{\textbf{Ablation study on key components of our model and training strategy.} We conduct comprehensive ablation experiments on representative datasets from the CoSOD, SOD, and SIS tasks.  We use bold to mark the best results.}
\label{tab:ablation_studies_unified}
\vspace{-2mm}
\begin{tabular}{m{2.5cm}|cccc|cc|cccc}
\toprule[1.25pt]

\rowcolor[RGB]{241, 241, 241}
& \multicolumn{4}{c|}{\textbf{SOD Task} (ECSSD \cite{yan2013hierarchical})} & \multicolumn{2}{c|}{\textbf{SIS Task} (ILSO \cite{li2017instance})} & \multicolumn{4}{c}{\textbf{CoSOD Task} (CoSOD3k \cite{fan2021rethinking})} \\
\rowcolor[RGB]{241, 241, 241}
\multicolumn{1}{c|}{\multirow{-2}{*}{\textbf{Method}}} & $S_{m} \uparrow$ & $E_{\xi} \uparrow$ & $F_{\beta} \uparrow$ & $\mathcal{M} \downarrow$ & $\text{AP}_{70} \uparrow$ & $\text{AP}_{50} \uparrow$ & $S_{m} \uparrow$ & $E_{\xi} \uparrow$ & $F_{\beta} \uparrow$ & $\mathcal{M} \downarrow$  \\
\midrule

\rowcolor[RGB]{214,220,229}
\multicolumn{11}{c}{\em Ablation Study on RL Algorithm} \\
\midrule
$\circ$ SFT & 0.890 & 0.929 & 0.906 & 0.046 & 0.683 & 0.711 & 0.853 & 0.897 & 0.861 & 0.057 \\
$\circ$ $\text{SFT+GRPO}$\cite{guo2025deepseek} & 0.895 & 0.928 & 0.889 & 0.049 & 0.895 & \textbf{0.926} & 0.899 & 0.944 & 0.900 & 0.035 \\
$\circ$ $\text{SFT+DrGRPO}$~\cite{liu2025understanding} & 0.915 & 0.948 & 0.931 &	0.035 & 0.802 & 0.830 & 0.901 & 0.941 & 0.903 & 0.033 \\
$\circ$ $\text{SFT+DAPO}$~\cite{yu2025dapo} & 0.929 &	0.961 &	0.938 & 0.026 & 0.716 & 0.738 & 0.901 & 0.942 & 0.899 & 0.031 \\
$\circ$ $\text{SFT+GSPO}$~\cite{zheng2025gspo} & 0.894  & 0.923 & 0.899 & 0.044  & 0.755 & 0.780  & 0.897 & 0.937 & 0.894 & 0.319 \\
\rowcolor[RGB]{241, 241, 241}
$\bullet$ $\textbf{SFT+CGPO}$ & \textbf{0.935} & \textbf{0.968} & \textbf{0.952} & \textbf{0.022} & \textbf{0.902} & 0.922 & \textbf{0.907} & \textbf{0.950} & \textbf{0.912} & \textbf{0.030}  \\
\midrule

\rowcolor[RGB]{214,220,229}
\multicolumn{11}{c}{\em Ablation Study for CoT Reasoning} \\
\midrule
$\circ$ SFT & 0.890 & 0.929 & 0.906 & 0.046  & 0.683 & 0.711  & 0.853 & 0.897 & 0.861 & 0.057 \\
$\circ$ $\text{SFT}_{\text{w/o CoT}}$ & 0.870 & 0.908 & 0.883 & 0.062 & 0.525 & 0.542 & 0.887 & 0.933 & 0.903 & 0.039  \\
$\circ$ $\text{SFT}_{\text{w/o CoT}} + \text{CGPO}$ & 0.912 & 0.951 & 0.930 & 0.032 & 0.864 & 0.902  & \textbf{0.911} & \textbf{0.956} & \textbf{0.919} & \textbf{0.027} \\
\rowcolor[RGB]{241, 241, 241}
$\bullet$ $\textbf{SFT+CGPO}$ & \textbf{0.935} & \textbf{0.968} & \textbf{0.952} & \textbf{0.022} & \textbf{0.902} & \textbf{0.922}  & 0.907 & 0.950 & 0.912 & 0.030 \\
\midrule

\rowcolor[RGB]{214,220,229}
\multicolumn{11}{c}{\em Ablation Study for Multi-task Unified Training} \\
\midrule
$\circ$ Separate Training  & 0.925 & 0.958 & 0.938 & 0.027 & 0.898 & 0.921 & 0.903 &	0.947 & 0.907 &	0.031 \\
\rowcolor[RGB]{241, 241, 241}
$\bullet$ \textbf{Unified Training} & \textbf{0.935} & \textbf{0.968} & \textbf{0.952} & \textbf{0.022} & \textbf{0.902} & \textbf{0.922} & \textbf{0.907} & \textbf{0.950} & \textbf{0.912} & \textbf{0.030} \\
\bottomrule[1.25pt]
\end{tabular}
\vspace{-3mm}
\end{table*}

\begin{table}[t]
\centering
\footnotesize
\renewcommand{\arraystretch}{1.1}
\renewcommand{\tabcolsep}{1.0mm}
\caption{\textbf{Training overhead comparison on a single H20 GPU: CGPO vs. GRPO and its other variants (DrGRPO, DAPO, GSPO).} \textbf{Traces}: reasoning traces per sample; \textbf{Mem}: peak GPU memory;
\textbf{Step Time}: average training time per step; \textbf{Total Time}: wall-clock time for 3k RL steps.}
\label{tab:efficiency_comparison}
\vspace{-2mm}
\begin{tabular}{m{1.9cm}|c|c|c|c}
\toprule[1.25pt]
\rowcolor[RGB]{241, 241, 241}
\textbf{Method} & \textbf{Traces} & \textbf{Mem} (GB) & \textbf{Step Time} (s) & \textbf{Total Time} (h)\\
\midrule
$\circ$ GRPO\cite{guo2025deepseek} & 8 & 92.4 & 213.8 & 178.2 \\
$\circ$ GRPO w/o KL & 8 & 72.4 & 152.5 & 127.0 \\
$\circ$ DrGRPO~\cite{liu2025understanding} & 8 & 65.0 & 163.7 & 136.4 \\
$\circ$ DAPO~\cite{yu2025dapo} & 8 & 73.1 & 195.3 & 162.7 \\
$\circ$ GSPO~\cite{zheng2025gspo} & 8 & 71.0 & 162.8 & 135.7 \\
\rowcolor[RGB]{241, 241, 241}
$\bullet$ \textbf{CGPO} & \textbf{1} & \textbf{38.7}  & \textbf{50.6}  & \textbf{42.2} \\
\bottomrule[1.25pt]
\end{tabular}
\vspace{-4mm}
\end{table}

\subsection{Implementation Details}
\Paragraph{Training Data.} For SFT data construction, we use Gemini-2.5-Pro~\cite{comanici2025gemini} to generate CoT annotations conditioned on ground-truth masks, collecting:
2640 images from DUTS-TR~\cite{wang2017learning} for SOD, 2467 images from ILSO~\cite{li2017instance}, SIS10K~\cite{pei2022transformer}, and SOC~\cite{fan2018salient} for SIS, and 1262 image groups (8 images each) from CoCoSeg~\cite{wang2019robust} for CoSOD. For CGPO training data, we use 3000 examples per task from the same sources, comprising 1800 held-out examples for RL updates and 1200 SFT examples for ISR.

\Paragraph{Model Architecture.} We employ Qwen2.5-VL-7B-Instruct~\cite{bai2025qwen2} as our MLLM, freezing its vision encoder and fine-tuning only the language component via LoRA~\cite{hu2022lora}. For the referring segmentation model, we adopt EVF-SAM-multitask~\cite{zhang2024evf} and further adapt it for saliency detection. Specifically, we fine-tune the mask decoder using structured referring expressions from our SFT data as prompts, supervised by binary cross-entropy against ground-truth masks.

\Paragraph{Training Details.} The training process uses AdamW ($\beta_1=0.9$, $\beta_2=0.95$), with 3,000 SFT steps at a learning rate of $1\mathrm{e}{-6}$, followed by 3,000 CGPO RL steps at a learning rate of $1\mathrm{e}{-4}$. We implement ISR by alternating RL and SFT updates every 5 steps: for every 3 CGPO RL steps, we insert 2 SFT steps. All experiments are conducted on L20 GPUs using PyTorch~\cite{paszke2019pytorch}.

\subsection{Ablation Study}
\vspace{-1mm}

We validate our core contributions through ablation studies on representative datasets of three tasks: ECSSD (SOD), ILSO (SIS), and CoCA (CoSOD). Our analysis focuses on three aspects: (1) the effectiveness of our CGPO algorithm; (2) the necessity of explicit CoT reasoning; and (3) the benefits of multi-task unified training.

\subsubsection{Effectiveness of CGPO} 
\vspace{-1mm}
We assess the efficacy of our proposed CGPO by comparing it against standard GRPO and its recent representative variants, namely DrGRPO~\cite{liu2025understanding}, DAPO~\cite{yu2025dapo}, and GSPO~\cite{zheng2025gspo}, under identical experimental settings, which involve initializing with SFT followed by RL. As shown in Table~\ref{tab:ablation_studies_unified}, CGPO consistently outperforms GRPO and its variants across all three tasks, while avoiding performance degradation in metrics such as $E_{\xi}$ and $F_{\beta}$, which is observed in GRPO and GSPO compared to the SFT baseline in SOD (ECSSD). This improvement is attributed to two key factors: (1) using per-sample reward–confidence discrepancy as the advantage signal to prioritize informative responses; (2) eliminating group-mean baseline to prevent signal dilution. 

More importantly, as shown in Table~\ref{tab:efficiency_comparison}, CGPO achieves superior performance with significantly lower training overhead, as it eliminates multi-sample generation. It requires only 38.7 GB peak GPU memory and 42.2 hours for RL training, in contrast to GRPO and its variants, which consume $\geq$ 65.0 GB memory and $\geq$ 127.0 training hours.



\subsubsection{Effectiveness of CoT Reasoning} 
We evaluate the impact of explicit CoT reasoning using four variants (Table~\ref{tab:ablation_studies_unified}): 
(1) \text{SFT} with full CoT supervision; 
(2) $\text{SFT}_\text{w/o CoT}$ with direct instruction-to-answer mapping; 
(3) $\text{SFT}_\text{w/o CoT}+\text{CGPO}$ applying CGPO to the no-CoT model; 
and (4) \text{SFT+CGPO}, our full CoT-aware approach. 

Results show that explicit CoT reasoning consistently boosts SOD and SIS performance at both SFT and CGPO stages, as these tasks require context analysis and instance disambiguation, benefiting from stepwise, attribute-aware reasoning. 
CoSOD gains little from CoT since it only needs identifying a shared semantic category, making direct group-to-category mapping more stable for RL optimization. 
Thus, CoT proves crucial for fine-grained, compositional tasks, but less so for coarse, consensus-based ones.

\subsubsection{Effectiveness of Unified Training}
\vspace{-1mm}
\label{sec:multitask_ablation}

We compare separate training (three independently trained, task-specific models) against multi-task unified training (a single shared model trained jointly on all three tasks). As shown in Table~\ref{tab:ablation_studies_unified}, unified training achieves on-par or slightly better performance across SOD (ECSSD), SIS (ILSO), and CoSOD (CoSOD3k), despite using only one parameter set. This indicates strong task compatibility in the shared representation space, with no negative interference. Crucially, unified training matches or exceeds the performance of three separate models while offering clear gains in deployment simplicity, memory efficiency, and maintainability.





\begin{figure*}[t]
\begin{minipage}[c]{1\textwidth}
\centering
\footnotesize
\renewcommand{\arraystretch}{1.1}
\renewcommand{\tabcolsep}{3.2mm} 

\captionof{table}{\textbf{Comparison experiments with SOTA closed-source and opened-source MLLMs on the ECSSD, ILSO, and CoSOD3k datasets.} We use bold and underline to mark the best and second-best excellent results, respectively.}
\label{tab:sota_closed_source_comparison_formatted}
\vspace{-2mm}

\begin{tabular}{m{3.7cm}|cccc|cc|cccc}
\toprule[1.25pt]
\rowcolor[RGB]{241, 241, 241}
& \multicolumn{4}{c|}{\textbf{\textbf{SOD Task} (ECSSD  \cite{yan2013hierarchical})}} & \multicolumn{2}{c|}{\textbf{SIS Task} (\textbf{ILSO}  \cite{li2017instance})} & \multicolumn{4}{c}{\textbf{CoSOD Task} (\textbf{CoSOD3k}  \cite{fan2021rethinking})} \\
\rowcolor[RGB]{241, 241, 241}
\multicolumn{1}{c|}{\multirow{-2}{*}{\textbf{Method}}} & $S_{m} \uparrow$ & $E_{\xi} \uparrow$ & $F_{\beta} \uparrow$ & $\mathcal{M} \downarrow$ & $\text{AP}_{70} \uparrow$ & $\text{AP}_{50} \uparrow$ & $S_{m} \uparrow$ & $E_{\xi} \uparrow$ & $F_{\beta} \uparrow$ & $\mathcal{M} \downarrow$  \\
\midrule

\rowcolor[RGB]{214,220,229}
\multicolumn{11}{c}{\em Closed-source MLLMs} \\
\midrule
$\circ$ Gemini-2.5-Pro \cite{comanici2025gemini} & \underline{0.898} & \underline{0.929} & \underline{0.894} & \underline{0.047} & 0.632 & 0.673 & 0.868 & 0.911 & 0.858 & 0.045 \\
$\circ$ Claude-Sonet-4-5 \cite{anthropic2025sonnet45system} & 0.868 & 0.893 & 0.855 & 0.072 & \underline{0.644} & \underline{0.678} & \underline{0.904} & 0.947 & 0.907 & \textbf{0.030} \\
$\circ$ GPT-4o \cite{hurst2024gpt} & 0.707 & 0.727 & 0.680 & 0.203 & 0.529 & 0.558 & \underline{0.904} & \underline{0.948} & \underline{0.909} & \underline{0.031}\\
$\circ$ Qwen-VL-Max \cite{bai2023qwen} & 0.823 & 0.846 & 0.808 & 0.108 & 0.608 & 0.635 & 0.878 & 0.910 & 0.860 & 0.038 \\
$\circ$ Grok-4 \cite{xai2025grok} & 0.790 & 0.808 & 0.753 & 0.116 & 0.561 & 0.597 & 0.900 & 0.941 & 0.900 & 0.032 \\
\midrule

\rowcolor[RGB]{214,220,229}
\multicolumn{11}{c}{\em Opened-source MLLMs} \\
\midrule
$\circ$ Qwen2.5-VL-7B-Instruct \cite{bai2025qwen2} & 0.518 & 0.531 & 0.510 & 0.372 & 0.558 & 0.590  & 0.808 & 0.850 & 0.785 & 0.086  \\
$\circ$ Qwen3-VL-32B-Thinking \cite{yang2025qwen3}  & \underline{0.830} & \underline{0.853} & \underline{0.814} & \underline{0.099} & \underline{0.625} & \underline{0.655} & 0.776 & 0.770 & 0.676 & 0.075  \\
$\circ$ InternVL3-9B \cite{zhu2025internvl3}  & 0.576 & 0.580 & 0.559 & 0.325 & 0.501 & 0.545 & 0.866 & 0.909 & 0.856 & 0.048 \\
$\circ$ LLaVA-OneVision-7B \cite{li2024llava}  & 0.796 & 0.816 & 0.792 & 0.139 & 0.456 & 0.482 & \underline{0.875} & \underline{0.920} & \underline{0.866} & \underline{0.045} \\
$\circ$ GLM-4.1V-9B-Thinking \cite{V2025glm4.1} & 0.777 & 0.792 & 0.762 & 0.155 & 0.361 & 0.414 & 0.863 & 0.890 & 0.833 & \underline{0.045} \\
\midrule
\rowcolor[RGB]{241, 241, 241}
$\bullet$ \textbf{Saliency-R1(Ours)} & \textbf{0.935} & \textbf{0.968} & \textbf{0.952} & \textbf{0.022} & \textbf{0.902} & \textbf{0.922} & \textbf{0.907} & \textbf{0.950} & \textbf{0.912} & \textbf{0.030}  \\

\bottomrule[1.25pt]
\end{tabular}
\end{minipage}
\vspace{-1mm}
\end{figure*}

\begin{table*}[t]
\centering
\caption{\textbf{Quantitative comparisons with other SOTA methods in conventional SOD, SIS, and CoSOD tasks.} We conduct the comparison on nine
benchmark datasets. We use bold and underline to mark the best and second-best excellent results, respectively.}
\label{tab:unified_comparison_three-task}
\vspace{-2mm}
\resizebox{\textwidth}{!}{
\begin{tabular}{>{\arraybackslash}m{3.3cm}|c|c|cccc|cccc|cccc}
\toprule[1.25pt]

\rowcolor[RGB]{214,220,229}
\multicolumn{15}{c}{\normalsize \em Salient Object Detection (SOD)} \\
\midrule

& \multirow{2}{*}{Pub.} & \multirow{2}{*}{\shortstack{Training \\ Images}} & \multicolumn{4}{c|}{DUT-O \cite{yang2013saliency}} & \multicolumn{4}{c|}{ECSSD \cite{yan2013hierarchical}} & \multicolumn{4}{c}{PASCAL-S \cite{li2014secrets}} \\
\multicolumn{1}{c|}{\multirow{-2}{*}{Methods}} & & & $S_m \uparrow$ & $E_\xi \uparrow$ & $\text{F}_\beta \uparrow$ & $\mathcal{M} \downarrow$ & $S_m \uparrow$ & $E_\xi \uparrow$ & $\text{F}_\beta \uparrow$ & $\mathcal{M} \downarrow$ & $S_m \uparrow$ & $E_\xi \uparrow$ & $\text{F}_\beta \uparrow$ & $\mathcal{M} \downarrow$ \\
\midrule
$\circ$ VST \cite{liu2021visual} & ICCV2021 & 10553 &
0.850 & 0.888 & 0.800 & 0.058 &
0.932 & 0.964 & 0.944 & 0.033 &
0.873 & 0.920 & 0.871 & 0.060
\\
$\circ$ MENet \cite{wang2023pixels}  & CVPR2023 & 10553 &
0.850 & 0.879 & 0.792 & \underline{0.045} &
0.928 & 0.956 & 0.939 & 0.031 &
0.871 & 0.914 & 0.872 & 0.055
\\
$\circ$ SelfReformer \cite{yun2023towards}  & TMM2023 & 10553 &
0.861 & 0.894 & 0.807 & \textbf{0.043} &
\textbf{0.936} & \underline{0.965} & \underline{0.948} & \underline{0.027} &
\underline{0.881} & \underline{0.926} & \underline{0.883} & \underline{0.052}
\\
\rowcolor[RGB]{241, 241, 241}
$\bullet$ \textbf{$\text{Saliency-R1}(\text{Ours})$}  & - & \textbf{4440} &
\textbf{0.864} & \textbf{0.907} & \textbf{0.833} & 0.049 &
\underline{0.935} & \textbf{0.968} & \textbf{0.952} & \textbf{0.022} &
\textbf{0.887} & \textbf{0.935} & \textbf{0.891} & \textbf{0.042}
\\ \midrule

\rowcolor[RGB]{214,220,229}
\multicolumn{15}{c}{\normalsize \em Salient Instance Segmentation (SIS)} \\
\midrule
& \multirow{2}{*}{Pub.} & \multirow{2}{*}{\shortstack{Training \\ Images}} & \multicolumn{4}{c|}{ILSO \cite{li2017instance}} & \multicolumn{4}{c|}{SIS10K \cite{pei2022transformer}} & \multicolumn{4}{c}{SOC \cite{fan2018salient}} \\
\multicolumn{1}{c|}{\multirow{-2}{*}{Methods}} & & & \multicolumn{2}{c}{$\text{AP}_{70} \uparrow$} & \multicolumn{2}{c|}{$\text{AP}_{50} \uparrow$} & \multicolumn{2}{c}{$\text{AP}_{70} \uparrow$} & \multicolumn{2}{c|}{$\text{AP}_{50} \uparrow$} & \multicolumn{2}{c}{$\text{AP}_{70} \uparrow$} & \multicolumn{2}{c}{$\text{AP}_{50} \uparrow$} \\
\midrule
$\circ$ S4Net \cite{fan2019s4net}  & CVPR2019 & \textbf{500} &
\multicolumn{2}{c}{0.636} & \multicolumn{2}{c|}{0.867} &
\multicolumn{2}{c}{0.670} & \multicolumn{2}{c|}{0.833} &
\multicolumn{2}{c}{0.420} & \multicolumn{2}{c}{0.646}
\\
$\circ$ SCNet \cite{pei2020salient}  & Neurocomputing2020 & 14207 &
\multicolumn{2}{c}{0.674} & \multicolumn{2}{c|}{0.846} &
\multicolumn{2}{c}{0.692} & \multicolumn{2}{c|}{0.692} &
\multicolumn{2}{c}{0.414} & \multicolumn{2}{c}{0.609}
\\
$\circ$ RDPNet \cite{wu2021regularized}  & TIP2021 & 2900 &
\multicolumn{2}{c}{\underline{0.885}} & \multicolumn{2}{c|}{0.734} &
\multicolumn{2}{c}{0.694} & \multicolumn{2}{c|}{0.820} &
\multicolumn{2}{c}{0.482} & \multicolumn{2}{c}{0.606}
\\
$\circ$ OQTR \cite{pei2022transformer}  & TMM2022 & 9330 &
\multicolumn{2}{c}{0.799} & \multicolumn{2}{c|}{\underline{0.904}} &
\multicolumn{2}{c}{\underline{0.817}} & \multicolumn{2}{c|}{\textbf{0.881}} &
\multicolumn{2}{c}{\textbf{0.725}} & \multicolumn{2}{c}{\textbf{0.895}}
\\
\rowcolor[RGB]{241, 241, 241}
$\bullet$ \textbf{$\text{Saliency-R1}(\text{Ours})$} & - & 4267 &
\multicolumn{2}{c}{\textbf{0.902}} & \multicolumn{2}{c|}{\textbf{0.922}} &
\multicolumn{2}{c}{\textbf{0.843}} & \multicolumn{2}{c|}{\underline{0.870}} &
\multicolumn{2}{c}{\underline{0.714}} & \multicolumn{2}{c}{\underline{0.745}}
\\
\midrule

\rowcolor[RGB]{214,220,229}
\multicolumn{15}{c}{\normalsize \em Co-Salient Object Detection (CoSOD)} \\
\midrule
& \multirow{2}{*}{Pub.} & \multirow{2}{*}{\shortstack{Training \\ Images}} & \multicolumn{4}{c|}{CoCA \cite{zhang2020gicd}} & \multicolumn{4}{c|}{CoSal2015 \cite{zhang2016detection}} & \multicolumn{4}{c}{CoSOD3k \cite{fan2021rethinking}} \\
\multicolumn{1}{c|}{\multirow{-2}{*}{Methods}} & & & $S_m \uparrow$ & $E_\xi \uparrow$ & $\text{F}_\beta \uparrow$ & $\mathcal{M} \downarrow$ & $S_m \uparrow$ & $E_\xi \uparrow$ & $\text{F}_\beta \uparrow$ & $\mathcal{M} \downarrow$ & $S_m \uparrow$ & $E_\xi \uparrow$ & $\text{F}_\beta \uparrow$ & $\mathcal{M} \downarrow$ \\
\midrule
$\circ$ GCoNet+ \cite{zheng2023gconet+}  & TPAMI2023 & 208250 &
0.738 & 0.814 & 0.637 & 0.081 &
0.881 & 0.926 & 0.891 & 0.055 &
0.843 & 0.901 & 0.834 & 0.061
\\
$\circ$ CONDA \cite{li2024conda}  & ECCV2024 & 208250 &
0.763 & 0.839 & 0.685 & 0.089 &
0.900 & 0.938 & 0.908 & 0.045 &
0.862 & 0.911 & 0.853 & 0.056
\\
$\circ$ DMT+ \cite{li2025advanced}  & TPAMI2025 & 208250 &
0.779 & 0.834 & 0.689 & 0.076 &
0.899 & 0.935 & 0.903 & 0.047 &
0.866 & 0.912 & 0.854 & 0.054
\\
$\circ$ VCP \cite{wang2025visual}  & CVPR2025 & 208250 &
\underline{0.819} & \underline{0.871} & \underline{0.752} & \underline{0.054} &
\underline{0.927} & \underline{0.962} & \underline{0.941} & \underline{0.030} &
\underline{0.895} & \underline{0.938} & \underline{0.893} & \underline{0.043}
\\
\rowcolor[RGB]{241, 241, 241}
$\bullet$ \textbf{$\text{Saliency-R1}(\text{Ours})$} & - & \textbf{24496} &
\textbf{0.899} & \textbf{0.946} & \textbf{0.871} & \textbf{0.023} &
\textbf{0.928} & \textbf{0.965} & \textbf{0.942} & \textbf{0.024} &
\textbf{0.907} & \textbf{0.950} & \textbf{0.912} & \textbf{0.030}
\\

\bottomrule[1.25pt]
\end{tabular}
}
\vspace{-3mm}
\end{table*}

\vspace{-1mm}
\subsection{Comparison with MLLMs}
\vspace{-1mm}
We compare Saliency-R1 with several advanced MLLMs, including closed-source models such as GPT-4o \cite{hurst2024gpt}, Gemini-2.5-Pro \cite{comanici2025gemini}, Claude-Sonnet-4.5 \cite{anthropic2025sonnet45system}, Qwen-VL-Max \cite{bai2023qwen}, and Grok-4 \cite{xai2025grok}, as well as open-source models such as Qwen2.4-VL-7B-Instruct \cite{bai2025qwen2}, Qwen3-VL-32B-Thinking \cite{yang2025qwen3}, InternVL3-9B \cite{zhu2025internvl3}, LLaVA-OneVision-Qwen2-7B \cite{li2024llava}, and GLM-4.1V-9B-Thinking \cite{V2025glm4.1}. Although these MLLMs possess strong general reasoning capabilities, they show limited performance on visual saliency understanding, particularly in SOD and SIS tasks.

As shown in Table~\ref{tab:sota_closed_source_comparison_formatted}, our Saliency-R1 achieves consistent improvements across all tasks. Especially in SIS (ILSO), our model achieves an AP$_{50}$ of 0.922, surpassing GPT-4o (0.558) by over 36\% and Qwen3-VL-32B-Thinking (0.655) by 26.7\%, which clearly demonstrates its superior instance-level reasoning capability. 

These results indicate that Saliency-R1 effectively compensates for the deficiencies of current MLLMs in saliency reasoning. By integrating confidence-guided CoT reasoning with unified training, our model bridges the gap between general multimodal understanding and saliency reasoning.


\subsection{Comparison With State-of-the-Art Methods}
\vspace{-1mm}
To further evaluate the effectiveness of Saliency-R1, we compare it with representative state-of-the-art (SOTA) methods in each task domain. For SOD, we choose VST \cite{liu2021visual}, MENet \cite{wang2023pixels}, and SelFReformer \cite{yun2023towards}. For SIS, we benchmark against advanced category-agnostic instance segmentation methods including SCNet \cite{pei2020salient}, S4Net~\cite{fan2019s4net}, RDPNet~\cite{wu2021regularized}, and OQTR~\cite{pei2022transformer}. Within CoSOD, the comparison involves competitive methods like GCoNet+~\cite{zheng2023gconet+}, CONDA~\cite{li2024conda}, DMT+~\cite{li2023discriminative}, and VCP~\cite{wang2025visual}. 

As shown in Table~\ref{tab:unified_comparison_three-task}, our model consistently outperforms most compared methods, even though it is trained with far less data
than task-specific models. In SOD, it outperforms strong transformer- and CNN-based methods. For SIS, the model shows clear improvements, producing more coherent instance-level predictions. Its advantage is most pronounced in CoSOD, on the CoCA dataset, Saliency-R1 achieves an $S_m$ of 0.899, surpassing the previous best method VCP (0.819) by about 8\%, demonstrating outstanding capability in capturing cross-image consistency.
\vspace{-1mm}
\section{Conclusion}
\vspace{-1mm}
\label{sec:conclusion}

This paper proposes Saliency-R1 to incentivize unified saliency reasoning capability in MLLM through confidence-guided reinforcement learning. 
This design encourages the model to align its multimodal reasoning with spatial saliency cues, enabling coherent and interpretable visual reasoning. Experimental results on SOD, SIS, and CoSOD benchmarks demonstrate that Saliency-R1 effectively bridges multimodal reasoning and pixel-level visual saliency perception, outperforming both existing MLLMs and task-specific models. 

\clearpage
\setcounter{page}{1}
\maketitlesupplementary




This supplementary material provides essential implementation details and analysis omitted from the main paper due to space limits, including: (1) full task instruction templates; 
(2) complete formulation of the reward function, especially the Instance-Aligned S-measure (IASM) for SIS; 
(3) the ``output-to-reasoning'' prompts used to generate SFT data; 
(4) parameter-scale comparison with other MLLMs;
(5) qualitative comparisons with other MLLMs and specialize SOTA methods.

\section{Task Instruction Design}
\label{sec:task_instruction}

\begin{figure}[h!]
\centering
\includegraphics[width=1.0\linewidth]{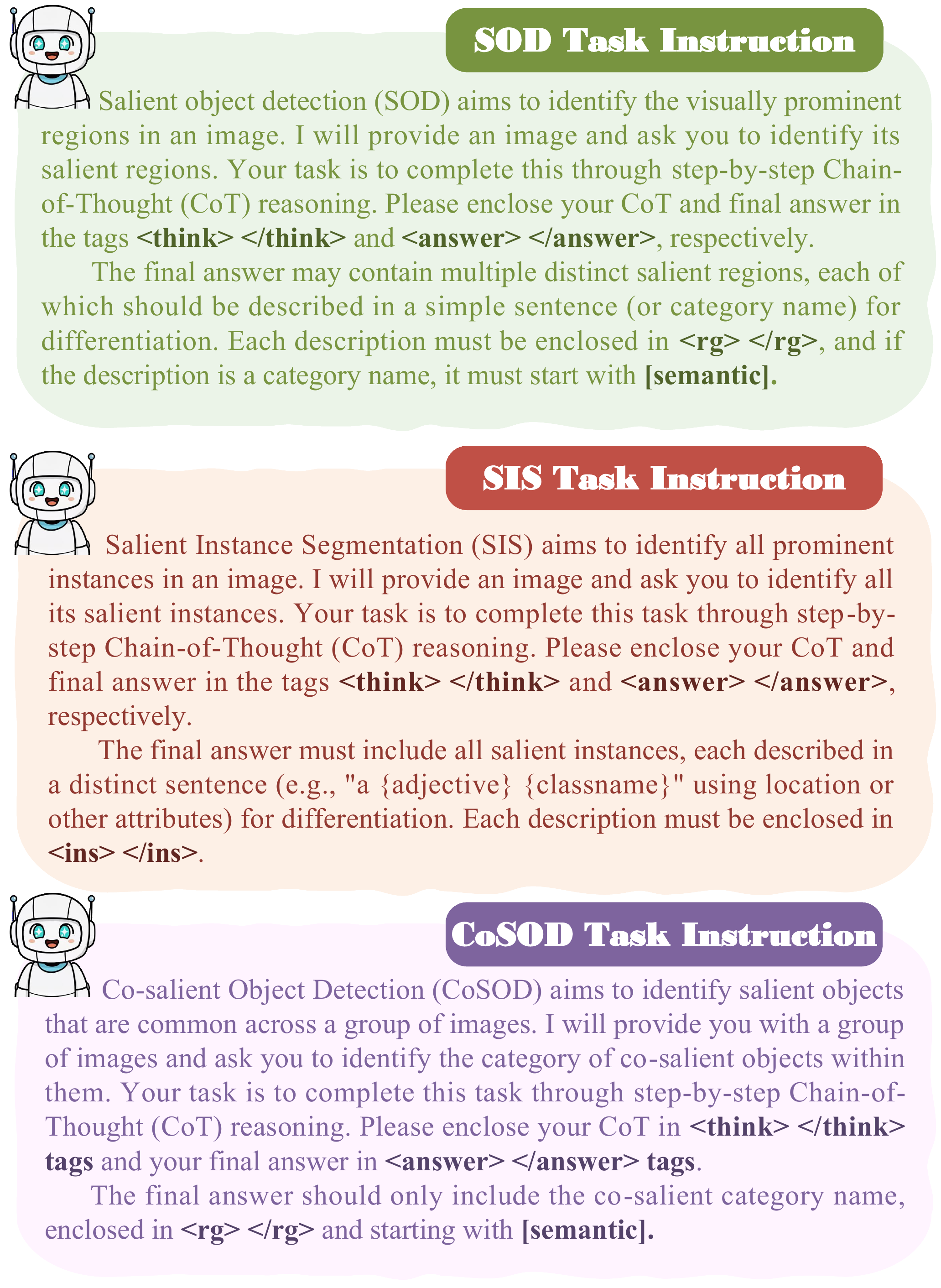}
\vspace{-6mm}
\caption{\textbf{Task instruction design for the three saliency tasks.}}
\label{fig:instruction_design}
\end{figure}

Figure~\ref{fig:instruction_design} illustrates the task instruction templates employed during the inference and confidence-guided RL training of Saliency-R1. Each instruction explicitly requires the MLLM to generate step-by-step CoT reasoning, adhering to strict formatting constraints: the reasoning trace must be enclosed in \texttt{<think>}~$\ldots$~\texttt{</think>}, and the final answer in \texttt{<answer>}~$\ldots$~\texttt{</answer>}. Crucially, the instructions are task-aware and prescribe distinct output semantics aligned with task granularity. 

For SOD, the final answer may contain multiple \texttt{<rg>} blocks, each describing a salient region (\eg, \texttt{<rg>a red cup</rg>}). When the description is a pure category name, it must be prefixed with \texttt{[semantic]} (\eg, \texttt{<rg>[semantic] dog</rg>}). 

For SIS, each salient instance must be expressed using an attribute-rich, disambiguating expression within an \texttt{<ins>} tag (\eg, \texttt{<ins>a red cup on the left</ins>}). 

For CoSOD, only a single \texttt{<rg>[semantic]\ $s$</rg>} expression is allowed, where $s$ denotes the shared semantic category across the image group (\eg, \texttt{<rg>[semantic] airplane</rg>}), ensuring cross-image consensus. 

\section{Reward Definition Details}
\label{sec:reward_details}

To align the MLLM’s textual reasoning with both task correctness and downstream parseability, our total reward is defined as a weighted sum:
\begin{equation}
    r = \lambda \cdot r_{\text{corr}} + (1 - \lambda) \cdot r_{\text{fmt}}, \quad \lambda = 0.5,
    \label{eq:total_reward}
\end{equation}
where $r_{\text{corr}}$ evaluates segmentation quality, and $r_{\text{fmt}}$ enforces strict adherence to the structured CoT interface. The weighting $\lambda = 0.5$ is selected based on empirical validation to balance optimization of reasoning fidelity and output safety. Below, we detail each component.

\subsection{Task-Adaptive Correctness Reward}
The definition of $r_{\text{corr}}$ is tailored to each saliency task to accommodate distinct input configurations and evaluation granularity.
\vspace{-4mm}
\paragraph{SOD.} Given parsed region expressions $\mathcal{E}^{\text{SOD}}_i$ ($i=1,\dots,N$), the predicted binary mask $\boldsymbol{M}\in \{0, 1\}^{H\times W}$ is obtained by pixel-wise logical OR:
\begin{equation}
    \boldsymbol{M} = \bigvee_{i=1}^N \mathcal{S}\big(x^{\text{SOD}}, \mathcal{E}^{\text{SOD}}_i\big),
    \label{eq:sod_mask}
\end{equation}
where $\mathcal{S}(\cdot)$ denotes the referring segmenter (EVF-SAM). Let $\boldsymbol{G} \in \{0,1\}^{H \times W}$ be the ground-truth mask. We adopt the Structure-measure $\mathcal{S}_m$~\cite{fan2017structure} as the correctness reward:
\begin{equation}
    r_{\text{corr}} = \mathcal{S}_m(\boldsymbol{M}, \boldsymbol{G}).
    \label{eq:sod_reward}
\end{equation}

\paragraph{CoSOD.} In CoSOD, masks are generated for each image $x^{\text{CoSOD}}_k$ with a shared expression $\mathcal{E}^{\text{CoSOD}}$:
\begin{equation}
    \boldsymbol{M}_k = \mathcal{S}\big(x^{\text{CoSOD}}_k, \mathcal{E}^{\text{CoSOD}}\big), \quad k=1,\dots,K.
\end{equation}
The ground-truth mask for the $k$-th image is $\boldsymbol{G}_k$, and the reward is the average S-measure across the group:
\begin{equation}
    r_{\text{corr}} = \frac{1}{K} \sum_{k=1}^K \mathcal{S}_m(\boldsymbol{M}_k, \boldsymbol{G}_k).
    \label{eq:cosod_reward}
\end{equation}

\paragraph{SIS.} Here, the standard S-measure is insufficient due to the need for instance correspondence. We introduce the Instance-Aligned S-measure (IASM):

Given $M$ predicted instance masks $\{\boldsymbol{M}_j\}_{j=1}^M$ and $I$ ground-truth masks $\{\boldsymbol{G}_i\}_{i=1}^I$, we construct the IoU matrix $\mathbf{U} \in \mathbb{R}^{I \times M}$ with entries

\begin{equation}
    U_{ij} = \frac{\| \boldsymbol{G}_i \cap \boldsymbol{M}_j \|}{\| \boldsymbol{G}_i \cup \boldsymbol{M}_j \|}.
\end{equation}

The Hungarian algorithm identifies the optimal one-to-one matching $\mathcal{M}$ to maximize total IoU, retaining pairs with $U_{ij} \geq \tau$ ($\tau = 0.1$). Unmatched instances are paired with zero masks, yielding aligned pairs $\{(\boldsymbol{G}'_\ell, \boldsymbol{M}'_\ell)\}_{\ell=1}^L$ (where $L = I + M - |\mathcal{M}|$). The correctness reward is defined as:

\begin{equation}
    r_{\text{corr}} = \frac{1}{L} \sum_{\ell=1}^L \mathcal{S}_m(\boldsymbol{M}'_\ell, \boldsymbol{G}'_\ell).
    \label{eq:iasm_reward}
\end{equation}

This formulation penalizes both over- and under-segmentation, along with instance misassignment, which is crucial for instance-level reasoning.

\subsection{Structured Format Reward}

Our pipeline requires precise parsing of \texttt{<think>} and \texttt{<answer>} blocks, alongside task-specific tags (\texttt{<rg>}, \texttt{<ins>}, \texttt{[semantic]}). Any malformed response disrupts the system, thus $r_{\text{fmt}}$ acts as a hard interface contract:

\begin{equation}
    r_{\text{fmt}} = r_{\text{struct}} + r_{\text{tag}},
\end{equation}
where $r_{\text{struct}} = 0.5$ if and only if the response contains exactly one well-formed \texttt{<think>} block and exactly one well-formed \texttt{<answer>} block (without nesting, truncation, or duplication); otherwise $r_{\text{struct}} = 0$. For $r_{\text{tag}}$, it equals $0.5$ if and only if the content inside \texttt{<answer>} adheres strictly to Eq.~(2) of the main paper:

\begin{itemize} 
\item \textbf{SOD}: $\mathcal{E} = \{\texttt{<rg>} d_i \texttt{</rg>}\}_{i=1}^N$, with $N \geq 1$ (at least one \texttt{<rg>} tag, each properly closed);
\item \textbf{CoSOD}: $\mathcal{E} = \texttt{<rg>[semantic] } s \texttt{ </rg>}$ (exactly one \texttt{<rg>} tag, starting with \texttt{[semantic]});
\item \textbf{SIS}: $\mathcal{E} = \{\texttt{<ins>} \phi_j \texttt{</ins>}\}_{j=1}^M$, with $M \geq 1$ (at least one \texttt{<ins>} tag, each properly closed).
\end{itemize}

Any deviation, such as missing tags, extraneous or missing closing brackets, incorrect tag types (\eg, \texttt{<ins>} in SOD), or malformed \texttt{[semantic]} annotations, results in $r_{\text{tag}} = 0$.

This design ensures that only fully compliant outputs receive $r_{\text{fmt}} = 1.0$, thus providing a clear and strong reinforcement learning signal for interface safety.

\section{Visualization of Model Predictions}
\label{sec:Visualization of Model Predictions}

\begin{figure*}
\centering
\includegraphics[width=1.0\linewidth]{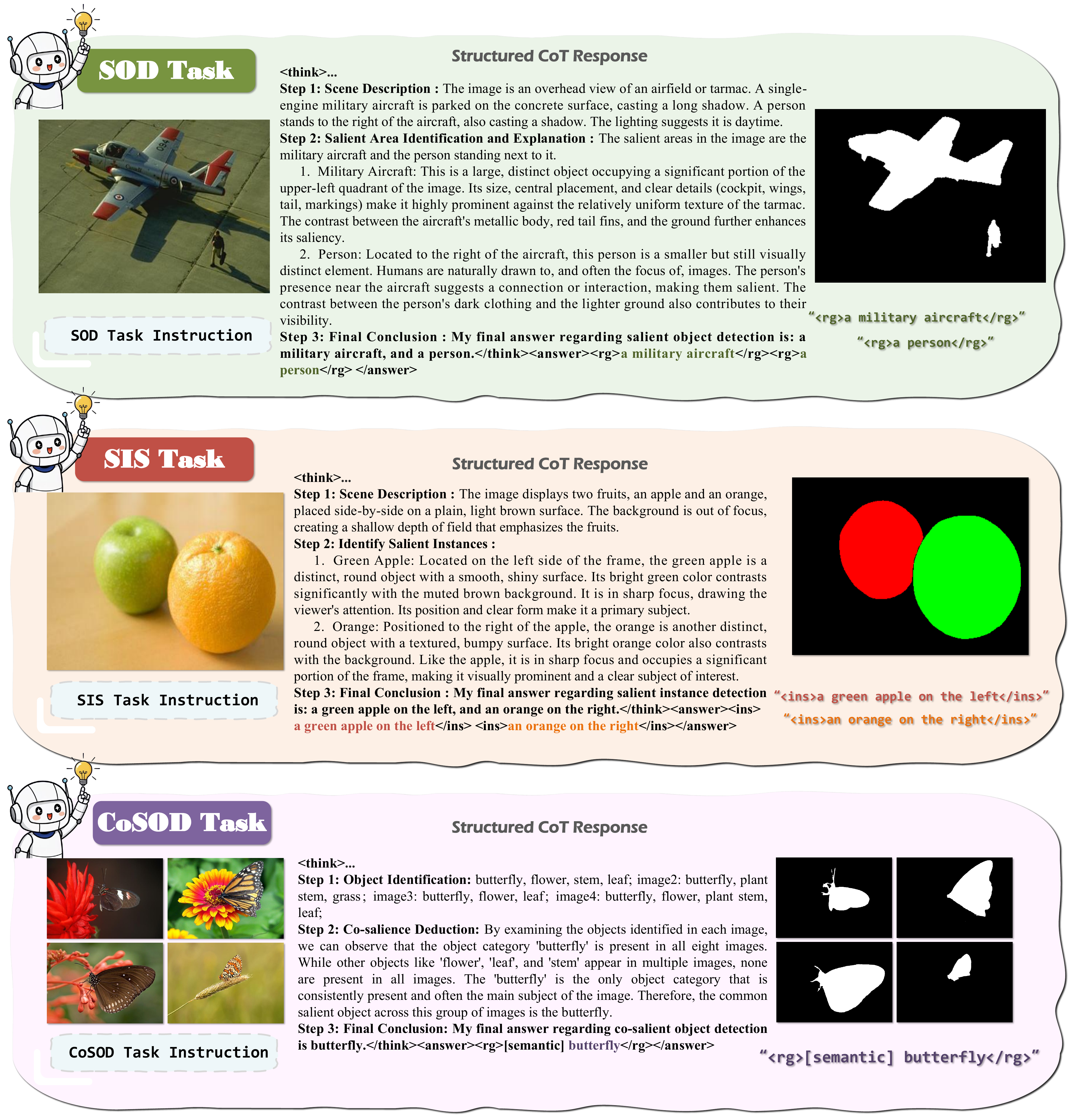}
\vspace{-4mm}
\caption{\textbf{Qualitative examples of Saliency-R1’s reasoning and segmentation on SOD, SIS, and CoSOD.} Each shows multimodal inputs, structured CoT responses, parsed referring expressions, and final segmentation masks.}
\vspace{-4mm}
\label{fig:short-prediction-answer}
\end{figure*}


Figure~\ref{fig:short-prediction-answer} illustrates representative outputs of Saliency-R1 across the three saliency tasks, \ie, SOD, SIS, and CoSOD. This demonstrates how structured CoT reasoning facilitates precise segmentation.

In the SOD task, the model produces two region-level referring expressions, \ie, ``a military aircraft'' and ``a person'', which are enclosed in \texttt{<rg>} tags. These expressions are processed independently by EVF-SAM, and their masks are unioned to form the final salient region map.

In the SIS task, the model differentiates two overlapping instances using attribute-rich descriptors, \ie, ``a green apple on the left'' and ``an orange on the right'', within \texttt{<ins>} tags. EVF-SAM generates separate binary masks for each, allowing for accurate instance-level disambiguation without merging.

In the CoSOD task, the model analyzes four images, identifies ``butterfly'' as the only object category consistently present, and outputs a single semantic-tagged expression: \texttt{<rg>[semantic] butterfly</rg>}. This unified cue is applied across all images, yielding consistent co-salient masks.

Critically, all responses strictly adhere to the structured format (Eq.~(1)–(2)), ensuring parseability. The resulting masks exhibit high fidelity: complete object coverage, sharp boundaries, and robustness to occlusion or clutter, thereby validating the effectiveness of our Saliency-R1 interface.

\section{Prompts for SFT Data Generation via Output-to-Reasoning}
\label{sec:sft_prompts}

We adopt the ``output-to-reasoning" strategy, as introduced in Sec.~3.2.1, to generate high-quality CoT supervision for SFT. Instead of prompting the model to generate reasoning solely from instructions, we condition a powerful closed-source MLLM (Gemini-2.5-Pro~\cite{comanici2025gemini}) on both the raw visual input \textit{and} its corresponding ground-truth segmentation mask(s) as a visual cue. These ground-truth masks serve as precise saliency priors, allowing the model to anchor its stepwise reasoning in actual pixel-level evidence, thereby preserving naturalness and linguistic fluency. 

\begin{figure*}
\centering
\includegraphics[width=1.0\linewidth]{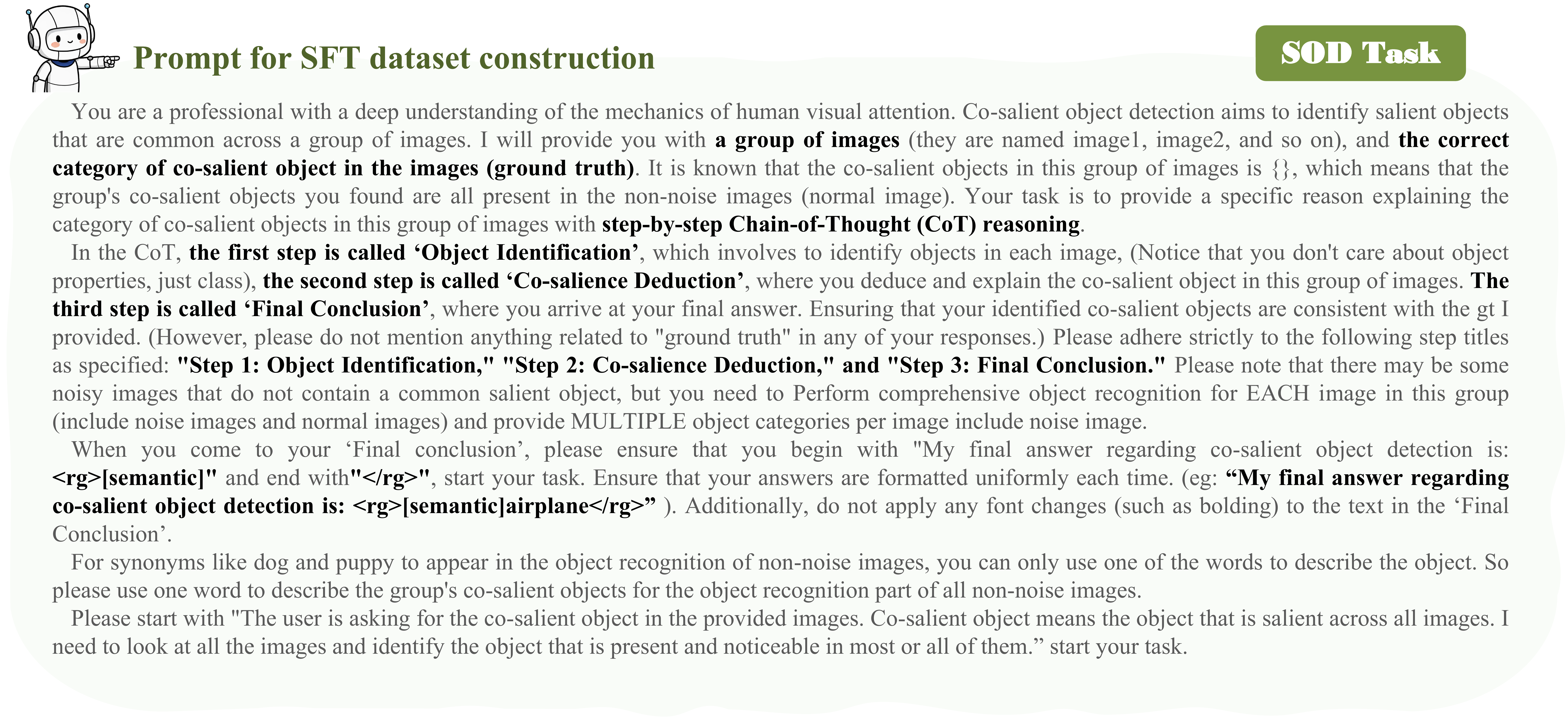}
\vspace{-8mm}
\caption{\textbf{Prompt design for SFT dataset on the salient object detection (SOD) tasks.}}
\label{fig:sft_data_prompt_sod}
\end{figure*}

\begin{figure*}
\centering
\includegraphics[width=1.0\linewidth]{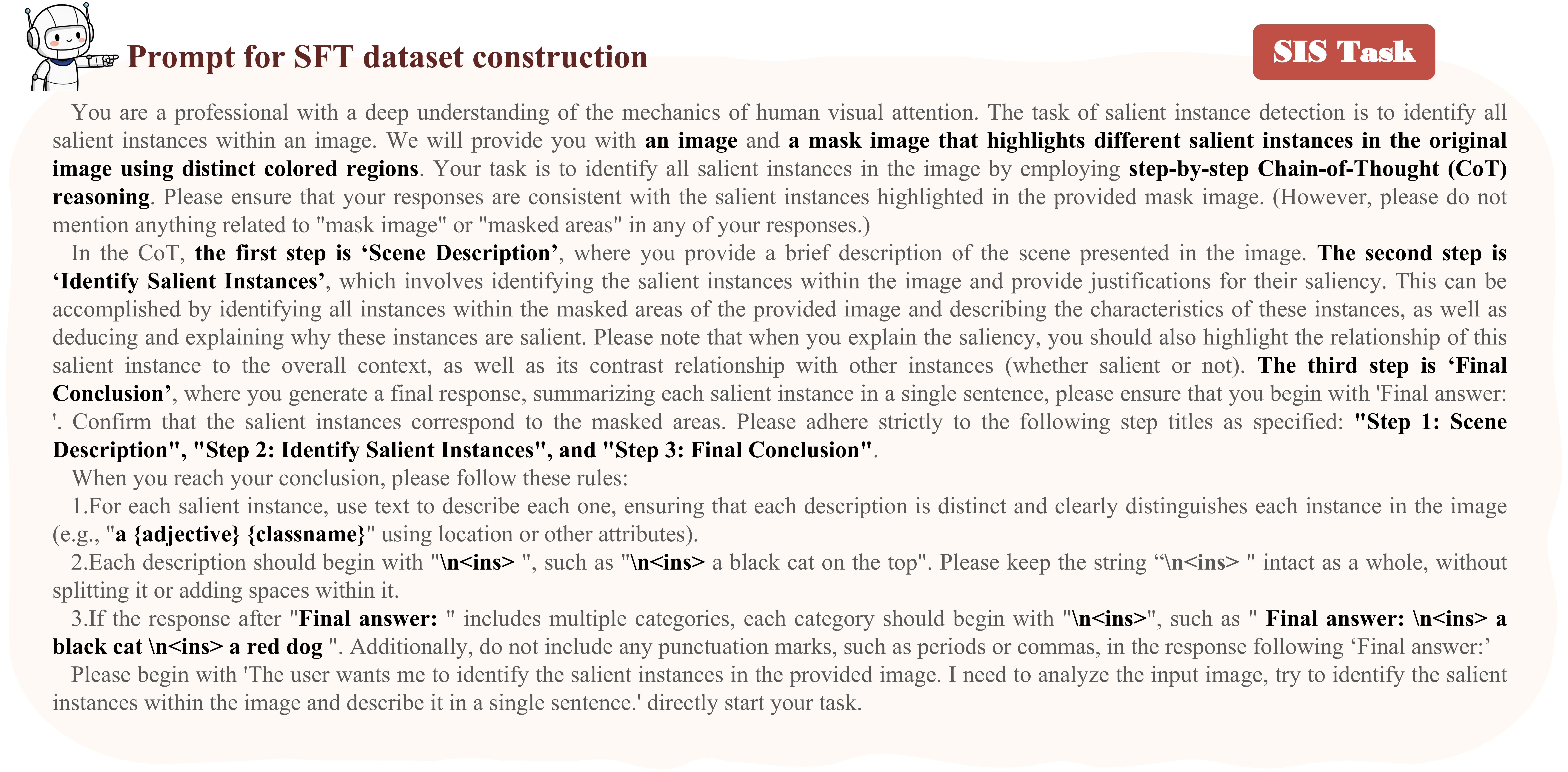}
\vspace{-8mm}
\caption{\textbf{Prompt design for SFT dataset on the salient instance segment (SIS) tasks.}}
\label{fig:sft_data_prompt_sis}
\end{figure*}

\begin{figure*}
\centering
\includegraphics[width=1.0\linewidth]{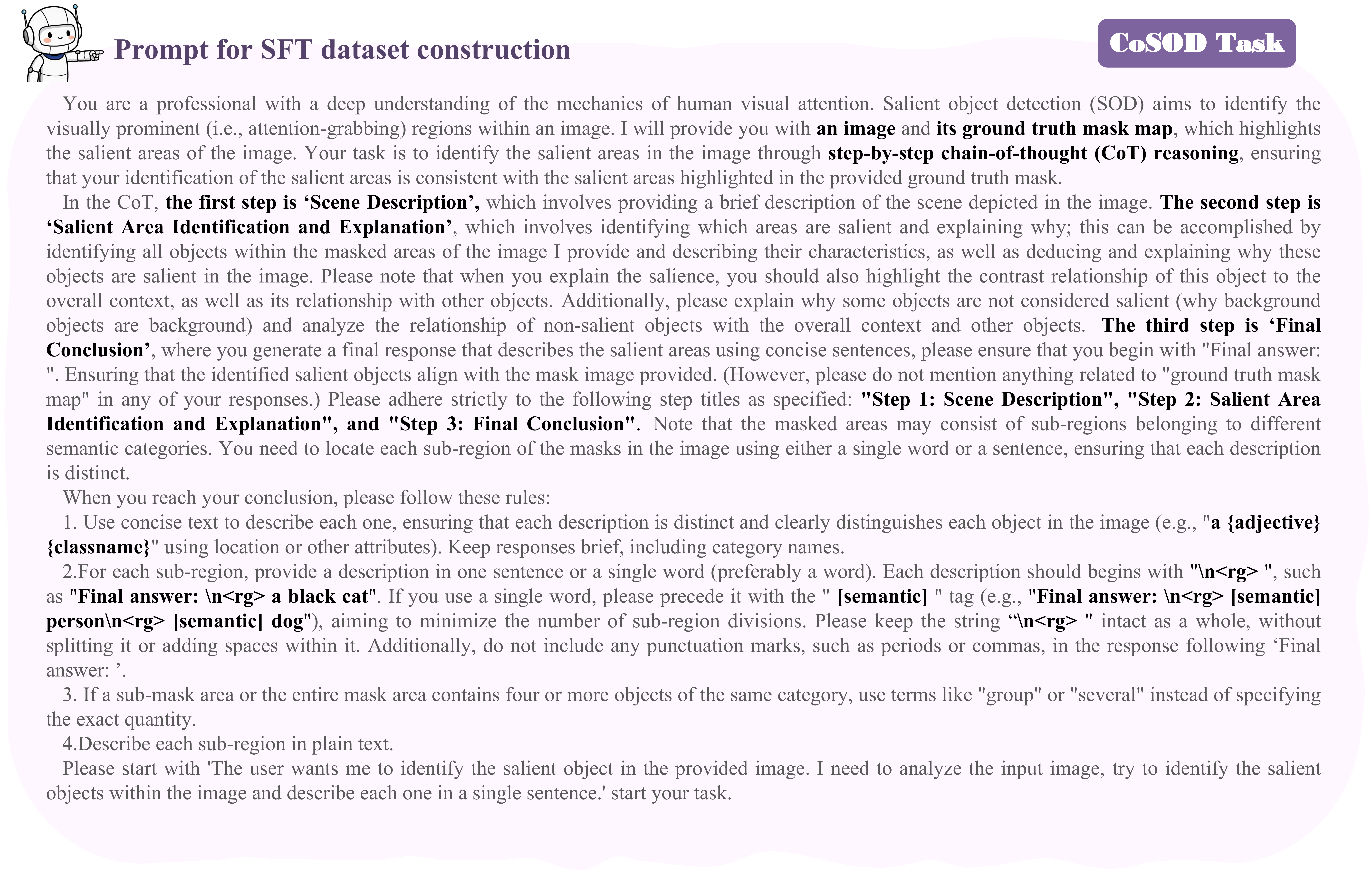}
\vspace{-8mm}
\caption{\textbf{Prompt design for SFT dataset on the Co-salient object detection (CoSOD) tasks.}}
\label{fig:sft_data_prompt_cosod}
\end{figure*}

As illustrated in Figure~\ref{fig:sft_data_prompt_sod}-\ref{fig:sft_data_prompt_cosod}, each task employs a tailored prompt template that guides the model through a fixed three-step CoT structure (scene/object identification → task-specific analysis → conclusion), while ensuring strict adherence to the formatting outlined in Eq.~(1)–(2). In the SOD task, where multiple salient regions may coexist, the model is instructed to enumerate all highlighted areas with concise and distinguishing descriptions within \texttt{<rg>} tags. For the SIS task, instance-wise mask cues prompt the model to produce attribute-rich, disambiguating expressions (\eg, ``a green apple on the left'') within \texttt{<ins>} tags. For the CoSOD task, the model leverages cross-image mask consistency to identify the shared semantic category, outputting this as \texttt{<rg>[semantic] s</rg>}.

This conditioning ensures faithful alignment between reasoning and segmentation, reduces hallucination, and generates high-granularity referring expressions, providing a robust foundation for subsequent CGPO refinement.

\begin{table}[t]
\centering
\footnotesize
\renewcommand{\arraystretch}{1.0}
\renewcommand{\tabcolsep}{3.0mm}
\caption{\textbf{Total parameter count of the MLLM backbone (including vision encoder) across models.} 
``Dense'' denotes models where all parameters are activated in every forward pass; ``MoE'' (Mixture of Experts) denotes models where only a subset of parameters (experts) are activated per token. ``Est.'' indicates community-derived parameter estimates for closed-source models (official values undisclosed). 
Open-source values follow official reports. All models use the same external EVF-SAM segmenter (not counted).}
\label{tab:param_total}
\vspace{-2mm}
\begin{tabular}{p{4.5cm}|c|c}
\toprule[1.25pt]
\rowcolor[RGB]{241, 241, 241}
\multicolumn{1}{c|}{\multirow{1}{*}{\textbf{Model}}} 
& \textbf{Type} & \textbf{Params (B)} \\
\midrule
\rowcolor[RGB]{214,220,229}
\multicolumn{3}{c}{\em Closed-source Models} \\
\midrule
$\circ$ Gemini-2.5-Pro~\cite{comanici2025gemini} & MoE & 175 (Est.) \\
$\circ$ Claude-Sonnet-4.5~\cite{anthropic2025sonnet45system} & Dense & 175+ (Est.) \\
$\circ$ GPT-4o~\cite{hurst2024gpt} & Dense & 200 (Est.) \\
$\circ$ Qwen-VL-Max~\cite{bai2023qwen} & MoE & 200+ (Est.) \\
$\circ$ Grok-4~\cite{xai2025grok} & MoE & 300+ (Est.) \\

\midrule
\rowcolor[RGB]{214,220,229}
\multicolumn{3}{c}{\em Open-source Models} \\
\midrule
$\circ$ Qwen2.5-VL-7B-Instruct~\cite{bai2025qwen2} & Dense & 7 \\
$\circ$ Qwen3-VL-32B-Thinking~\cite{yang2025qwen3} & Dense & 32 \\
$\circ$ InternVL3-9B~\cite{zhu2025internvl3} & Dense & 9 \\
$\circ$ LLaVA-OneVision-Qwen2-7B~\cite{li2024llava} & Dense & 7 \\
$\circ$ GLM-4.1V-9B-Thinking~\cite{V2025glm4.1} & Dense & 9 \\

\midrule
\rowcolor[RGB]{241, 241, 241}
$\bullet$ \textbf{Saliency-R1 (Ours)} & Dense & \textbf{7} \\
\bottomrule[1.25pt]
\end{tabular}
\label{tab:param_count}
\vspace{-4mm}
\end{table}

\section{Parameter Scale Comparison}
\label{sec:param_scale}

We also compare the parameter scales of our model with other MLLMs. As summarized in Table~\ref{tab:param_count}, our Saliency-R1 adopts Qwen2.5-VL-7B-Instruct as the MLLM backbone, with only about 7B dense parameters, substantially smaller than most closed-source counterparts (\eg, GPT-4o~\cite{hurst2024gpt}: $\sim$200B, Gemini-2.5-Pro~\cite{comanici2025gemini}: $\sim$175B MoE) and even smaller than many open-source alternatives (\eg, Qwen3-VL-32B-Thinking~\cite{yang2025qwen3}: 32B). Despite this compact scale, Saliency-R1 significantly outperforms all listed models across SOD, SIS, and CoSOD (see Table~3 in the main text), indicating that its superiority stems not from model size, but from the synergistic design of our \textbf{confidence-guided RL training (CGPO)} in aligning the MLLM’s reasoning with visual saliency priors, enabling more accurate, robust, and task-coherent CoT generation.

\section{Qualitative Comparison}
\label{sec:qualitative}
We conduct comprehensive qualitative analyses to demonstrate that Saliency-R1’s reasoning quality drives superior mask fidelity across heterogeneous saliency tasks. As shown in Figure~\ref{MLLM_SOD}-\ref{SOTA_SOD}, our method matches or exceeds the segmentation performance of both closed/open-source MLLMs and specialized saliency models across diverse and challenging scenarios, \eg, instance occlusion, scale variation, reflection interference, and cross-image consensus reasoning. These results highlight the effectiveness of our Saliency-R1 framework. 

\subsection{Comparison with MLLMs}
\begin{figure}[t] 
\begin{minipage}[c]{1.\columnwidth} 
	\scriptsize
	\renewcommand{\tabcolsep}{0.5pt} 
	\renewcommand{\arraystretch}{1.2} 
	\centering
        \begin{tabular}{ccccc}
	    \\

            \arrayrulecolor{red}
            \rotatebox[origin=c]{90}{\small (a)}
            &
		\makecell[c]{\includegraphics[width=0.225\linewidth,height=0.15\linewidth]{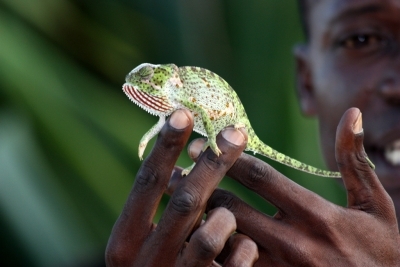}}
        &
		\makecell[c]{\includegraphics[width=0.225\linewidth,height=0.15\linewidth]{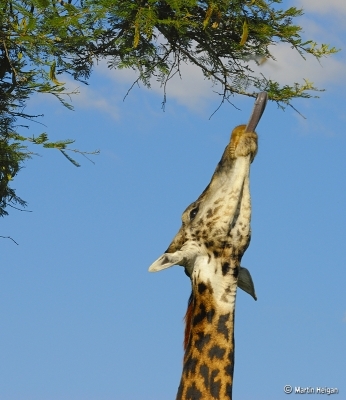}}
		&
		\makecell[c]{\includegraphics[width=0.225\linewidth,height=0.15\linewidth]{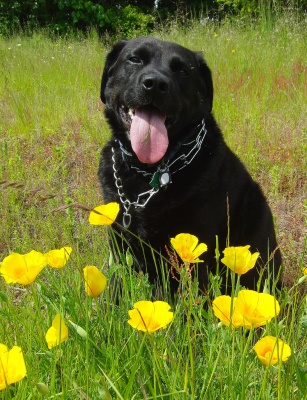}}
		&
		\makecell[c]{\includegraphics[width=0.225\linewidth,height=0.15\linewidth]{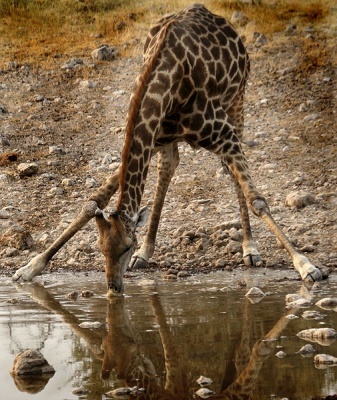}}
            \vspace{-0.5mm}
            \\

            \rotatebox[origin=c]{90}{\small (b)}
            &
		\makecell[c]{\includegraphics[width=0.225\linewidth,height=0.15\linewidth]{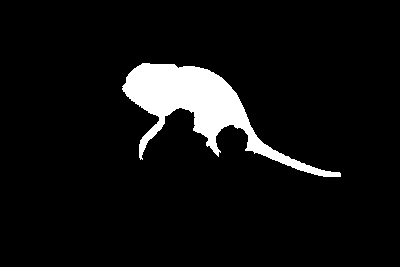}}
        &
		\makecell[c]{\includegraphics[width=0.225\linewidth,height=0.15\linewidth]{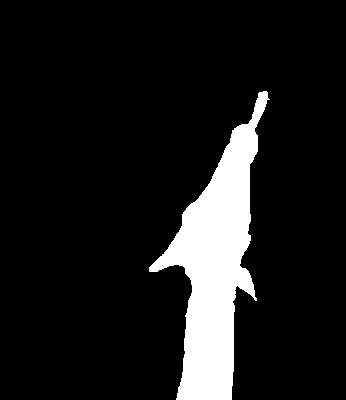}}
		&
		\makecell[c]{\includegraphics[width=0.225\linewidth,height=0.15\linewidth]{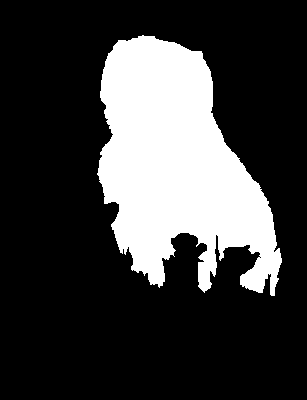}}
		&
		\makecell[c]{\includegraphics[width=0.225\linewidth,height=0.15\linewidth]{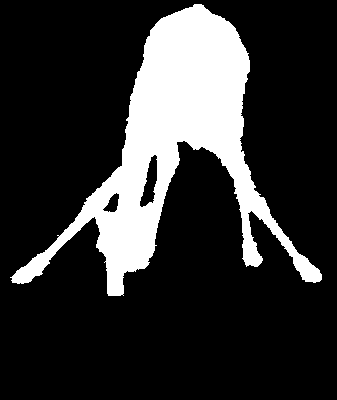}}
            \vspace{-0.5mm}
		\\

            \rotatebox[origin=c]{90}{\small (c)}  
            &
		\makecell[c]{\includegraphics[width=0.225\linewidth,height=0.15\linewidth]{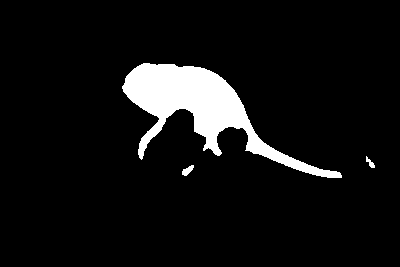}}
        &
		\makecell[c]{\includegraphics[width=0.225\linewidth,height=0.15\linewidth]{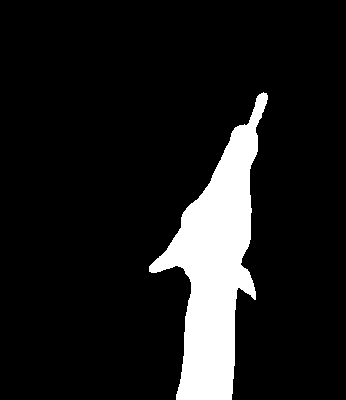}}
		&
		\makecell[c]{\includegraphics[width=0.225\linewidth,height=0.15\linewidth]{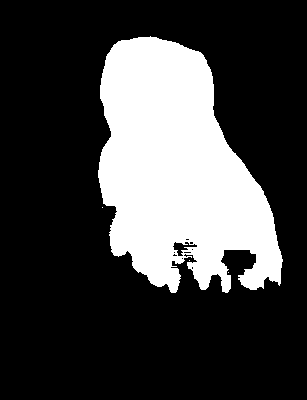}}
		&
		\makecell[c]{\includegraphics[width=0.225\linewidth,height=0.15\linewidth]{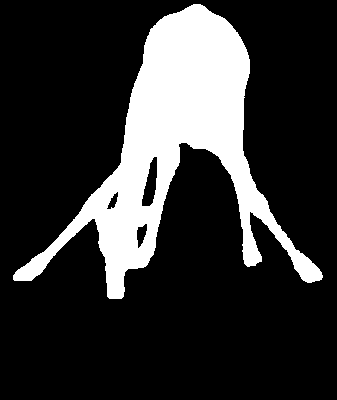}}      
            \vspace{-0.5mm}
		\\

            \rotatebox[origin=c]{90}{\small (d)}           
            &
		\makecell[c]{\includegraphics[width=0.225\linewidth,height=0.15\linewidth]{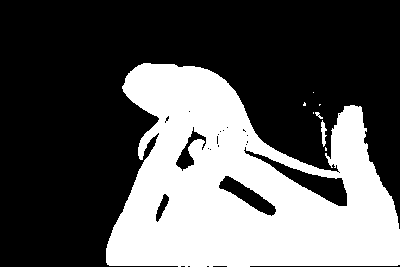}}
        &
		\makecell[c]{\includegraphics[width=0.225\linewidth,height=0.15\linewidth]{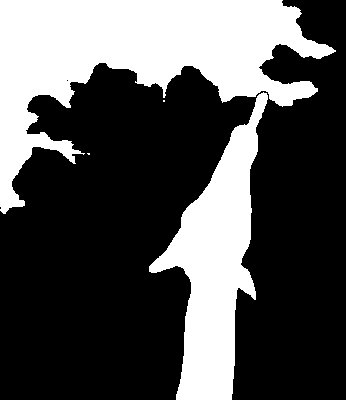}}
		&
		\makecell[c]{\includegraphics[width=0.225\linewidth,height=0.15\linewidth]{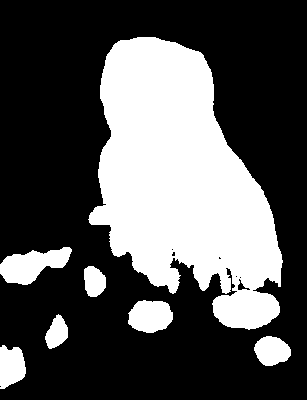}}
		&
		\makecell[c]{\includegraphics[width=0.225\linewidth,height=0.15\linewidth]{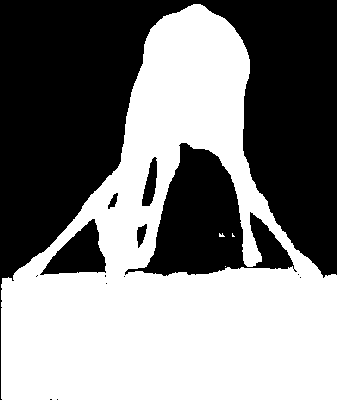}}  
            \vspace{-0.5mm}
		\\

             \rotatebox[origin=c]{90}{\small (e)}
            &
		\makecell[c]{\includegraphics[width=0.225\linewidth,height=0.15\linewidth]{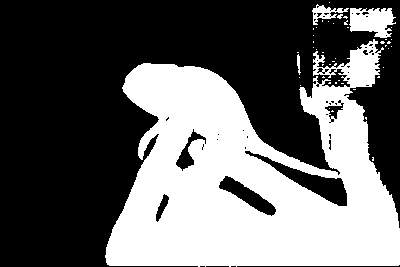}}
        &
		\makecell[c]{\includegraphics[width=0.225\linewidth,height=0.15\linewidth]{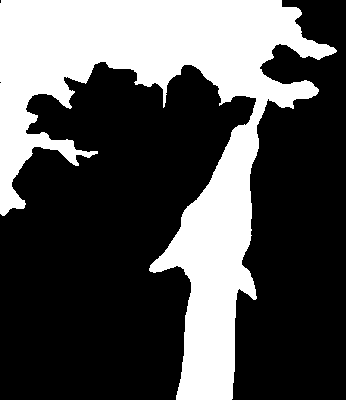}}
		&
		\makecell[c]{\includegraphics[width=0.225\linewidth,height=0.15\linewidth]{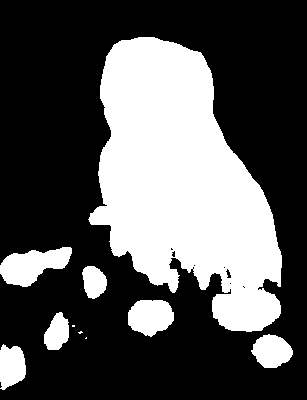}}
		&
		\makecell[c]{\includegraphics[width=0.225\linewidth,height=0.15\linewidth]{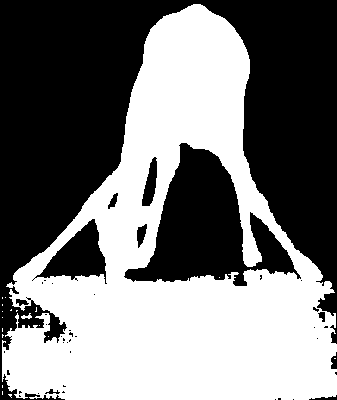}} 
            \vspace{-0.5mm}
		\\

            \rotatebox[origin=c]{90}{\small (f)}
            &
		\makecell[c]{\includegraphics[width=0.225\linewidth,height=0.15\linewidth]{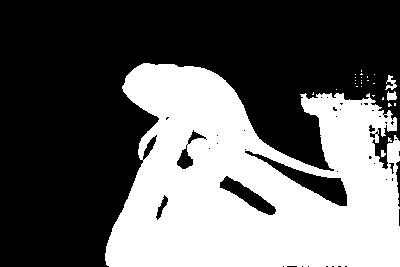}}
        &
		\makecell[c]{\includegraphics[width=0.225\linewidth,height=0.15\linewidth]{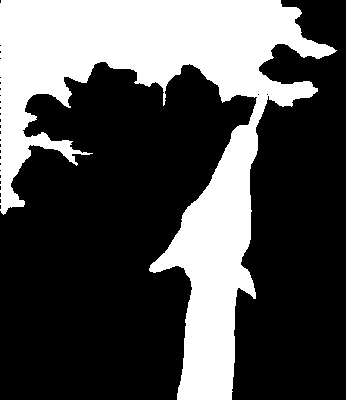}}
		&
		\makecell[c]{\includegraphics[width=0.225\linewidth,height=0.15\linewidth]{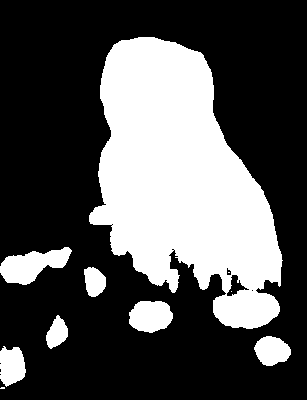}}
		&
		\makecell[c]{\includegraphics[width=0.225\linewidth,height=0.15\linewidth]{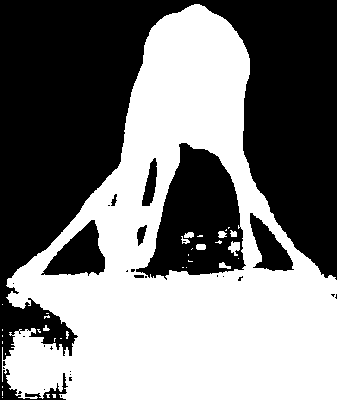}}   
            \vspace{-0.5mm}
		\\

            \rotatebox[origin=c]{90}{\small (g)}
            &
		\makecell[c]{\includegraphics[width=0.225\linewidth,height=0.15\linewidth]{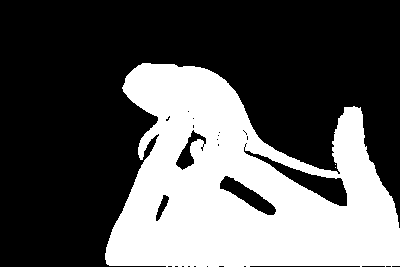}}
        &
		\makecell[c]{\includegraphics[width=0.225\linewidth,height=0.15\linewidth]{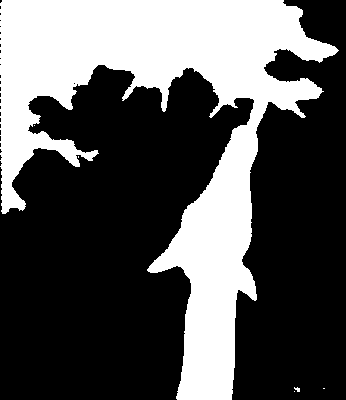}}
		&
		\makecell[c]{\includegraphics[width=0.225\linewidth,height=0.15\linewidth]{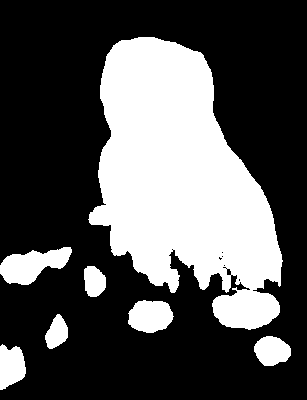}}
		&
		\makecell[c]{\includegraphics[width=0.225\linewidth,height=0.15\linewidth]{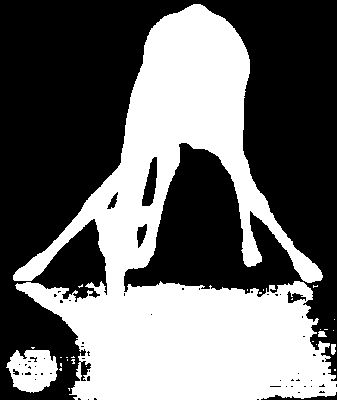}}     
            \vspace{-0.5mm}
		\\

        \rotatebox[origin=c]{90}{\small (h)}
            &
		\makecell[c]{\includegraphics[width=0.225\linewidth,height=0.15\linewidth]{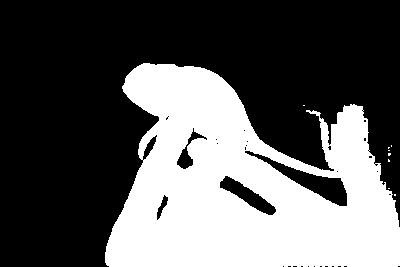}}
        &
		\makecell[c]{\includegraphics[width=0.225\linewidth,height=0.15\linewidth]{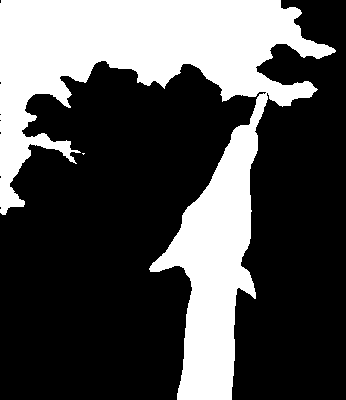}}
		&
		\makecell[c]{\includegraphics[width=0.225\linewidth,height=0.15\linewidth]{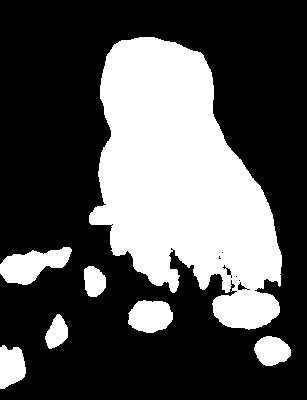}}
		&
		\makecell[c]{\includegraphics[width=0.225\linewidth,height=0.15\linewidth]{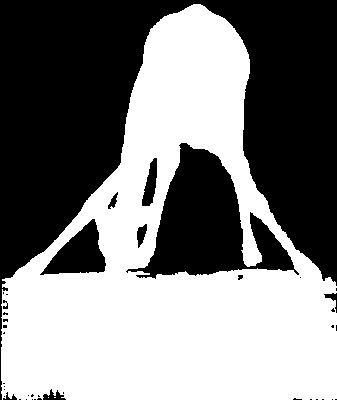}}     
            \vspace{-0.5mm}
		\\

        \rotatebox[origin=c]{90}{\small (i)}
            &
		\makecell[c]{\includegraphics[width=0.225\linewidth,height=0.15\linewidth]{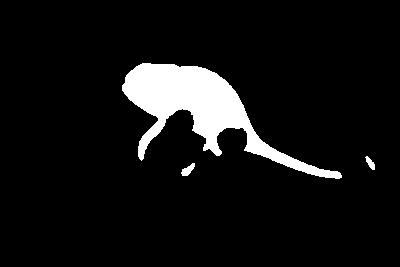}}
        &
		\makecell[c]{\includegraphics[width=0.225\linewidth,height=0.15\linewidth]{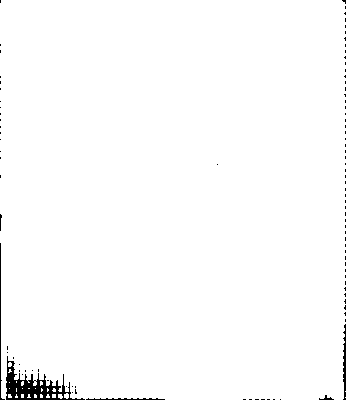}}
		&
		\makecell[c]{\includegraphics[width=0.225\linewidth,height=0.15\linewidth]{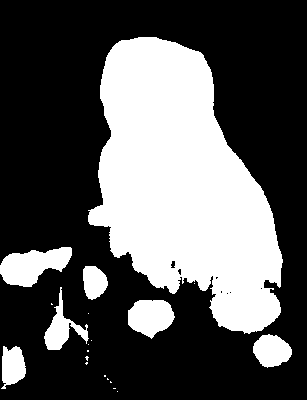}}
		&
		\makecell[c]{\includegraphics[width=0.225\linewidth,height=0.15\linewidth]{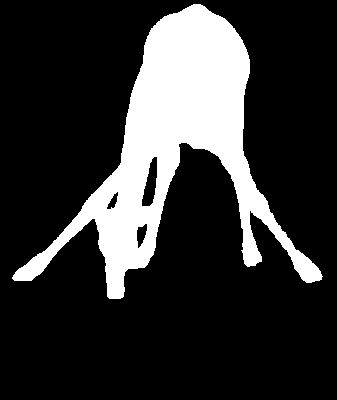}}     
            \vspace{-0.5mm}
		\\

        \rotatebox[origin=c]{90}{\small (j)}
            &
		\makecell[c]{\includegraphics[width=0.225\linewidth,height=0.15\linewidth]{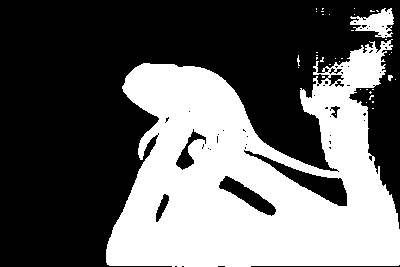}}
        &
		\makecell[c]{\includegraphics[width=0.225\linewidth,height=0.15\linewidth]{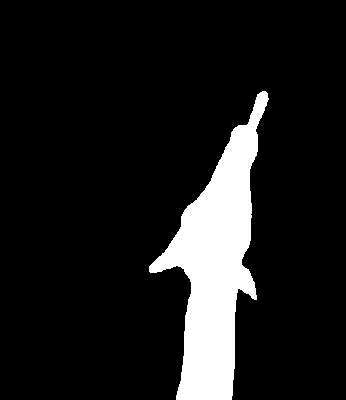}}
		&
		\makecell[c]{\includegraphics[width=0.225\linewidth,height=0.15\linewidth]{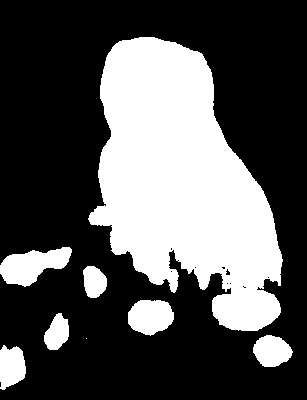}}
		&
		\makecell[c]{\includegraphics[width=0.225\linewidth,height=0.15\linewidth]{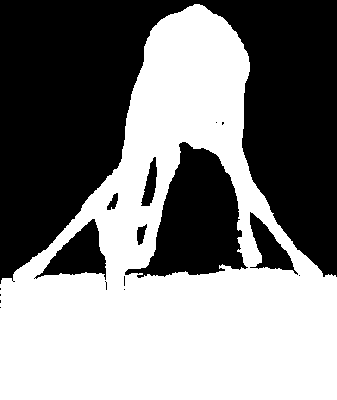}}     
            \vspace{-0.5mm}
		\\

		\end{tabular}
            \vspace{-2mm}
    \caption{\textbf{Qualitative comparison with other MLLMs on SOD.} \textbf{(a)} denotes the input image, \textbf{(b)} the ground-truth mask, \textbf{(c)} our Saliency-R1, followed by closed- and open-source MLLMs: \textbf{(d)} Claude-Sonnet-4.5, \textbf{(e)} Gemini-2.5-Pro, \textbf{(f)} GPT-4o, \textbf{(g)} Grok-4, \textbf{(h)} Qwen3-VL-32B-Thinking, \textbf{(i)} Qwen2.5-VL-7B-Instruct, and \textbf{(j)} LLaVA-OneVision-Qwen2-7B.}
    \label{MLLM_SOD}
\end{minipage}
\end{figure}

\begin{figure}[t] 
\begin{minipage}[c]{1.\columnwidth}
	\scriptsize
	\renewcommand{\tabcolsep}{0.5pt} 
	\renewcommand{\arraystretch}{1.2} 
	\centering
        \begin{tabular}{ccccc}
	    \\

            \arrayrulecolor{red}
            \rotatebox[origin=c]{90}{\small (a)}
            &
		\makecell[c]{\includegraphics[width=0.225\linewidth,height=0.15\linewidth]{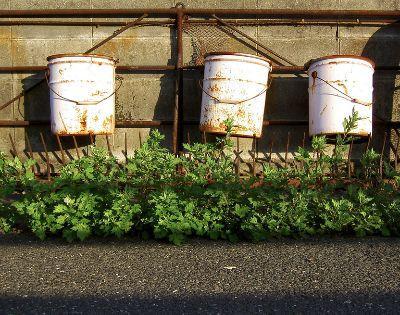}}
        &
		\makecell[c]{\includegraphics[width=0.225\linewidth,height=0.15\linewidth]{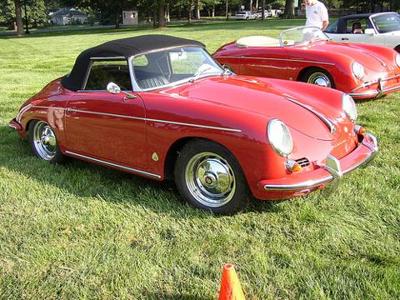}}
		&
		\makecell[c]{\includegraphics[width=0.225\linewidth,height=0.15\linewidth]{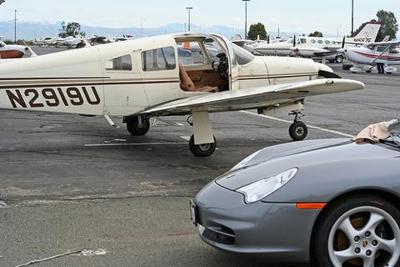}}
		&
		\makecell[c]{\includegraphics[width=0.225\linewidth,height=0.15\linewidth]{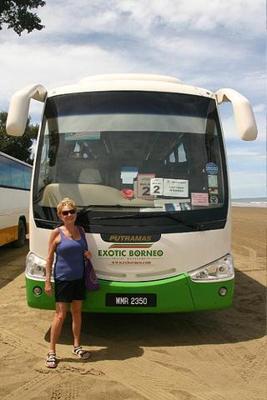}}
            \vspace{-0.5mm}
            \\

            \rotatebox[origin=c]{90}{\small (b)}
            &
		\makecell[c]{\includegraphics[width=0.225\linewidth,height=0.15\linewidth]{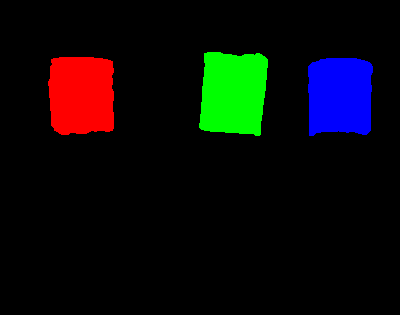}}
        &
		\makecell[c]{\includegraphics[width=0.225\linewidth,height=0.15\linewidth]{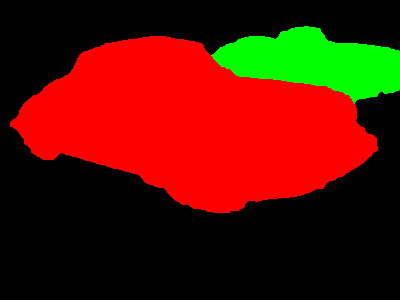}}
		&
		\makecell[c]{\includegraphics[width=0.225\linewidth,height=0.15\linewidth]{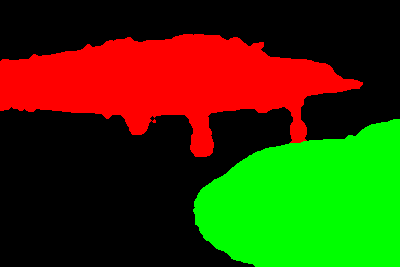}}
		&
		\makecell[c]{\includegraphics[width=0.225\linewidth,height=0.15\linewidth]{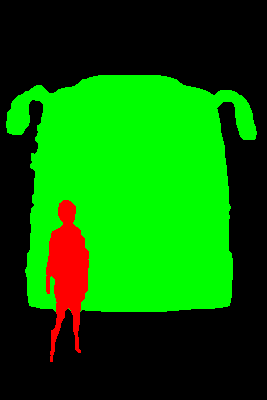}}
            \vspace{-0.5mm}
		\\

            \rotatebox[origin=c]{90}{\small (c)}  
            &
		\makecell[c]{\includegraphics[width=0.225\linewidth,height=0.15\linewidth]{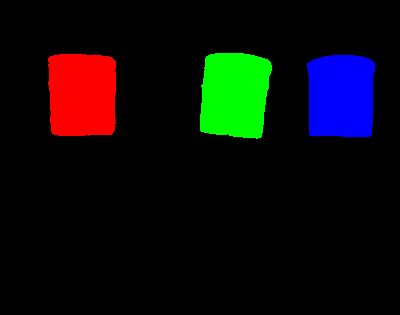}}
        &
		\makecell[c]{\includegraphics[width=0.225\linewidth,height=0.15\linewidth]{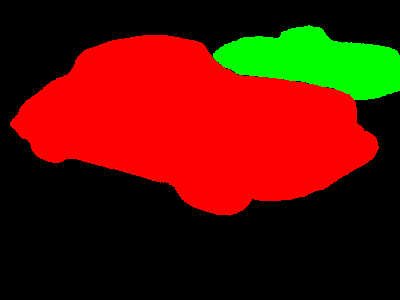}}
		&
		\makecell[c]{\includegraphics[width=0.225\linewidth,height=0.15\linewidth]{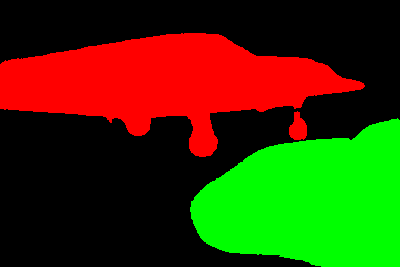}}
		&
		\makecell[c]{\includegraphics[width=0.225\linewidth,height=0.15\linewidth]{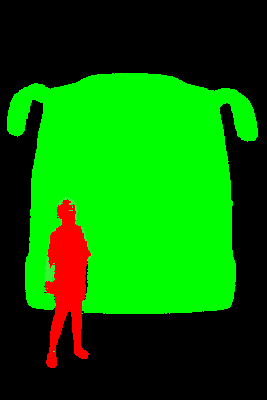}}   
            \vspace{-0.5mm}
		\\

            \rotatebox[origin=c]{90}{\small (d)}           
            &
		\makecell[c]{\includegraphics[width=0.225\linewidth,height=0.15\linewidth]{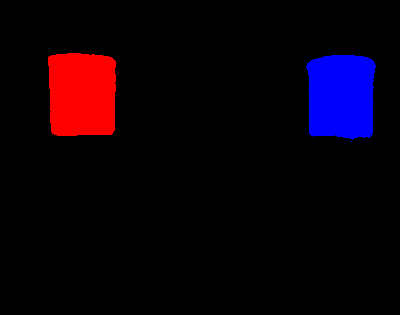}}
        &
		\makecell[c]{\includegraphics[width=0.225\linewidth,height=0.15\linewidth]{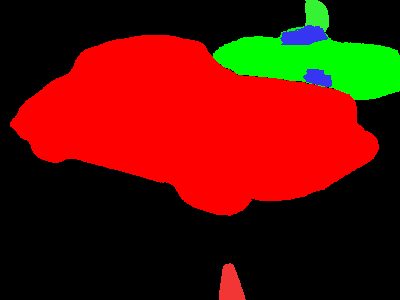}}
		&
		\makecell[c]{\includegraphics[width=0.225\linewidth,height=0.15\linewidth]{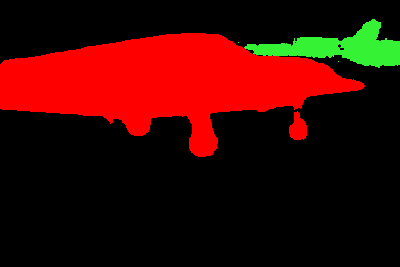}}
		&
		\makecell[c]{\includegraphics[width=0.225\linewidth,height=0.15\linewidth]{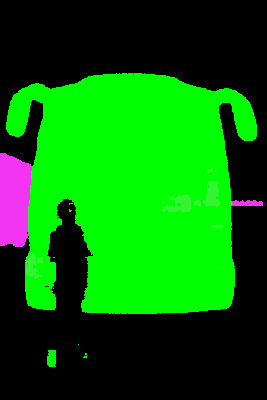}}
            \vspace{-0.5mm}
		\\

             \rotatebox[origin=c]{90}{\small (e)}
            &
		\makecell[c]{\includegraphics[width=0.225\linewidth,height=0.15\linewidth]{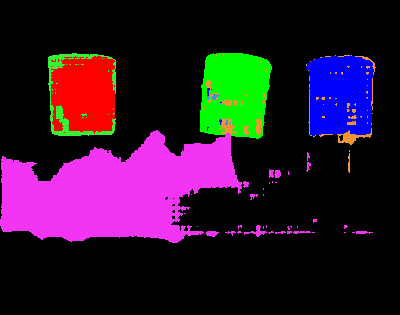}}
        &
		\makecell[c]{\includegraphics[width=0.225\linewidth,height=0.15\linewidth]{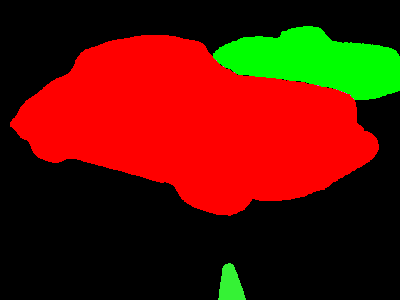}}
		&
		\makecell[c]{\includegraphics[width=0.225\linewidth,height=0.15\linewidth]{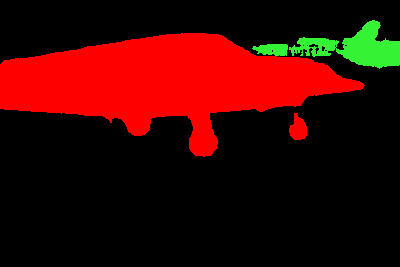}}
		&
		\makecell[c]{\includegraphics[width=0.225\linewidth,height=0.15\linewidth]{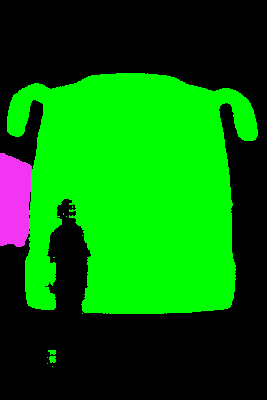}}
            \vspace{-0.5mm}
		\\

            \rotatebox[origin=c]{90}{\small (f)}
            &
		\makecell[c]{\includegraphics[width=0.225\linewidth,height=0.15\linewidth]{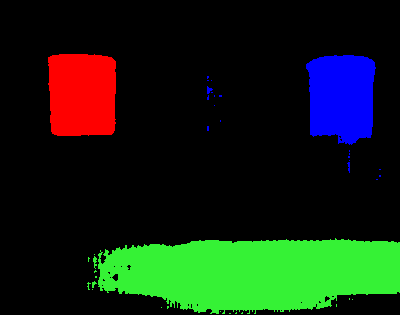}}
        &
		\makecell[c]{\includegraphics[width=0.225\linewidth,height=0.15\linewidth]{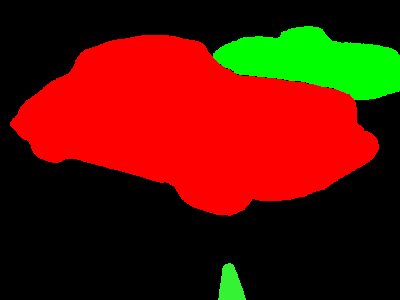}}
		&
		\makecell[c]{\includegraphics[width=0.225\linewidth,height=0.15\linewidth]{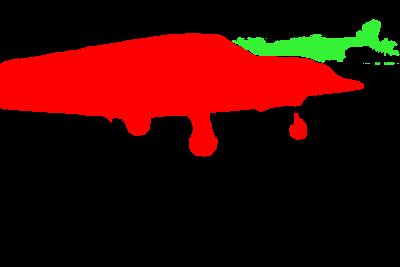}}
		&
		\makecell[c]{\includegraphics[width=0.225\linewidth,height=0.15\linewidth]{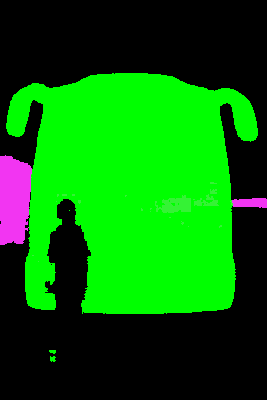}}
            \vspace{-0.5mm}
		\\

        \rotatebox[origin=c]{90}{\small (g)}
            &
		\makecell[c]{\includegraphics[width=0.225\linewidth,height=0.15\linewidth]{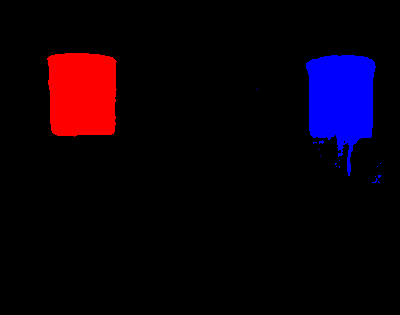}}
        &
		\makecell[c]{\includegraphics[width=0.225\linewidth,height=0.15\linewidth]{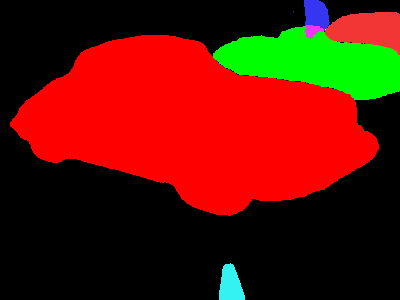}}
		&
		\makecell[c]{\includegraphics[width=0.225\linewidth,height=0.15\linewidth]{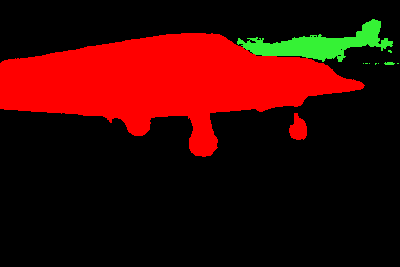}}
		&
		\makecell[c]{\includegraphics[width=0.225\linewidth,height=0.15\linewidth]{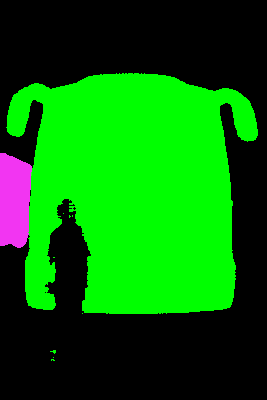}}
\\
        \rotatebox[origin=c]{90}{\small (h)}
            &
		\makecell[c]{\includegraphics[width=0.225\linewidth,height=0.15\linewidth]{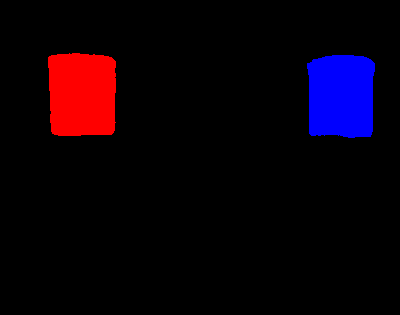}}
        &
		\makecell[c]{\includegraphics[width=0.225\linewidth,height=0.15\linewidth]{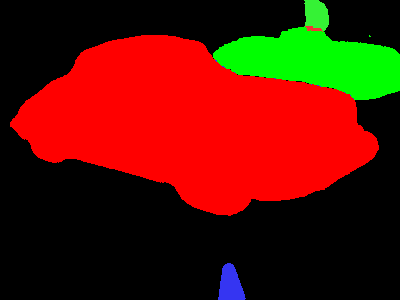}}
		&
		\makecell[c]{\includegraphics[width=0.225\linewidth,height=0.15\linewidth]{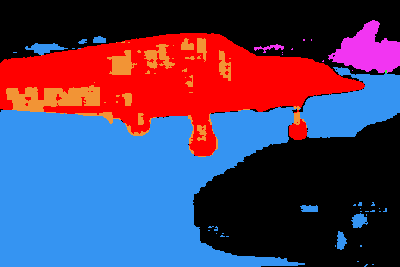}}
		&
		\makecell[c]{\includegraphics[width=0.225\linewidth,height=0.15\linewidth]{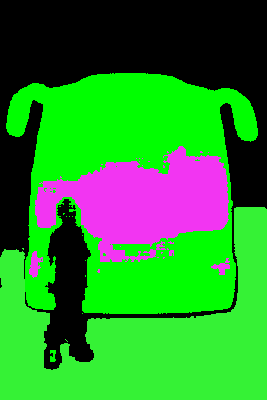}}     
            \vspace{-0.5mm}
		\\

        \rotatebox[origin=c]{90}{\small (i)}
            &
		\makecell[c]{\includegraphics[width=0.225\linewidth,height=0.15\linewidth]{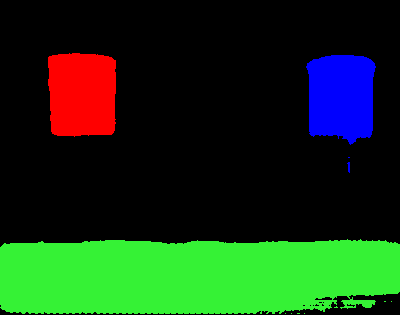}}
        &
		\makecell[c]{\includegraphics[width=0.225\linewidth,height=0.15\linewidth]{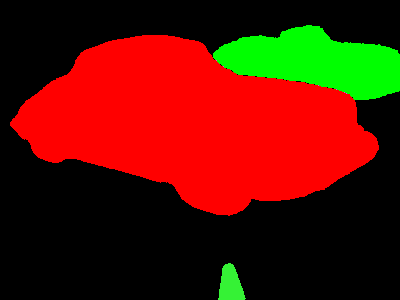}}
		&
		\makecell[c]{\includegraphics[width=0.225\linewidth,height=0.15\linewidth]{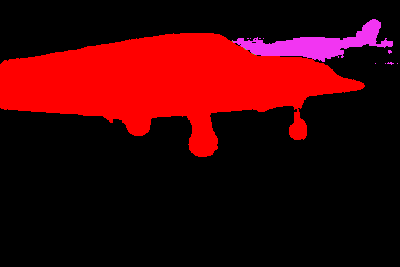}}
		&
		\makecell[c]{\includegraphics[width=0.225\linewidth,height=0.15\linewidth]{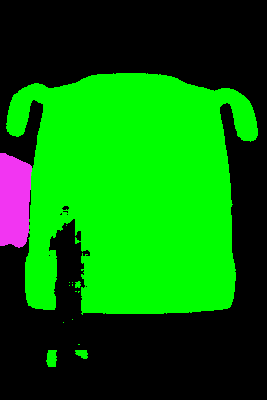}}     
            \vspace{-0.5mm}
		\\

        \rotatebox[origin=c]{90}{\small (j)}
            &
		\makecell[c]{\includegraphics[width=0.225\linewidth,height=0.15\linewidth]{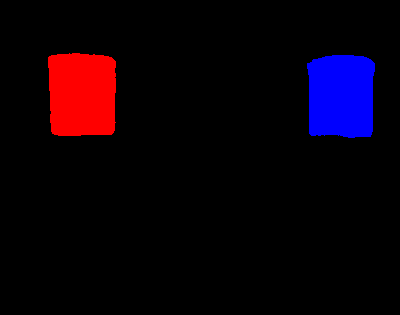}}
        &
		\makecell[c]{\includegraphics[width=0.225\linewidth,height=0.15\linewidth]{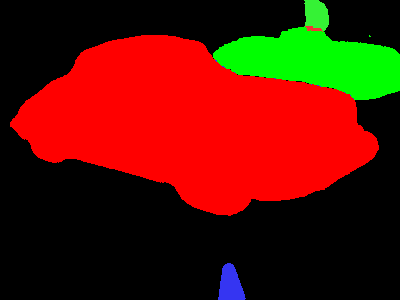}}
		&
		\makecell[c]{\includegraphics[width=0.225\linewidth,height=0.15\linewidth]{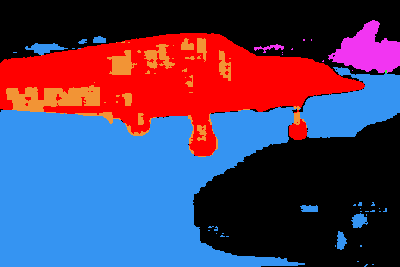}}
		&
		\makecell[c]{\includegraphics[width=0.225\linewidth,height=0.15\linewidth]{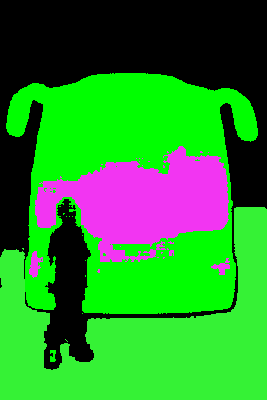}}     
            \vspace{-0.5mm}
		\\
		\end{tabular}
            \vspace{-2mm}
    \caption{\textbf{Qualitative comparison with other MLLMs on SIS.} \textbf{(a)} denotes the input image, \textbf{(b)} the ground-truth instance masks, \textbf{(c)} our Saliency-R1, followed by: \textbf{(d)} Claude-Sonnet-4.5, \textbf{(e)} Gemini-2.5-Pro, \textbf{(f)} GPT-4o, \textbf{(g)} Grok-4, \textbf{(h)} Qwen3-VL-32B-Thinking, \textbf{(i)} Qwen2.5-VL-7B-Instruct, and \textbf{(j)} LLaVA-OneVision-Qwen2-7B.}
    \label{MLLM_SIS}
\end{minipage}
\end{figure}

\begin{figure*}[t]
\begin{minipage}[c]{1.\textwidth}
	\scriptsize
	\renewcommand{\tabcolsep}{0.5pt} 
	\renewcommand{\arraystretch}{0.9} 
	\centering
        \begin{tabular}{cccccccccccccccc}
	    &
            \multicolumn{4}{c}{\cellcolor[RGB]
            {255,230,153}$\bold{penguin}$}
            &
            \multicolumn{4}{c}{\cellcolor[RGB]{244,177,131}$\bold{goldenfish}$}
	    &
            \multicolumn{4}{c}{\cellcolor[RGB]{169,209,142}$\bold{boat}$}
            \vspace{0.5mm}
	    \\

            \arrayrulecolor{red}
            \rotatebox[origin=c]{90}{\small (a)}
            &
		\makecell[c]{\includegraphics[width=0.08\linewidth,height=0.07\linewidth]{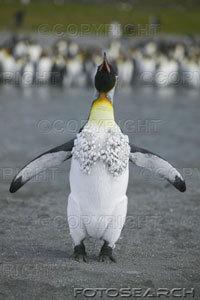}}
        &
		\makecell[c]{\includegraphics[width=0.08\linewidth,height=0.07\linewidth]{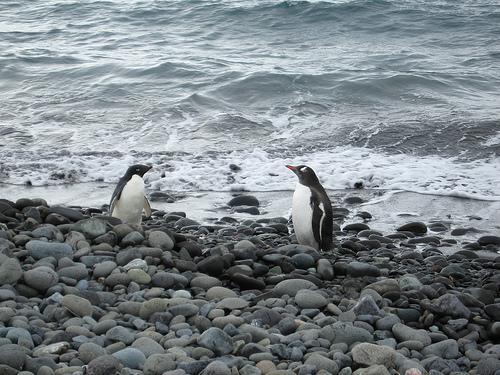}} 
		&
		\makecell[c]{\includegraphics[width=0.08\linewidth,height=0.07\linewidth]{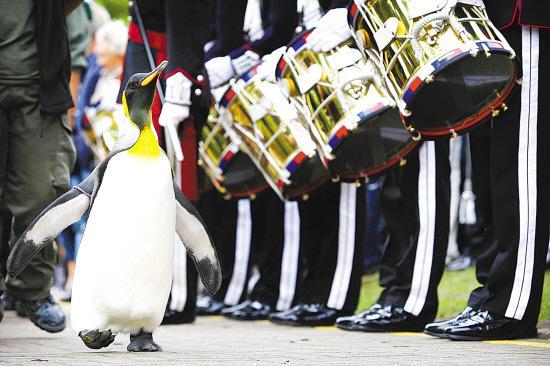}} 
		&
		\makecell[c]{\includegraphics[width=0.08\linewidth,height=0.07\linewidth]{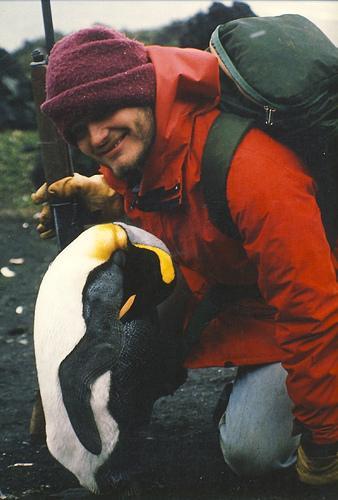}}
        &
		\makecell[c]{\includegraphics[width=0.08\linewidth,height=0.07\linewidth]{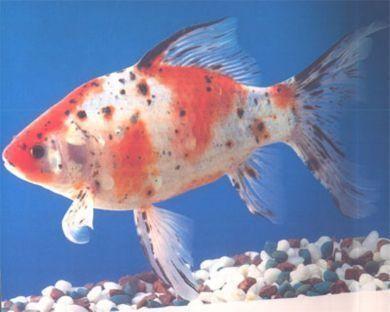}} 
		&
		\makecell[c]{\includegraphics[width=0.08\linewidth,height=0.07\linewidth]{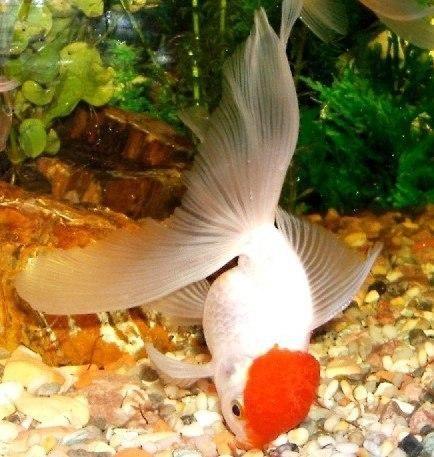}} 
		&
		\makecell[c]{\includegraphics[width=0.08\linewidth,height=0.07\linewidth]{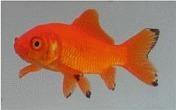}} 
		&
		\makecell[c]{\includegraphics[width=0.08\linewidth,height=0.07\linewidth]{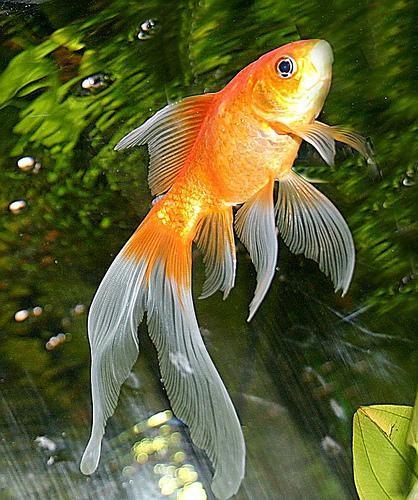}}         
		&
		\makecell[c]{\includegraphics[width=0.08\linewidth,height=0.07\linewidth]{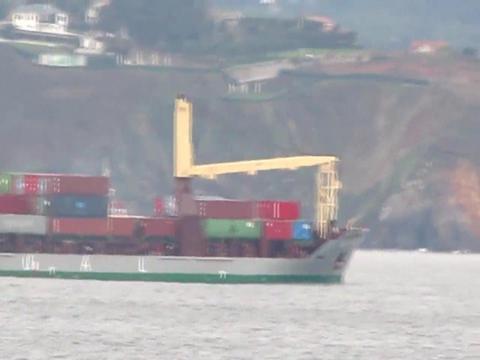}} 
		&
		\makecell[c]{\includegraphics[width=0.08\linewidth,height=0.07\linewidth]{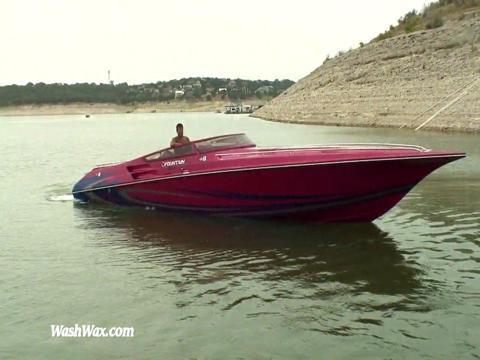}}
            &
		\makecell[c]{\includegraphics[width=0.08\linewidth,height=0.07\linewidth]{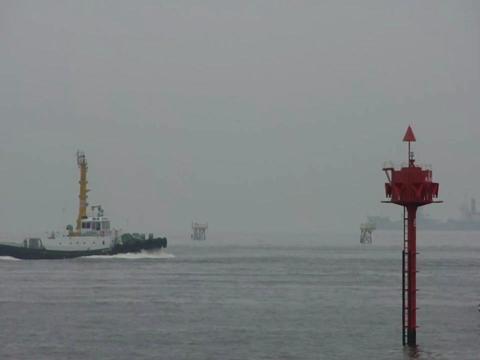}} 
		&
		\makecell[c]{\includegraphics[width=0.08\linewidth,height=0.07\linewidth]{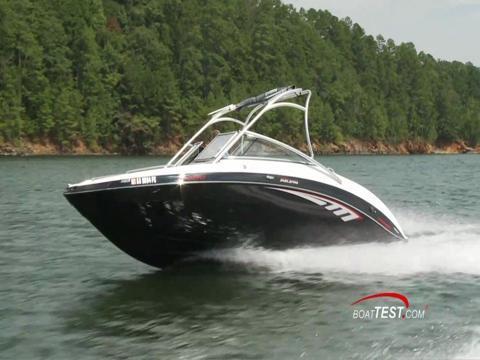}} 
 
            \vspace{-0.5mm}
            \\

            \rotatebox[origin=c]{90}{\small (b))}
            &
		\makecell[c]{\includegraphics[width=0.08\linewidth,height=0.07\linewidth]{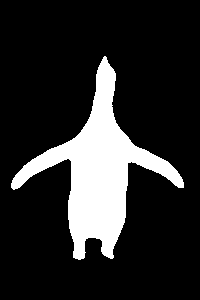}} 
		&
		\makecell[c]{\includegraphics[width=0.08\linewidth,height=0.07\linewidth]{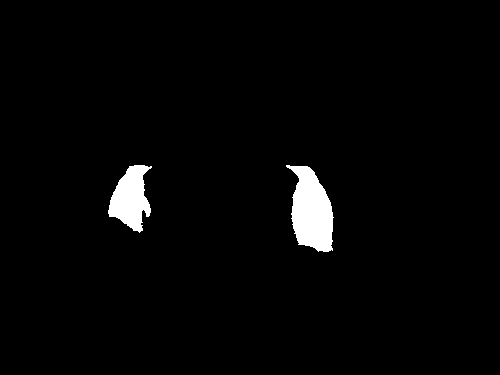}} 
		&
		\makecell[c]{\includegraphics[width=0.08\linewidth,height=0.07\linewidth]{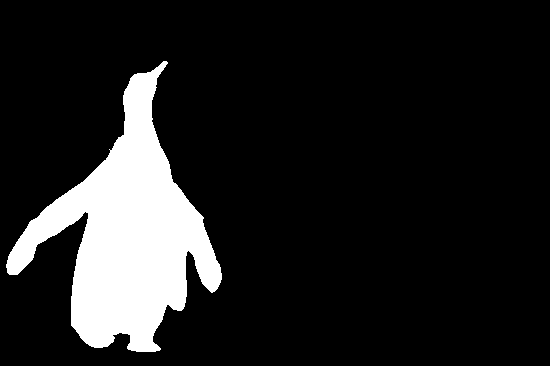}} 
		&
		\makecell[c]{\includegraphics[width=0.08\linewidth,height=0.07\linewidth]{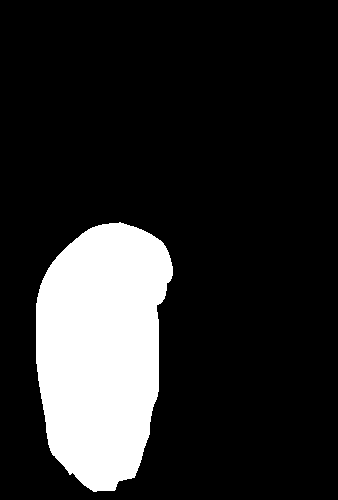}} 
		&
		\makecell[c]{\includegraphics[width=0.08\linewidth,height=0.07\linewidth]{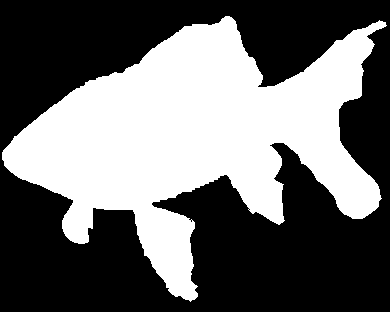}} 
		&
		\makecell[c]{\includegraphics[width=0.08\linewidth,height=0.07\linewidth]{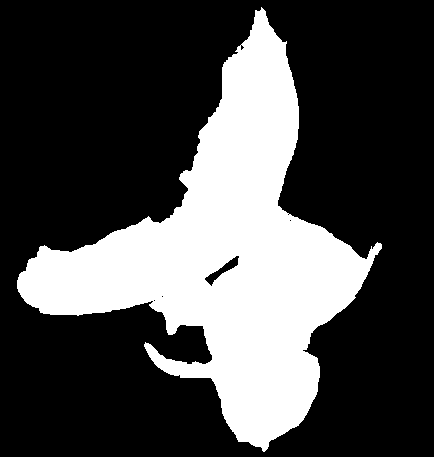}} 
		&
		\makecell[c]{\includegraphics[width=0.08\linewidth,height=0.07\linewidth]{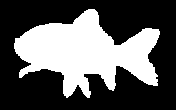}} 
		&
		\makecell[c]{\includegraphics[width=0.08\linewidth,height=0.07\linewidth]{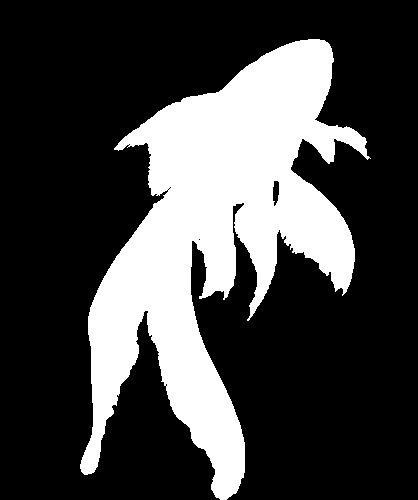}}
        &
		\makecell[c]{\includegraphics[width=0.08\linewidth,height=0.07\linewidth]{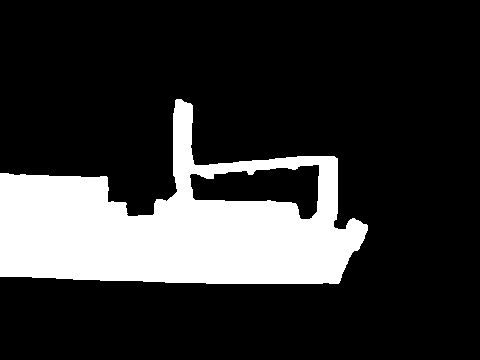}} 
		&
        \makecell[c]{\includegraphics[width=0.08\linewidth,height=0.07\linewidth]{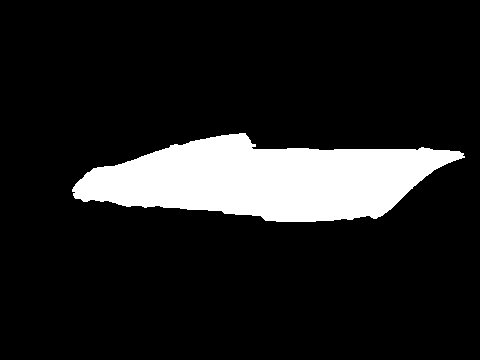}} 
		&
		\makecell[c]{\includegraphics[width=0.08\linewidth,height=0.07\linewidth]{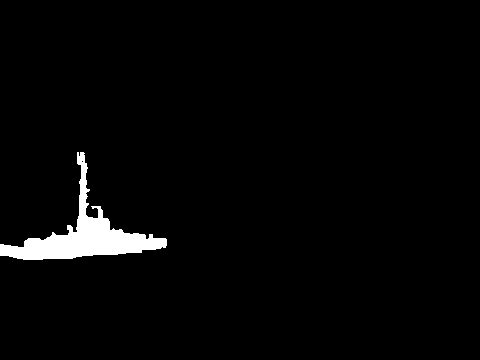}}
            &
		\makecell[c]{\includegraphics[width=0.08\linewidth,height=0.07\linewidth]{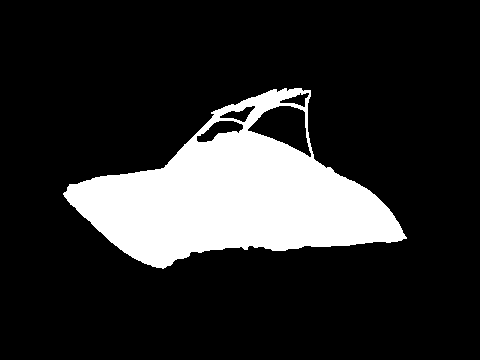}} 
            \vspace{-0.5mm}
		\\

            \rotatebox[origin=c]{90}{\small (c)}  
        &
		\makecell[c]{\includegraphics[width=0.08\linewidth,height=0.07\linewidth]{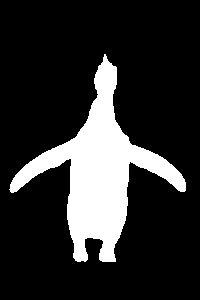}} 
		&
		\makecell[c]{\includegraphics[width=0.08\linewidth,height=0.07\linewidth]{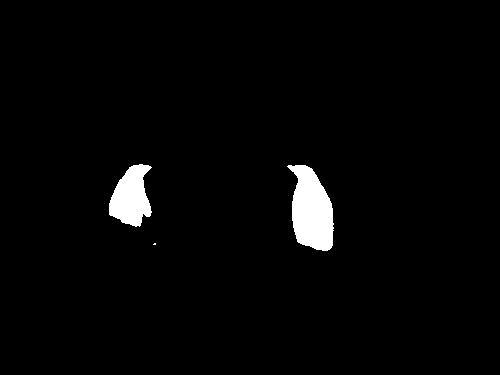}} 
		&
		\makecell[c]{\includegraphics[width=0.08\linewidth,height=0.07\linewidth]{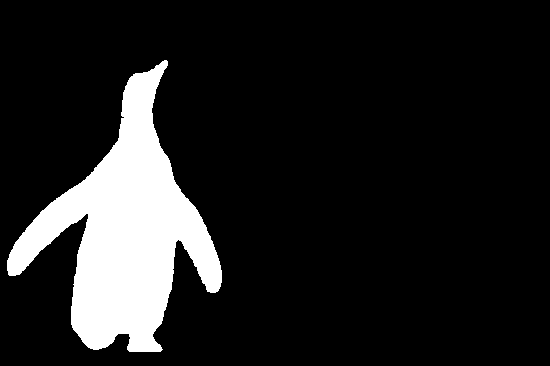}} 
		&
		\makecell[c]{\includegraphics[width=0.08\linewidth,height=0.07\linewidth]{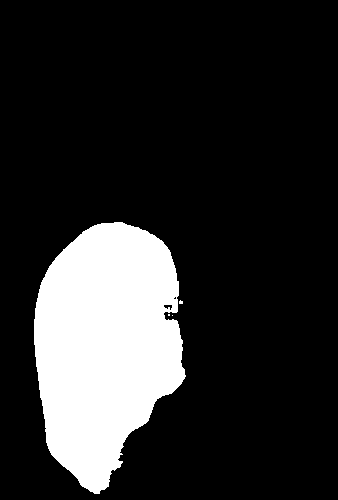}}        
		&
		\makecell[c]{\includegraphics[width=0.08\linewidth,height=0.07\linewidth]{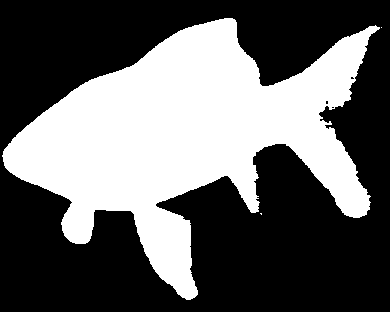}} 
		&
		\makecell[c]{\includegraphics[width=0.08\linewidth,height=0.07\linewidth]{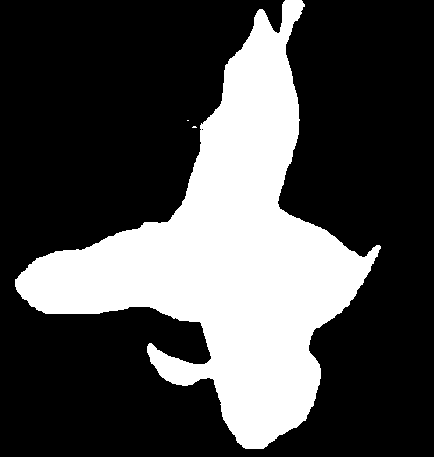}} 
		&
		\makecell[c]{\includegraphics[width=0.08\linewidth,height=0.07\linewidth]{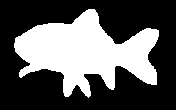}} 
		&
		\makecell[c]{\includegraphics[width=0.08\linewidth,height=0.07\linewidth]{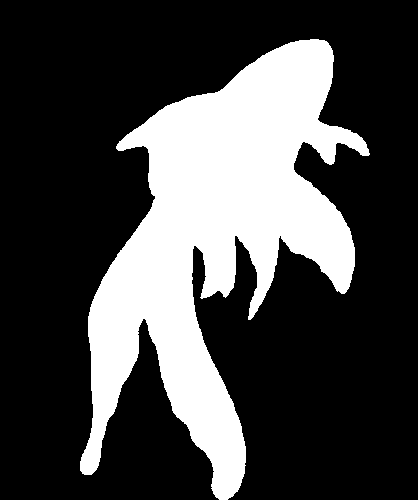}}
        &
		\makecell[c]{\includegraphics[width=0.08\linewidth,height=0.07\linewidth]{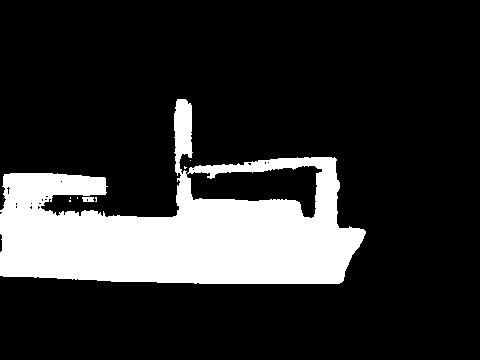}} 
		&
		\makecell[c]{\includegraphics[width=0.08\linewidth,height=0.07\linewidth]{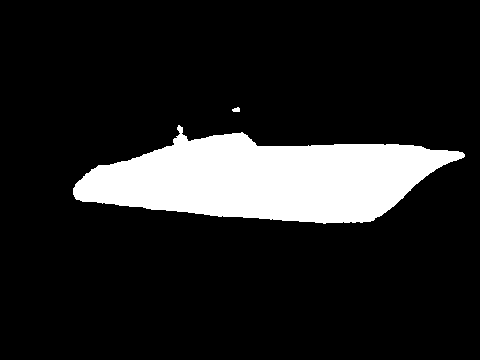}} 
		&
		\makecell[c]{\includegraphics[width=0.08\linewidth,height=0.07\linewidth]{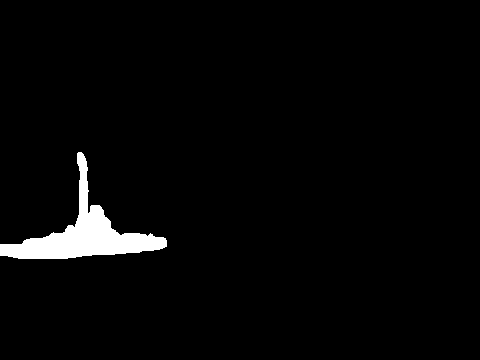}}
        &
		\makecell[c]{\includegraphics[width=0.08\linewidth,height=0.07\linewidth]{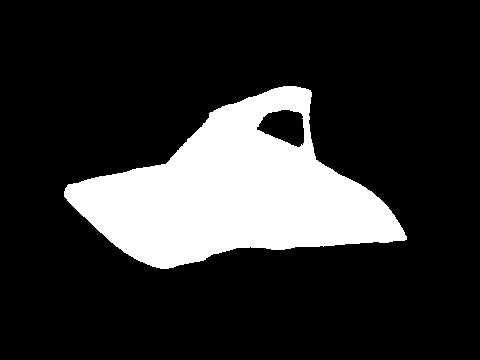}} 
            \vspace{-0.5mm}
		\\

            \rotatebox[origin=c]{90}{\small (d)}           
            &
		\makecell[c]{\includegraphics[width=0.08\linewidth,height=0.07\linewidth]{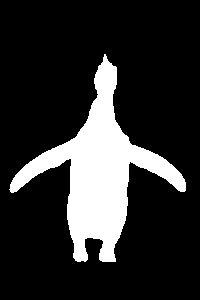}} 
		&
		\makecell[c]{\includegraphics[width=0.08\linewidth,height=0.07\linewidth]{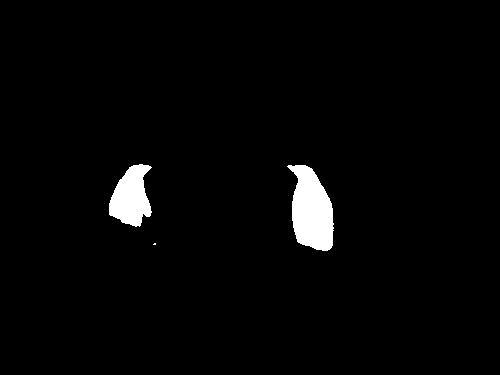}} 
		&
		\makecell[c]{\includegraphics[width=0.08\linewidth,height=0.07\linewidth]{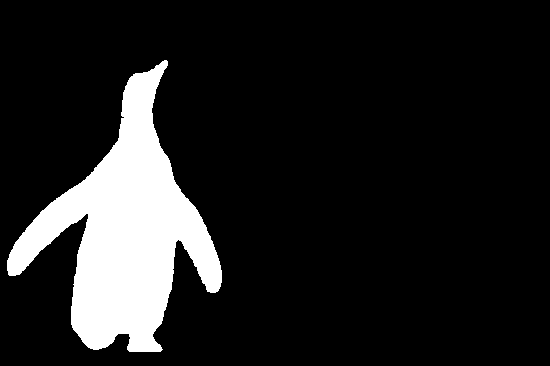}} 
		&
		\makecell[c]{\includegraphics[width=0.08\linewidth,height=0.07\linewidth]{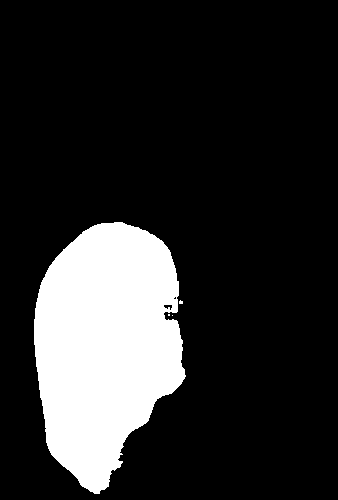}}        
		&
		\makecell[c]{\includegraphics[width=0.08\linewidth,height=0.07\linewidth]{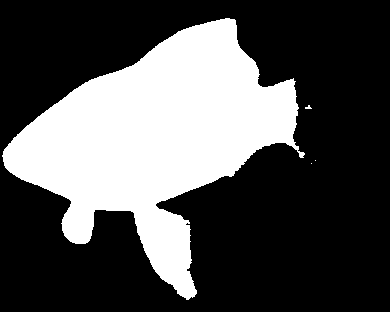}} 
		&
		\makecell[c]{\includegraphics[width=0.08\linewidth,height=0.07\linewidth]{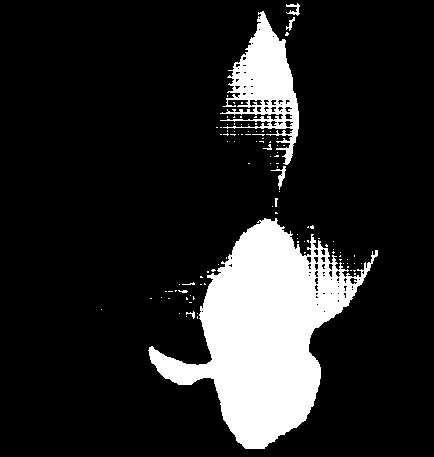}} 
		&
		\makecell[c]{\includegraphics[width=0.08\linewidth,height=0.07\linewidth]{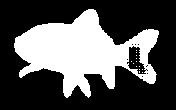}} 
		&
		\makecell[c]{\includegraphics[width=0.08\linewidth,height=0.07\linewidth]{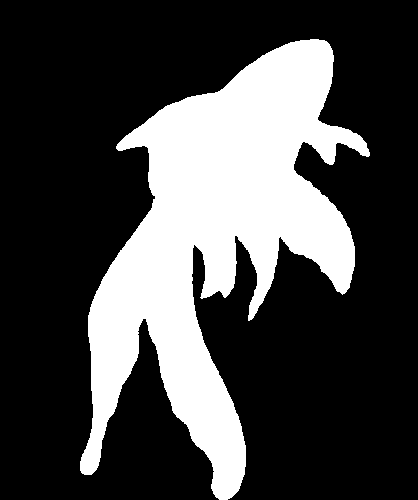}}
        &
		\makecell[c]{\includegraphics[width=0.08\linewidth,height=0.07\linewidth]{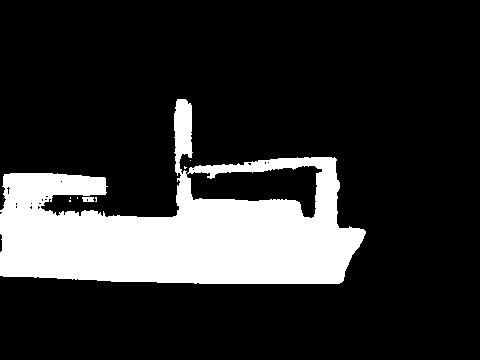}} 
		&
		\makecell[c]{\includegraphics[width=0.08\linewidth,height=0.07\linewidth]{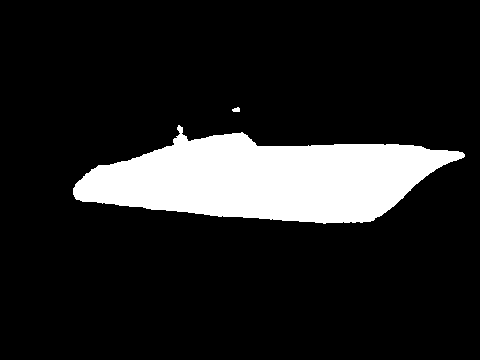}} 
		&
		\makecell[c]{\includegraphics[width=0.08\linewidth,height=0.07\linewidth]{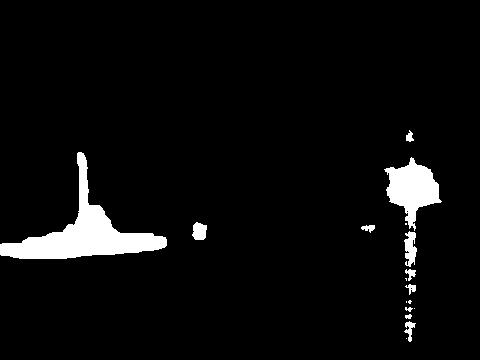}}
        &
		\makecell[c]{\includegraphics[width=0.08\linewidth,height=0.07\linewidth]{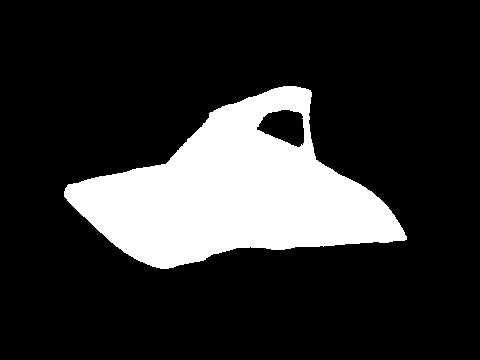}}
            \vspace{-0.5mm}
		\\

             \rotatebox[origin=c]{90}{\small (e)}
            &
		\makecell[c]{\includegraphics[width=0.08\linewidth,height=0.07\linewidth]{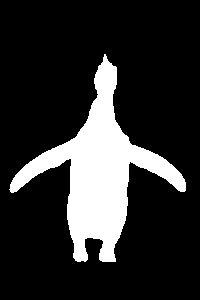}} 
		&
		\makecell[c]{\includegraphics[width=0.08\linewidth,height=0.07\linewidth]{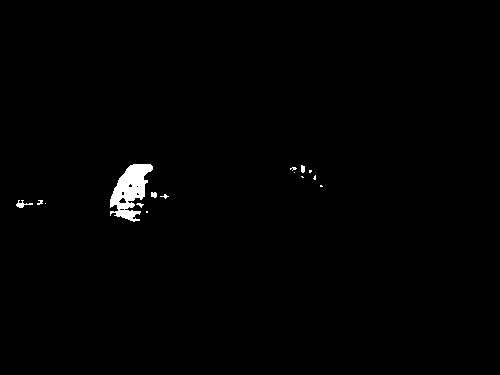}} 
		&
		\makecell[c]{\includegraphics[width=0.08\linewidth,height=0.07\linewidth]{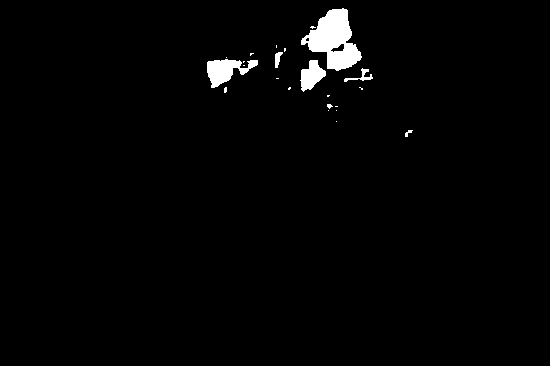}} 
		&
		\makecell[c]{\includegraphics[width=0.08\linewidth,height=0.07\linewidth]{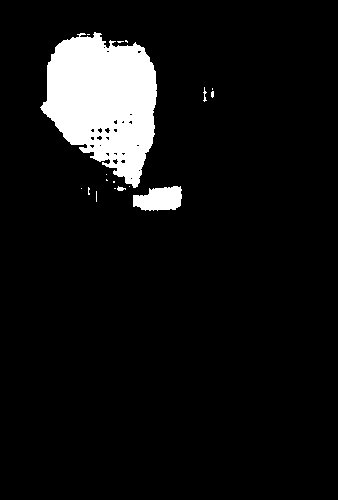}}        
		&
		\makecell[c]{\includegraphics[width=0.08\linewidth,height=0.07\linewidth]{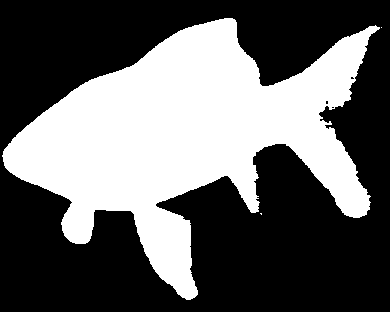}} 
		&
		\makecell[c]{\includegraphics[width=0.08\linewidth,height=0.07\linewidth]{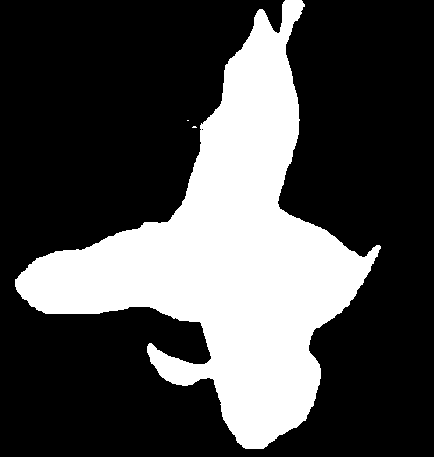}} 
		&
		\makecell[c]{\includegraphics[width=0.08\linewidth,height=0.07\linewidth]{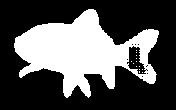}} 
		&
		\makecell[c]{\includegraphics[width=0.08\linewidth,height=0.07\linewidth]{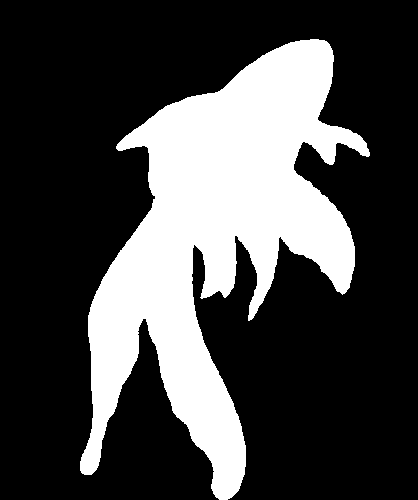}}
        &
		\makecell[c]{\includegraphics[width=0.08\linewidth,height=0.07\linewidth]{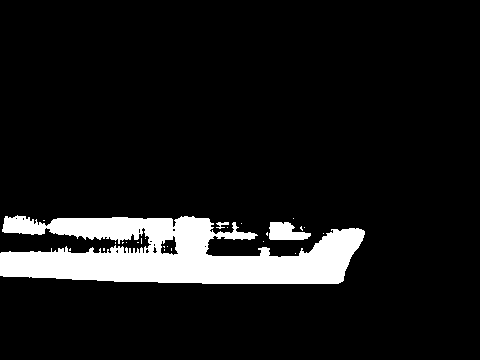}} 
		&
		\makecell[c]{\includegraphics[width=0.08\linewidth,height=0.07\linewidth]{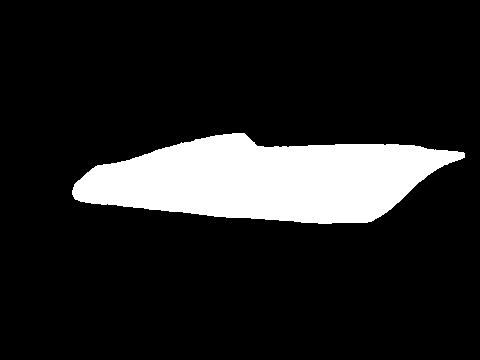}} 
		&
		\makecell[c]{\includegraphics[width=0.08\linewidth,height=0.07\linewidth]{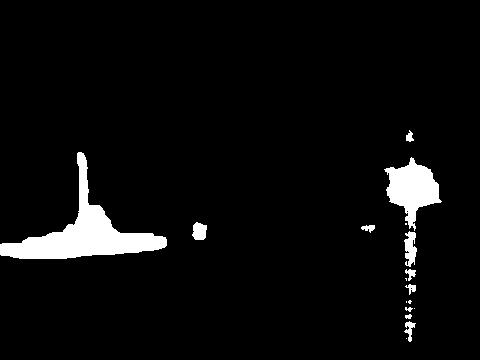}}
        &
		\makecell[c]{\includegraphics[width=0.08\linewidth,height=0.07\linewidth]{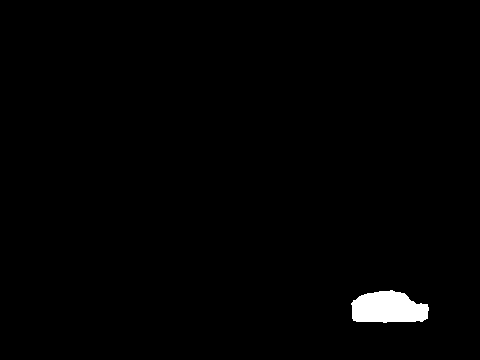}}
            \vspace{-0.5mm}
		\\

            \rotatebox[origin=c]{90}{\small (f)}
            &
		\makecell[c]{\includegraphics[width=0.08\linewidth,height=0.07\linewidth]{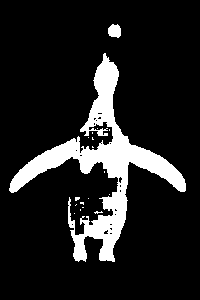}} 
		&
		\makecell[c]{\includegraphics[width=0.08\linewidth,height=0.07\linewidth]{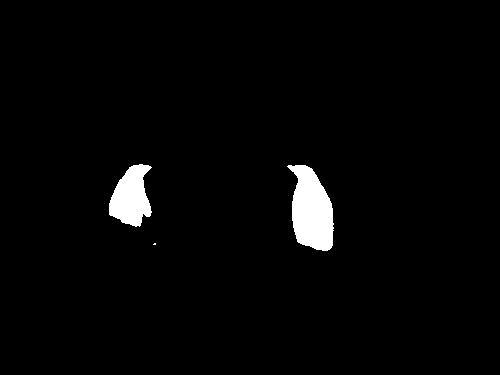}} 
		&
		\makecell[c]{\includegraphics[width=0.08\linewidth,height=0.07\linewidth]{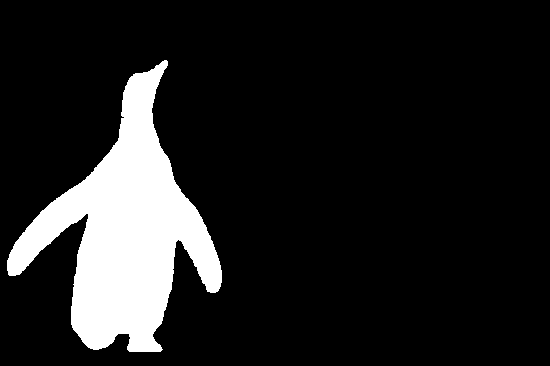}} 
		&
		\makecell[c]{\includegraphics[width=0.08\linewidth,height=0.07\linewidth]{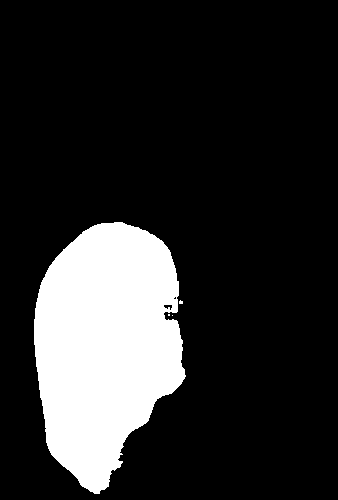}}        
		&
		\makecell[c]{\includegraphics[width=0.08\linewidth,height=0.07\linewidth]{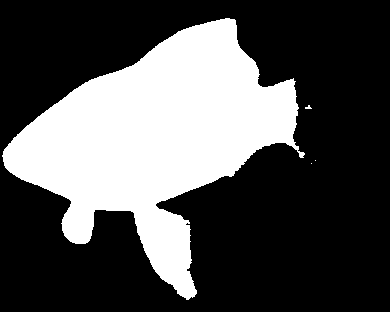}} 
		&
		\makecell[c]{\includegraphics[width=0.08\linewidth,height=0.07\linewidth]{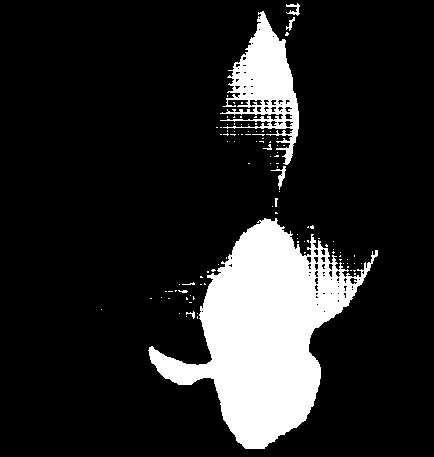}} 
		&
		\makecell[c]{\includegraphics[width=0.08\linewidth,height=0.07\linewidth]{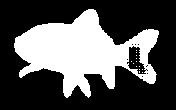}} 
		&
		\makecell[c]{\includegraphics[width=0.08\linewidth,height=0.07\linewidth]{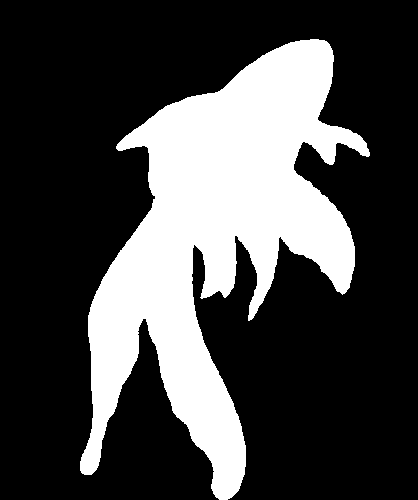}}
        &
		\makecell[c]{\includegraphics[width=0.08\linewidth,height=0.07\linewidth]{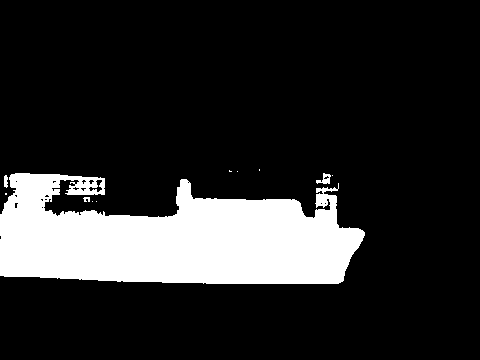}} 
		&
		\makecell[c]{\includegraphics[width=0.08\linewidth,height=0.07\linewidth]{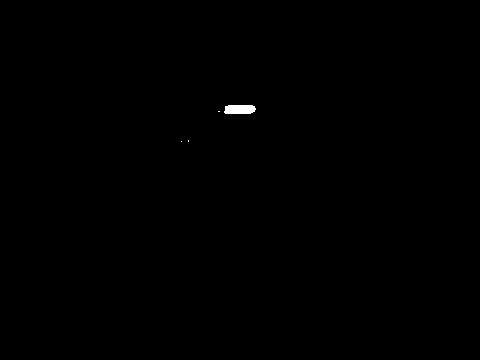}} 
		&
		\makecell[c]{\includegraphics[width=0.08\linewidth,height=0.07\linewidth]{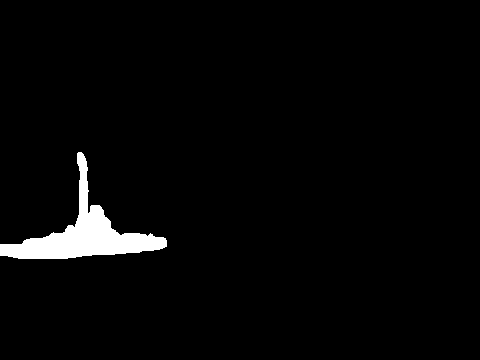}}
        &
		\makecell[c]{\includegraphics[width=0.08\linewidth,height=0.07\linewidth]{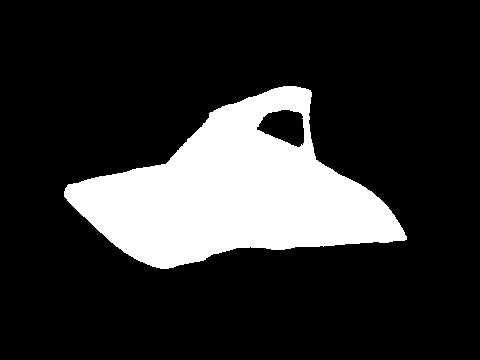}}
            \vspace{-0.5mm}
		\\

            \rotatebox[origin=c]{90}{\small (g)}
            &
		\makecell[c]{\includegraphics[width=0.08\linewidth,height=0.07\linewidth]{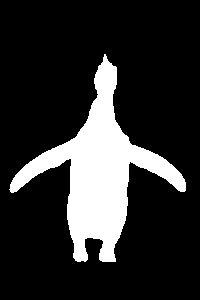}} 
		&
		\makecell[c]{\includegraphics[width=0.08\linewidth,height=0.07\linewidth]{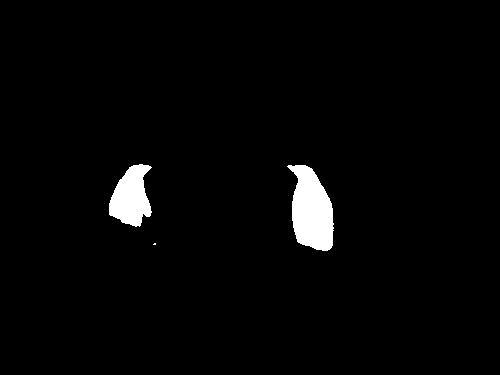}} 
		&
		\makecell[c]{\includegraphics[width=0.08\linewidth,height=0.07\linewidth]{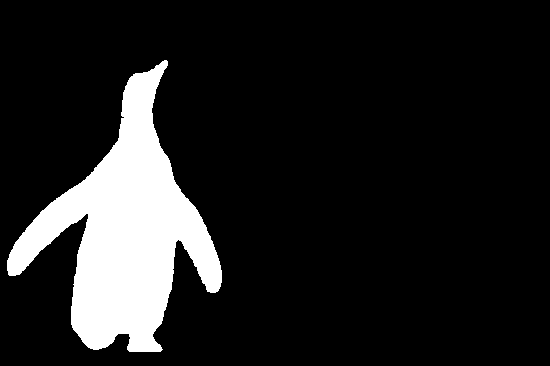}} 
		&
		\makecell[c]{\includegraphics[width=0.08\linewidth,height=0.07\linewidth]{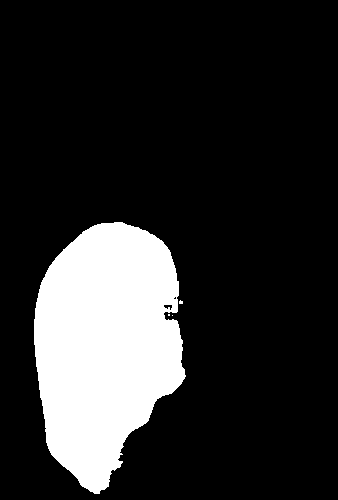}}        
		&
		\makecell[c]{\includegraphics[width=0.08\linewidth,height=0.07\linewidth]{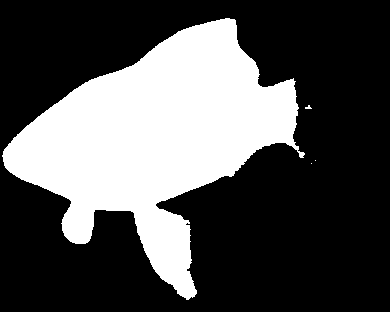}} 
		&
		\makecell[c]{\includegraphics[width=0.08\linewidth,height=0.07\linewidth]{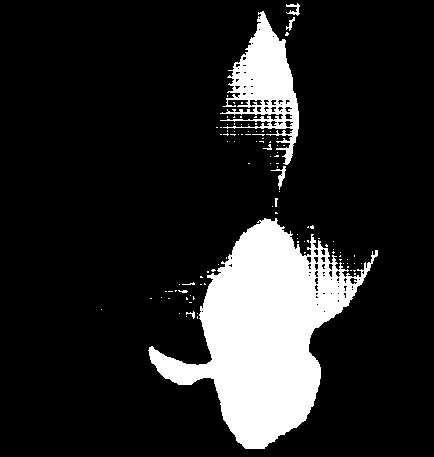}} 
		&
		\makecell[c]{\includegraphics[width=0.08\linewidth,height=0.07\linewidth]{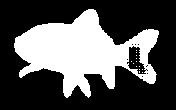}} 
		&
		\makecell[c]{\includegraphics[width=0.08\linewidth,height=0.07\linewidth]{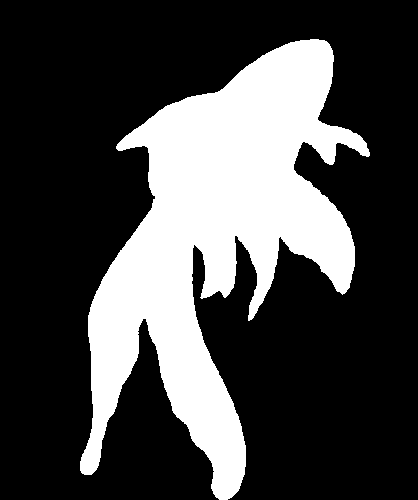}}
        &
		\makecell[c]{\includegraphics[width=0.08\linewidth,height=0.07\linewidth]{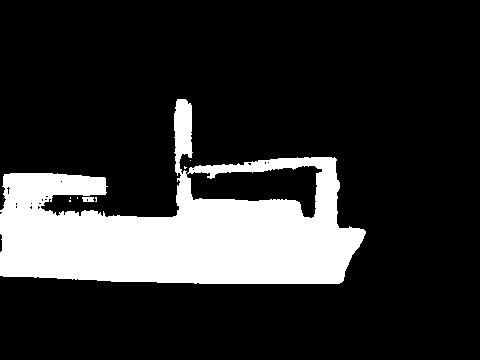}} 
		&
		\makecell[c]{\includegraphics[width=0.08\linewidth,height=0.07\linewidth]{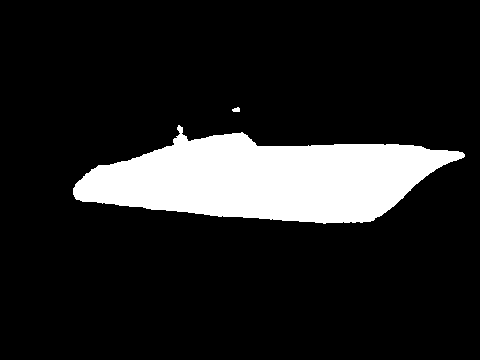}} 
		&
		\makecell[c]{\includegraphics[width=0.08\linewidth,height=0.07\linewidth]{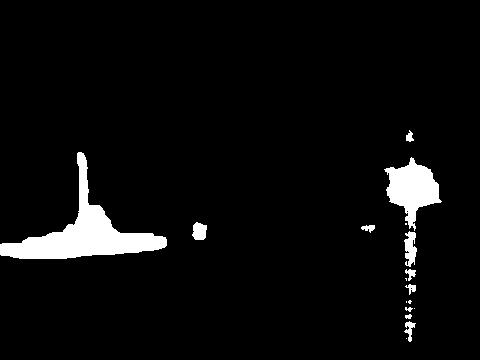}}
        &
		\makecell[c]{\includegraphics[width=0.08\linewidth,height=0.07\linewidth]{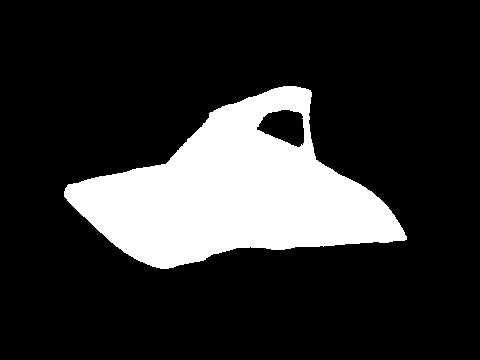}}
            \vspace{-0.5mm}
		\\

             \rotatebox[origin=c]{90}{\small (h)}
            &
		\makecell[c]{\includegraphics[width=0.08\linewidth,height=0.07\linewidth]{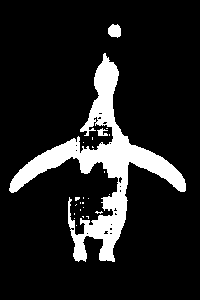}} 
		&
		\makecell[c]{\includegraphics[width=0.08\linewidth,height=0.07\linewidth]{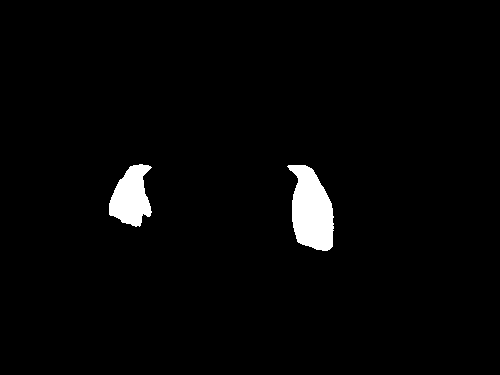}} 
		&
		\makecell[c]{\includegraphics[width=0.08\linewidth,height=0.07\linewidth]{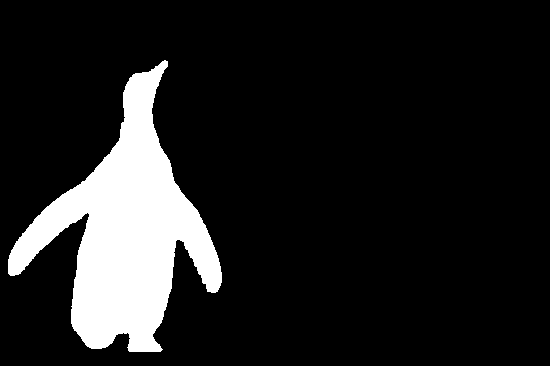}} 
		&
		\makecell[c]{\includegraphics[width=0.08\linewidth,height=0.07\linewidth]{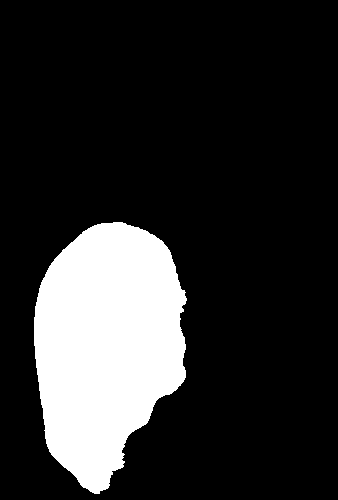}}        
		&
		\makecell[c]{\includegraphics[width=0.08\linewidth,height=0.07\linewidth]{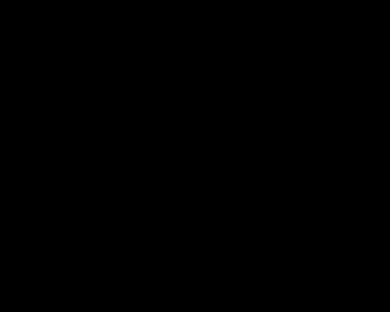}} 
		&
		\makecell[c]{\includegraphics[width=0.08\linewidth,height=0.07\linewidth]{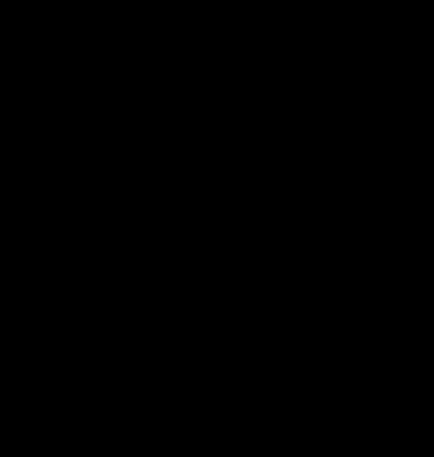}} 
		&
		\makecell[c]{\includegraphics[width=0.08\linewidth,height=0.07\linewidth]{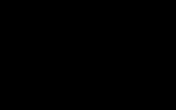}} 
		&
		\makecell[c]{\includegraphics[width=0.08\linewidth,height=0.07\linewidth]{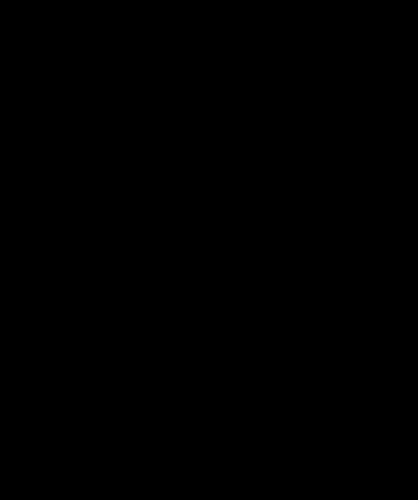}}
        &
		\makecell[c]{\includegraphics[width=0.08\linewidth,height=0.07\linewidth]{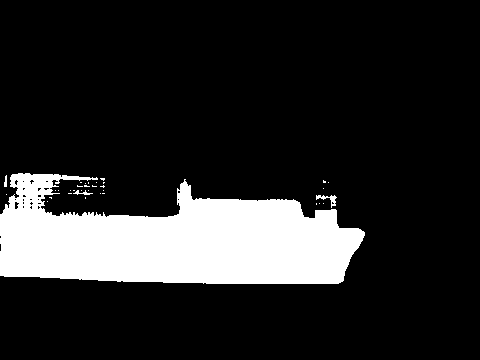}} 
		&
		\makecell[c]{\includegraphics[width=0.08\linewidth,height=0.07\linewidth]{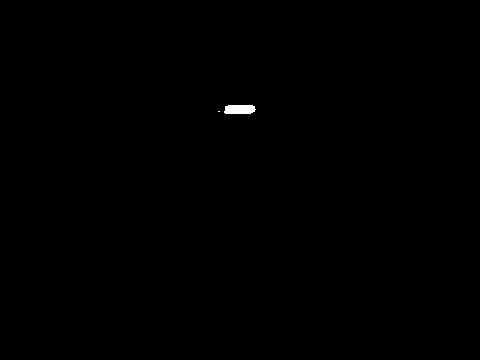}} 
		&
		\makecell[c]{\includegraphics[width=0.08\linewidth,height=0.07\linewidth]{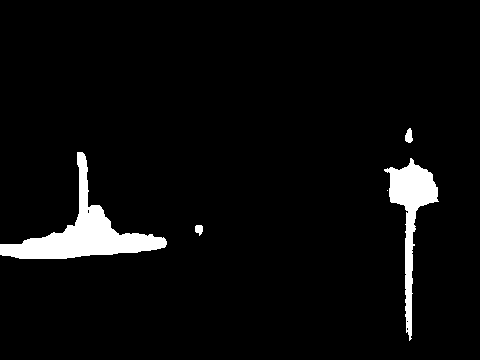}}
        &
		\makecell[c]{\includegraphics[width=0.08\linewidth,height=0.07\linewidth]{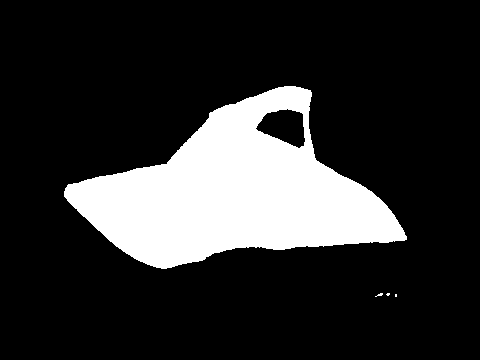}}
            \vspace{-0.5mm}
		\\

            \rotatebox[origin=c]{90}{\small (i)}
            &
		\makecell[c]{\includegraphics[width=0.08\linewidth,height=0.07\linewidth]{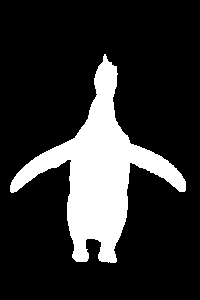}} 
		&
		\makecell[c]{\includegraphics[width=0.08\linewidth,height=0.07\linewidth]{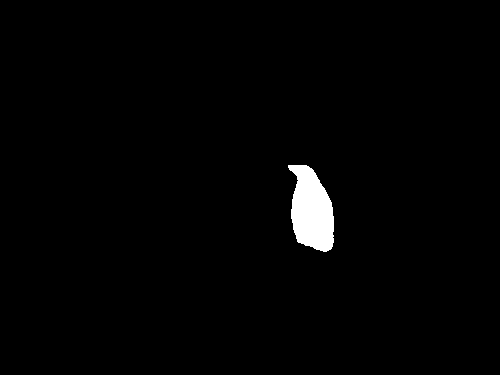}} 
		&
		\makecell[c]{\includegraphics[width=0.08\linewidth,height=0.07\linewidth]{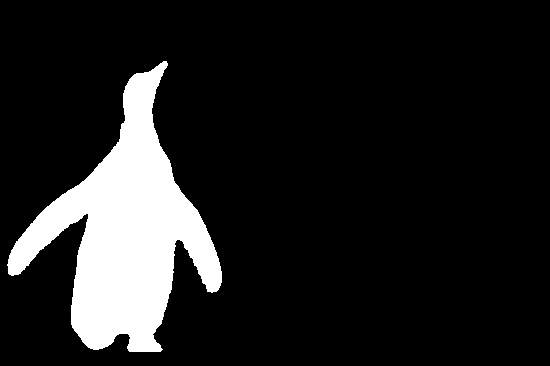}} 
		&
		\makecell[c]{\includegraphics[width=0.08\linewidth,height=0.07\linewidth]{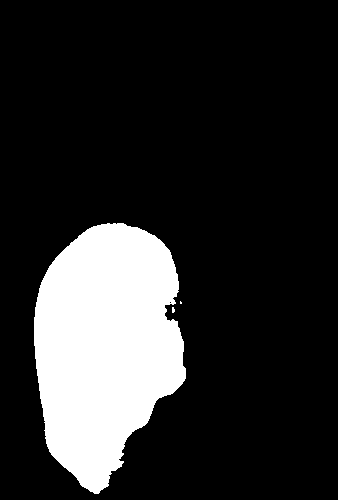}}        
		&
		\makecell[c]{\includegraphics[width=0.08\linewidth,height=0.07\linewidth]{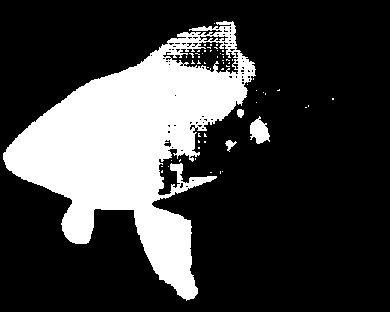}} 
		&
		\makecell[c]{\includegraphics[width=0.08\linewidth,height=0.07\linewidth]{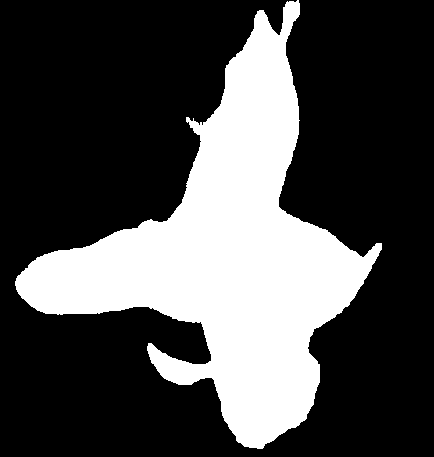}} 
		&
		\makecell[c]{\includegraphics[width=0.08\linewidth,height=0.07\linewidth]{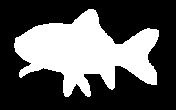}} 
		&
		\makecell[c]{\includegraphics[width=0.08\linewidth,height=0.07\linewidth]{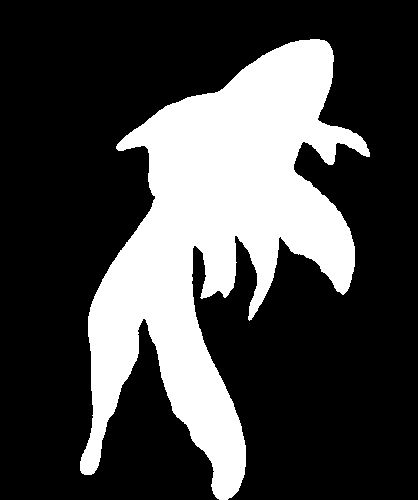}}
        &
		\makecell[c]{\includegraphics[width=0.08\linewidth,height=0.07\linewidth]{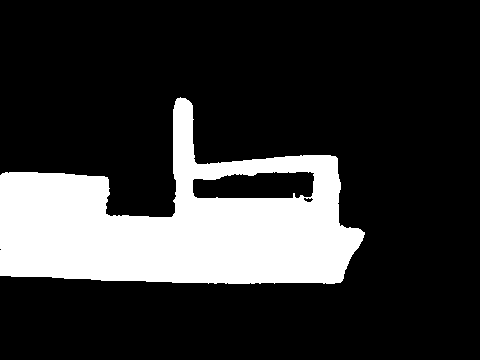}} 
		&
		\makecell[c]{\includegraphics[width=0.08\linewidth,height=0.07\linewidth]{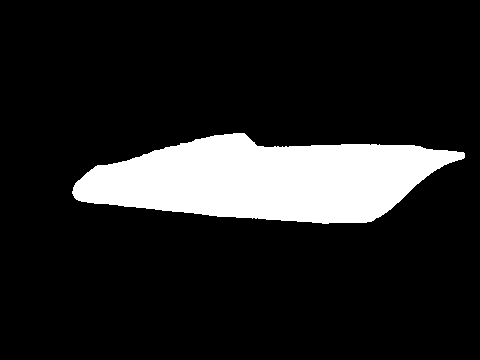}} 
		&
		\makecell[c]{\includegraphics[width=0.08\linewidth,height=0.07\linewidth]{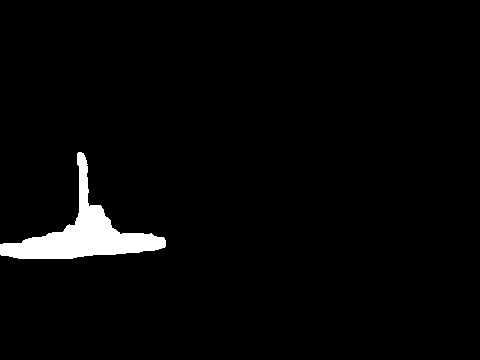}}
        &
		\makecell[c]{\includegraphics[width=0.08\linewidth,height=0.07\linewidth]{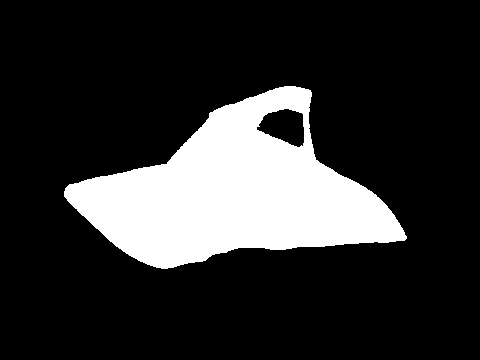}}
            \vspace{-0.5mm}
		\\

            \rotatebox[origin=c]{90}{\small (j)}
            &
		\makecell[c]{\includegraphics[width=0.08\linewidth,height=0.07\linewidth]{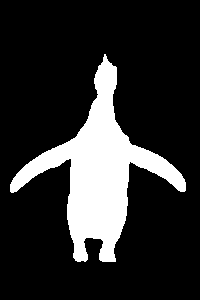}} 
		&
		\makecell[c]{\includegraphics[width=0.08\linewidth,height=0.07\linewidth]{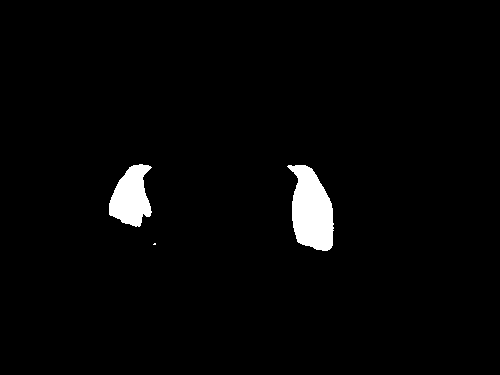}} 
		&
		\makecell[c]{\includegraphics[width=0.08\linewidth,height=0.07\linewidth]{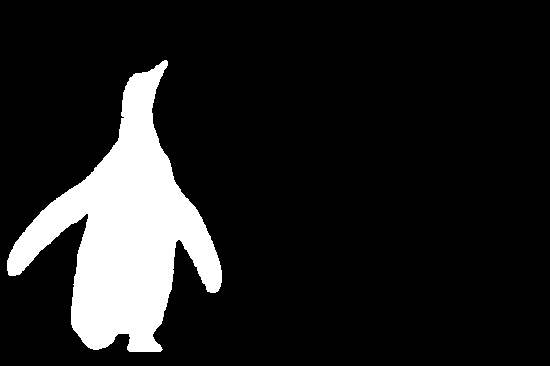}} 
		&
		\makecell[c]{\includegraphics[width=0.08\linewidth,height=0.07\linewidth]{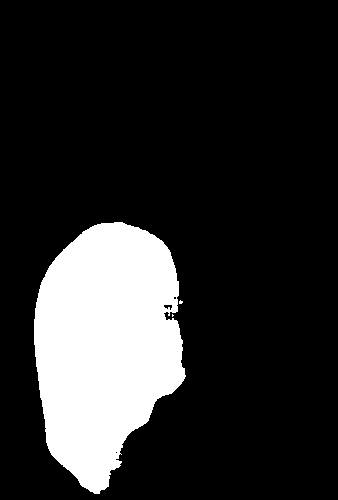}}        
		&
		\makecell[c]{\includegraphics[width=0.08\linewidth,height=0.07\linewidth]{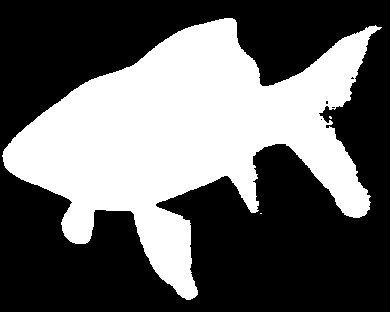}} 
		&
		\makecell[c]{\includegraphics[width=0.08\linewidth,height=0.07\linewidth]{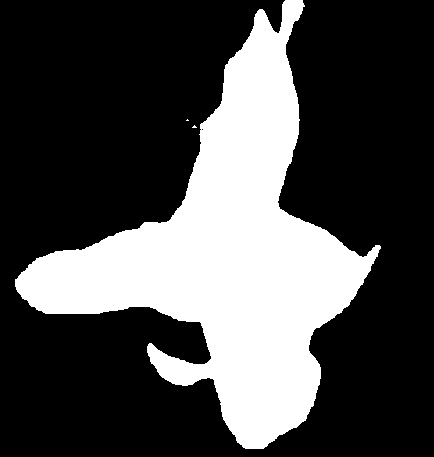}} 
		&
		\makecell[c]{\includegraphics[width=0.08\linewidth,height=0.07\linewidth]{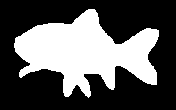}} 
		&
		\makecell[c]{\includegraphics[width=0.08\linewidth,height=0.07\linewidth]{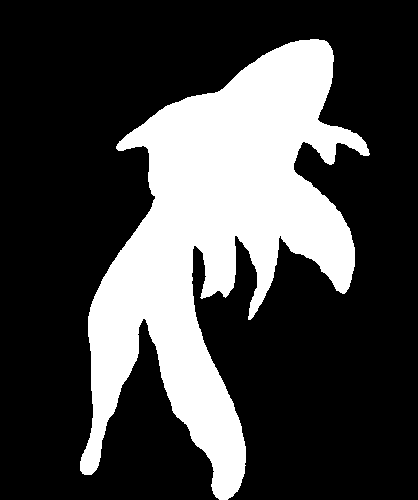}}
        &
		\makecell[c]{\includegraphics[width=0.08\linewidth,height=0.07\linewidth]{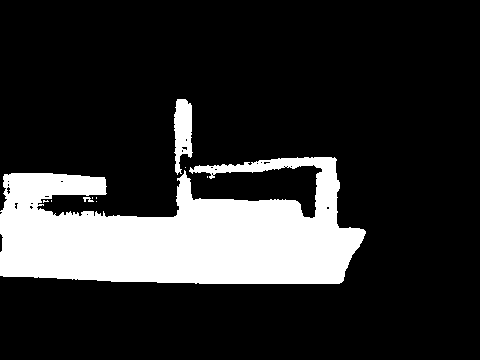}} 
		&
		\makecell[c]{\includegraphics[width=0.08\linewidth,height=0.07\linewidth]{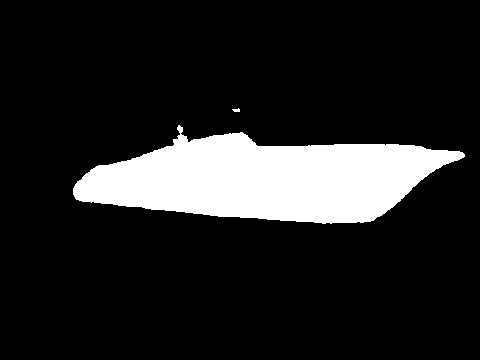}} 
		&
		\makecell[c]{\includegraphics[width=0.08\linewidth,height=0.07\linewidth]{Figures/SOTA_Maps/close-source/boat/boat_012_llava-onevision.png}}
        &
		\makecell[c]{\includegraphics[width=0.08\linewidth,height=0.07\linewidth]{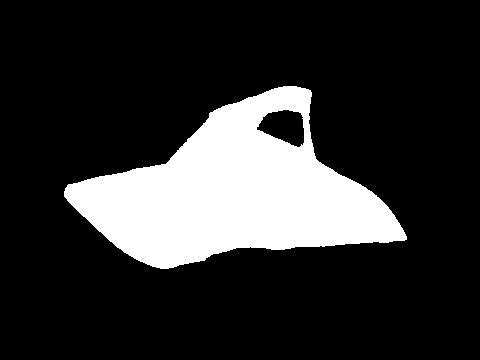}}
            \vspace{-0.5mm}
		\\

		\end{tabular}
            \vspace{-2mm}
    \caption{\textbf{Qualitative comparison with other MLLMs on CoSOD.} \textbf{(a)} denotes the input image group, \textbf{(b)} the ground-truth co-salient masks, \textbf{(c)} our Saliency-R1, followed by: \textbf{(d)} Claude-Sonnet-4.5, \textbf{(e)} Gemini-2.5-Pro, \textbf{(f)} GPT-4o, \textbf{(g)} Grok-4, \textbf{(h)} Qwen3-VL-32B-Thinking, \textbf{(i)} Qwen2.5-VL-7B-Instruct, and \textbf{(j)} LLaVA-OneVision-Qwen2-7B.}
    \label{MLLM_CoSOD}
\end{minipage}
\end{figure*}

Compared to closed- and open-source MLLMs (\eg, GPT-4o~\cite{hurst2024gpt}, Gemini-2.5-Pro~\cite{comanici2025gemini}, Qwen3-VL-32B-Thinking~\cite{yang2025qwen3}, LLaVA-OneVision-7B~\cite{li2024llava}, and so on), Saliency-R1 exhibits markedly superior visual grounding precision and segmentation coherence, as shown in Figure~\ref{MLLM_SOD} (SOD), Figure~\ref{MLLM_SIS} (SIS) and Figure~\ref{MLLM_CoSOD} (CoSOD).

In SOD, as shown in Figure~\ref{MLLM_SOD}, common failure modes include false-positive activation on background regions, \eg, the substrate beneath the chameleon (Column~1), the branches behind the bird (Column~2), the flowers surrounding the black dog (Column~3), and the water surface near the deer's legs (Column~4). Our method preserves fine structure and enforces foreground--background contrast via explicit CoT reasoning.

In SIS, as shown in Figure~\ref{MLLM_SIS}, common failure modes of MLLMs include under-prediction of instances (\eg, the buckets in Column~1, the car in Column~3, and the person in Column~4), over-prediction of non-instances (\eg, spurious masks on the ground in Columns~1 and~3, and on the person in Column~2). Our method, guided by explicit CoT reasoning that generates distinct \texttt{<ins>}-tagged expressions (\eg, ``a person standing in front of the bus''), enables EVF-SAM to robustly isolate each instance.

In the CoSOD task, as shown in Figure~\ref{MLLM_CoSOD}, MLLMs often hallucinate or overlook small co-salient instances under scale variation (\eg, distant boats in Column 11), while our model reliably identifies them. Additionally, in cluttered scenes (\eg, penguin and band in Column 3), Gemini-2.5-Pro segments background distractors, while our CoT-induced consensus reasoning (Step 2: Co-salience Deduction) suppresses non-common elements, yielding masks that are strictly aligned with ground truth.

These results highlight a key advantage: our CoT is not just descriptive, but functionally coupled to segmentation, each referring expression acts as a discriminative prompt for the segmenter.

\subsection{Comparison with Task-Specific Methods}

\begin{figure*}[t]
\begin{minipage}[c]{1.\textwidth}
	\scriptsize
	\renewcommand{\tabcolsep}{0.5pt} 
	\renewcommand{\arraystretch}{0.9} 
	\centering
        \begin{tabular}{cccccccccccccccc}
	    &
            \multicolumn{4}{c}{\cellcolor[RGB]
            {255,230,153}$\bold{backpack}$}
            &
            \multicolumn{4}{c}{\cellcolor[RGB]{244,177,131}$\bold{camera}$}
	    &
            \multicolumn{4}{c}{\cellcolor[RGB]{169,209,142}$\bold{Yellow duck}$}
            \vspace{0.5mm}
	    \\

            \arrayrulecolor{red}
            \rotatebox[origin=c]{90}{\small (a)}
            &
		\makecell[c]{\includegraphics[width=0.08\linewidth,height=0.06\linewidth]{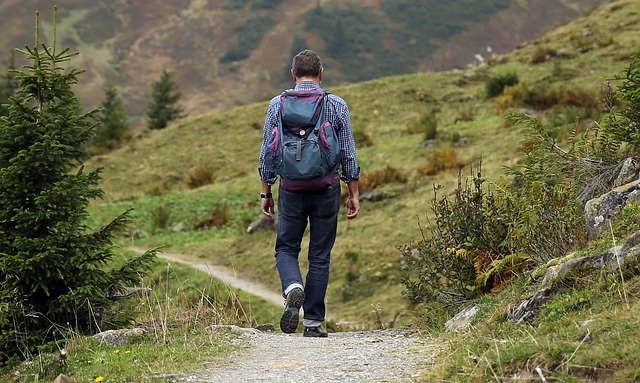}}
        &
		\makecell[c]{\includegraphics[width=0.08\linewidth,height=0.06\linewidth]{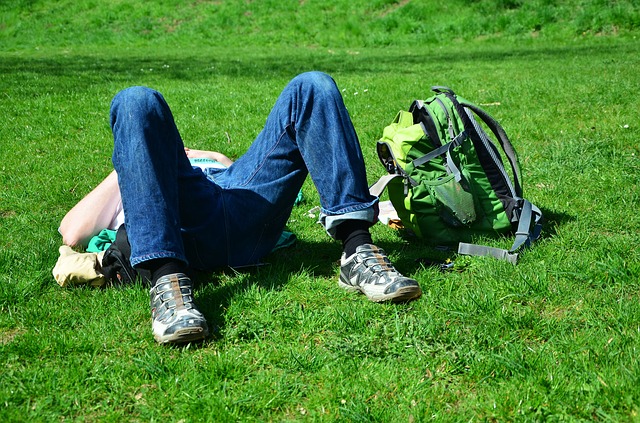}} 
		&
		\makecell[c]{\includegraphics[width=0.08\linewidth,height=0.06\linewidth]{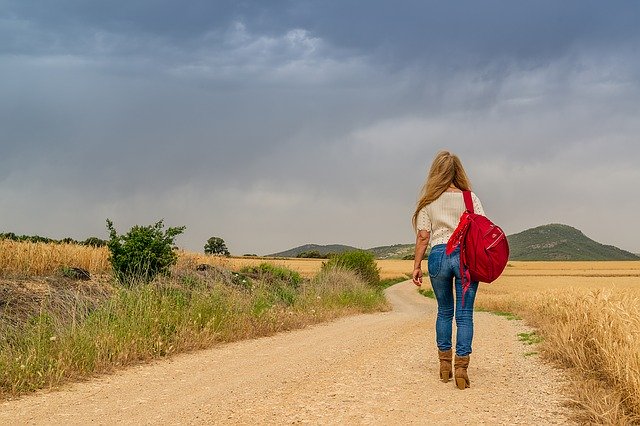}} 
		&
		\makecell[c]{\includegraphics[width=0.08\linewidth,height=0.06\linewidth]{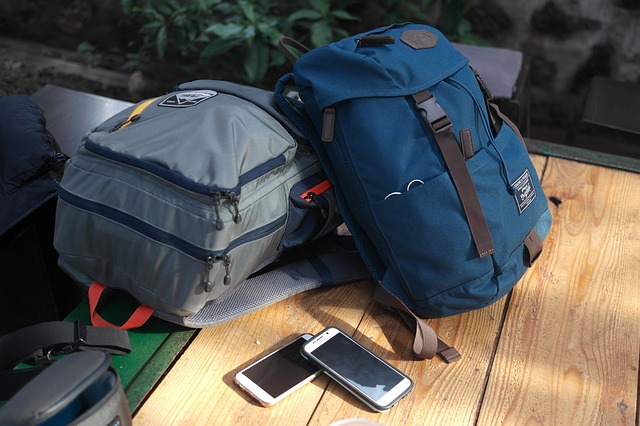}}
        &
		\makecell[c]{\includegraphics[width=0.08\linewidth,height=0.06\linewidth]{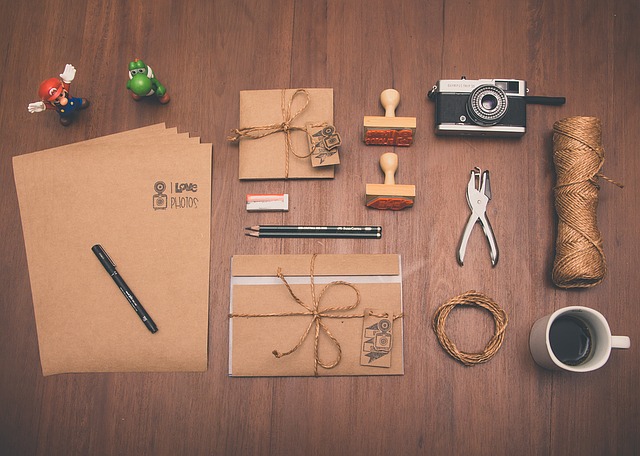}} 
		&
		\makecell[c]{\includegraphics[width=0.08\linewidth,height=0.06\linewidth]{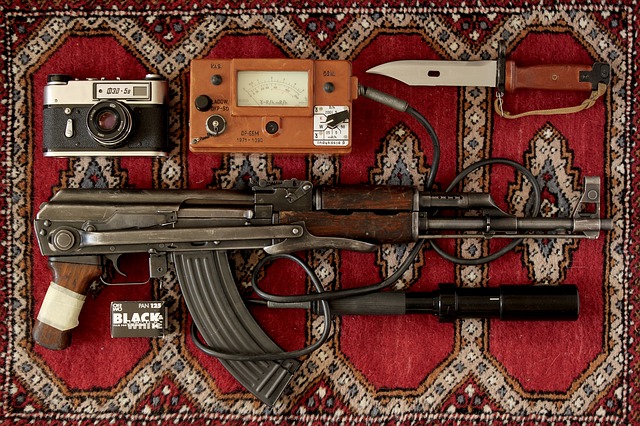}} 
		&
		\makecell[c]{\includegraphics[width=0.08\linewidth,height=0.06\linewidth]{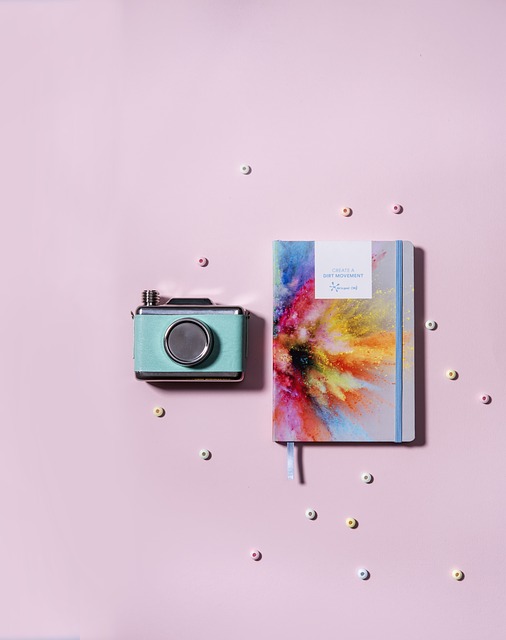}} 
		&
		\makecell[c]{\includegraphics[width=0.08\linewidth,height=0.06\linewidth]{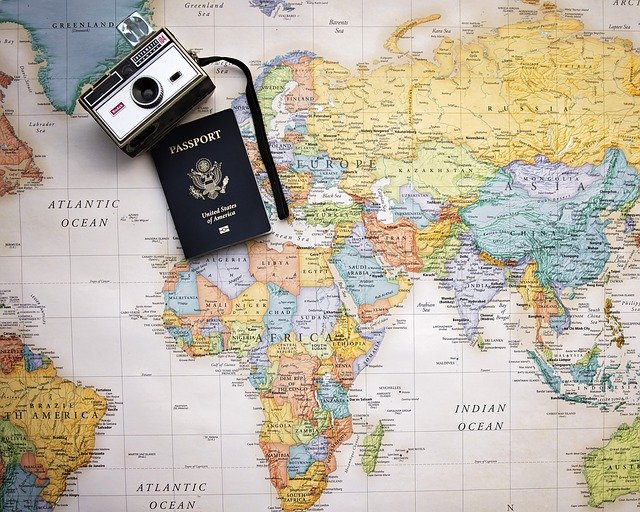}}         
		&
		\makecell[c]{\includegraphics[width=0.08\linewidth,height=0.06\linewidth]{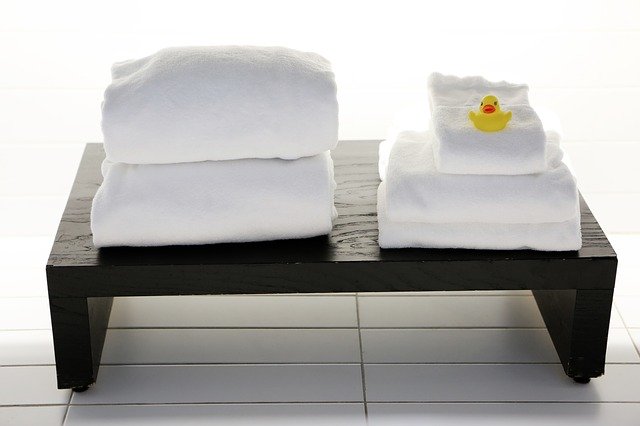}} 
		&
		\makecell[c]{\includegraphics[width=0.08\linewidth,height=0.06\linewidth]{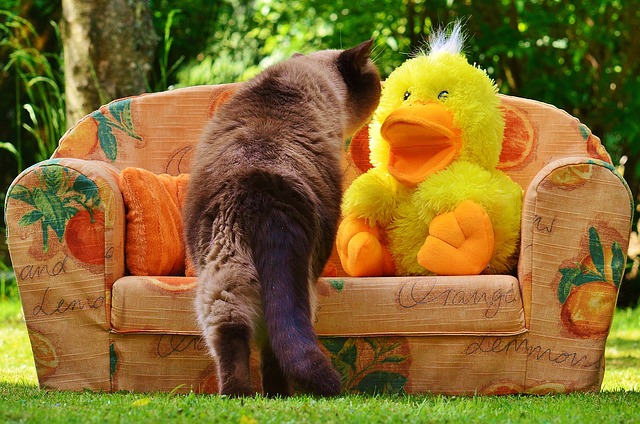}}
            &
		\makecell[c]{\includegraphics[width=0.08\linewidth,height=0.06\linewidth]{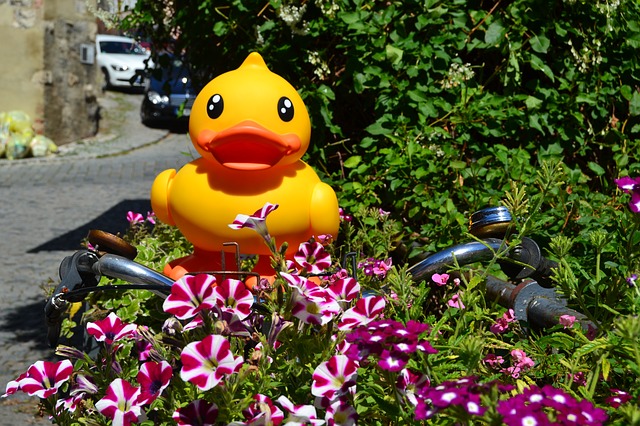}} 
		&
		\makecell[c]{\includegraphics[width=0.08\linewidth,height=0.06\linewidth]{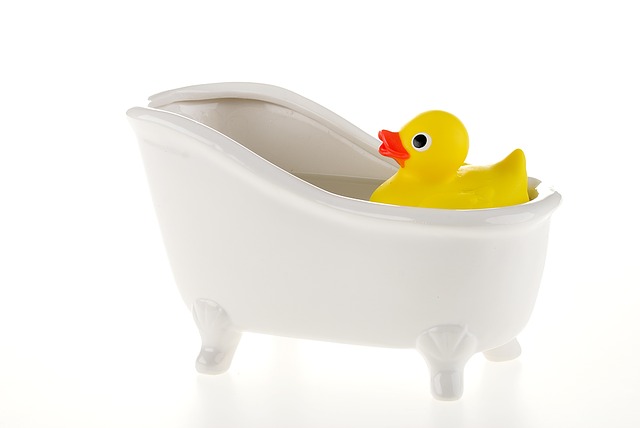}} 
 
            \vspace{-0.5mm}
            \\

            \rotatebox[origin=c]{90}{\small (b)}
            &
		\makecell[c]{\includegraphics[width=0.08\linewidth,height=0.06\linewidth]{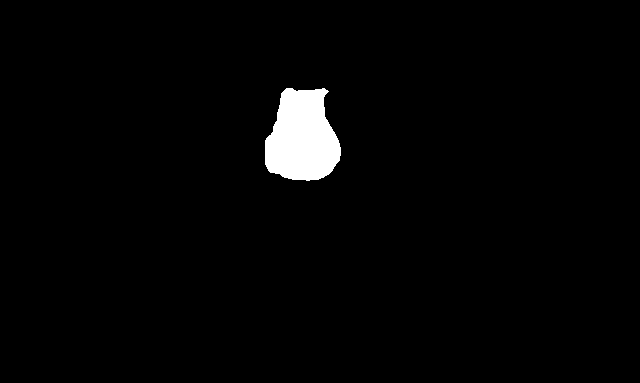}} 
		&
		\makecell[c]{\includegraphics[width=0.08\linewidth,height=0.06\linewidth]{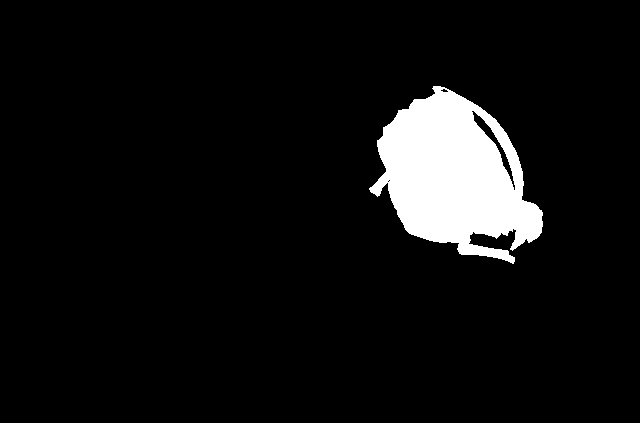}} 
		&
		\makecell[c]{\includegraphics[width=0.08\linewidth,height=0.06\linewidth]{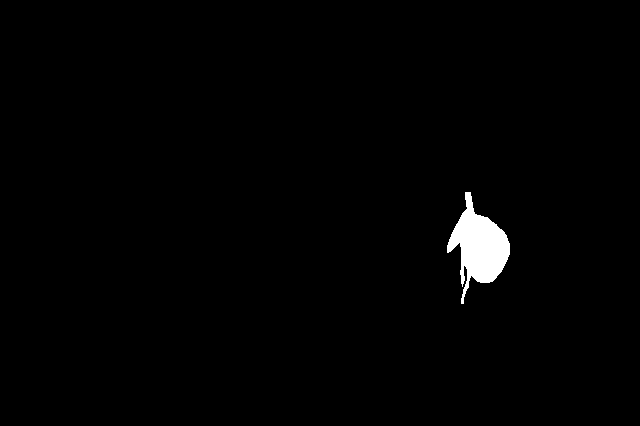}} 
		&
		\makecell[c]{\includegraphics[width=0.08\linewidth,height=0.06\linewidth]{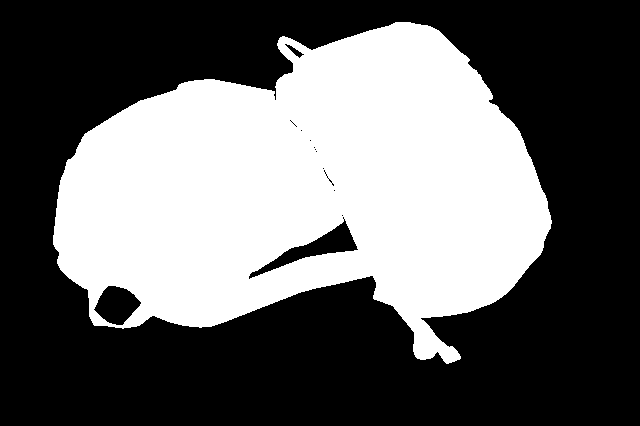}} 
		&
		\makecell[c]{\includegraphics[width=0.08\linewidth,height=0.06\linewidth]{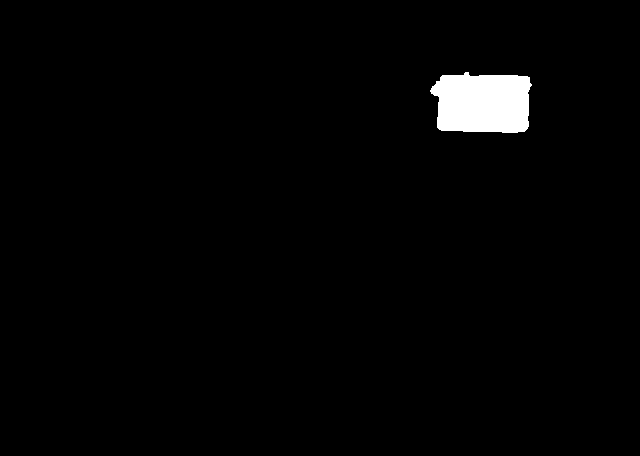}} 
		&
		\makecell[c]{\includegraphics[width=0.08\linewidth,height=0.06\linewidth]{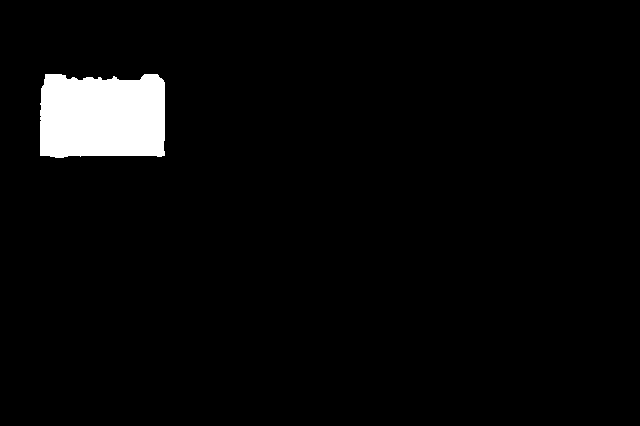}} 
		&
		\makecell[c]{\includegraphics[width=0.08\linewidth,height=0.06\linewidth]{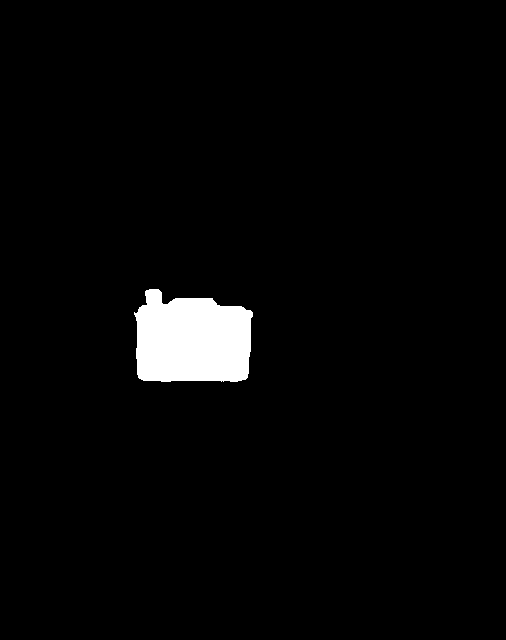}} 
		&
		\makecell[c]{\includegraphics[width=0.08\linewidth,height=0.06\linewidth]{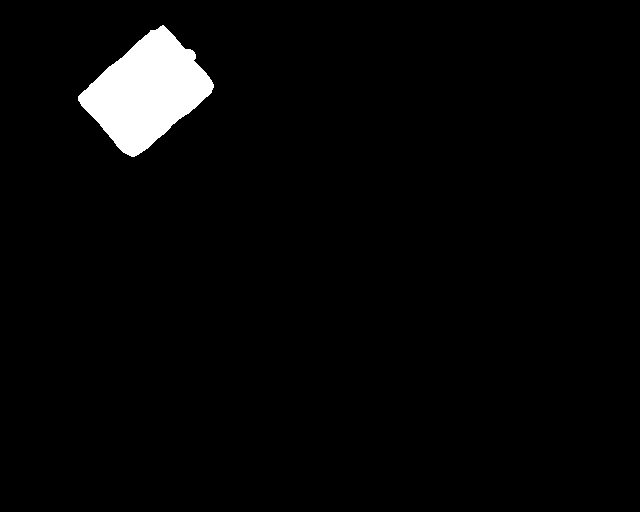}}
        &
		\makecell[c]{\includegraphics[width=0.08\linewidth,height=0.06\linewidth]{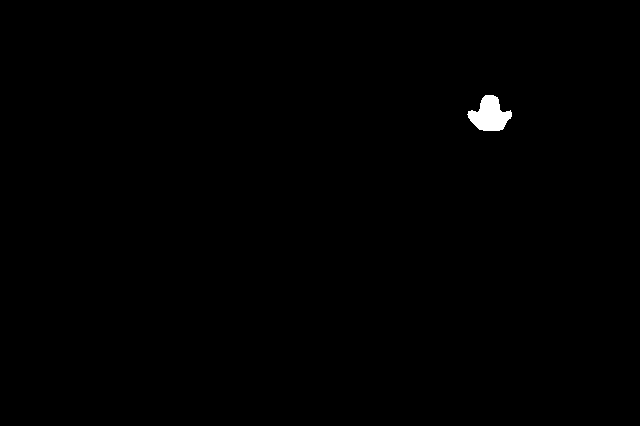}} 
		&
        \makecell[c]{\includegraphics[width=0.08\linewidth,height=0.06\linewidth]{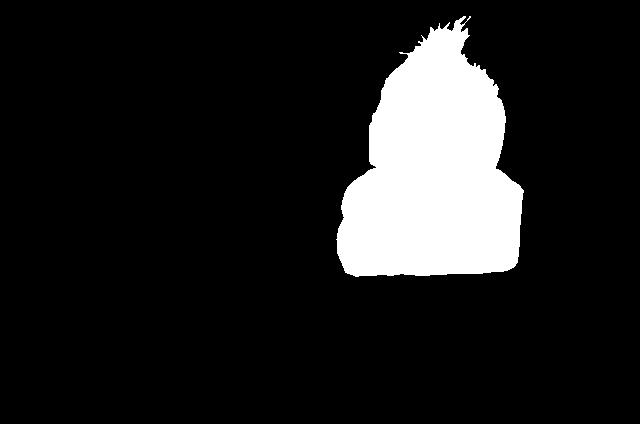}} 
		&
		\makecell[c]{\includegraphics[width=0.08\linewidth,height=0.06\linewidth]{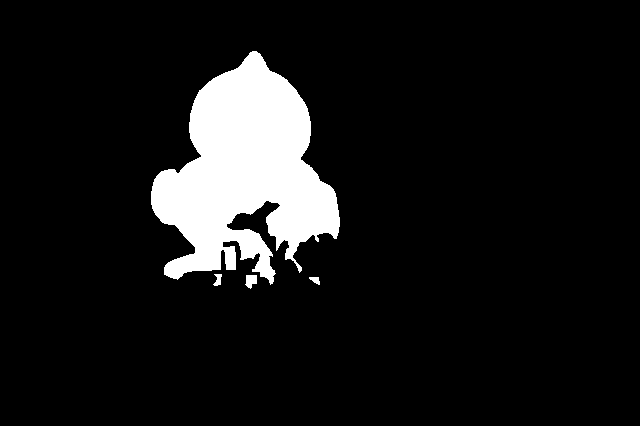}}
            &
		\makecell[c]{\includegraphics[width=0.08\linewidth,height=0.06\linewidth]{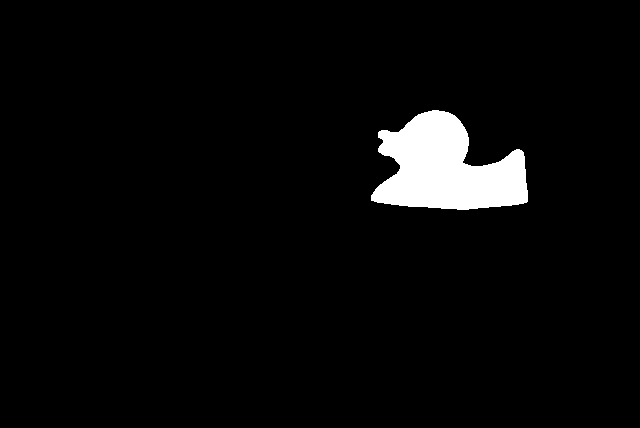}} 
            \vspace{-0.5mm}
		\\

            \rotatebox[origin=c]{90}{\small (c)}  
        &
		\makecell[c]{\includegraphics[width=0.08\linewidth,height=0.06\linewidth]{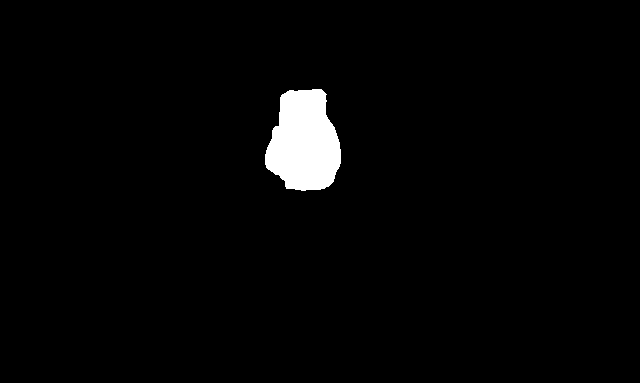}} 
		&
		\makecell[c]{\includegraphics[width=0.08\linewidth,height=0.06\linewidth]{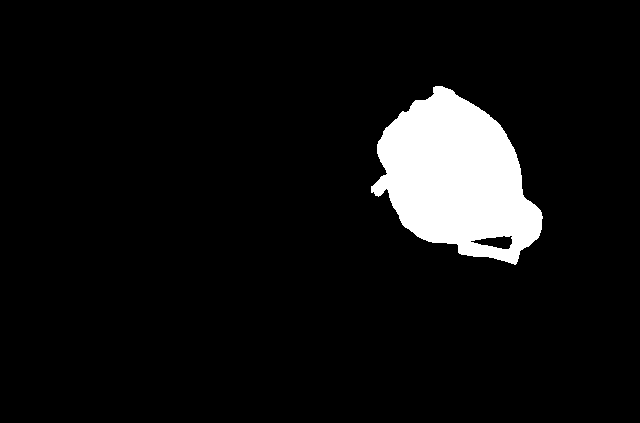}} 
		&
		\makecell[c]{\includegraphics[width=0.08\linewidth,height=0.06\linewidth]{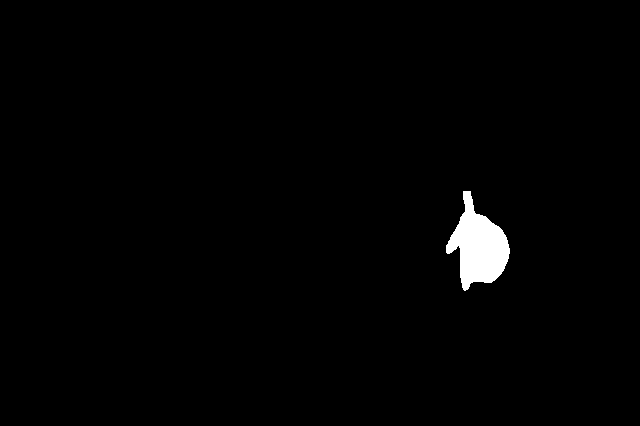}} 
		&
		\makecell[c]{\includegraphics[width=0.08\linewidth,height=0.06\linewidth]{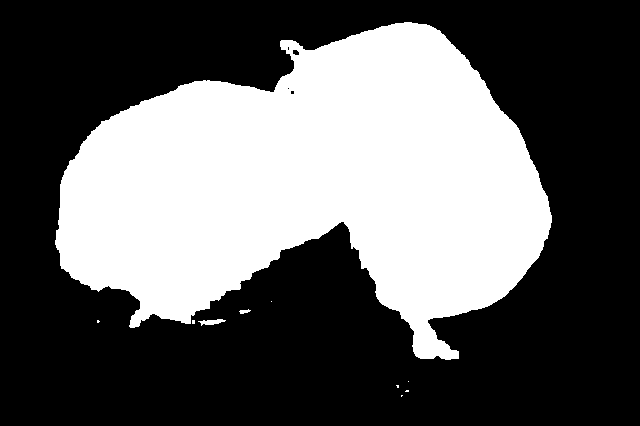}}        
		&
		\makecell[c]{\includegraphics[width=0.08\linewidth,height=0.06\linewidth]{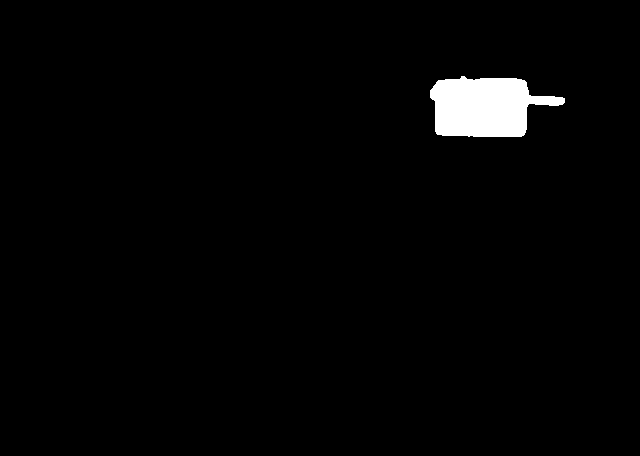}} 
		&
		\makecell[c]{\includegraphics[width=0.08\linewidth,height=0.06\linewidth]{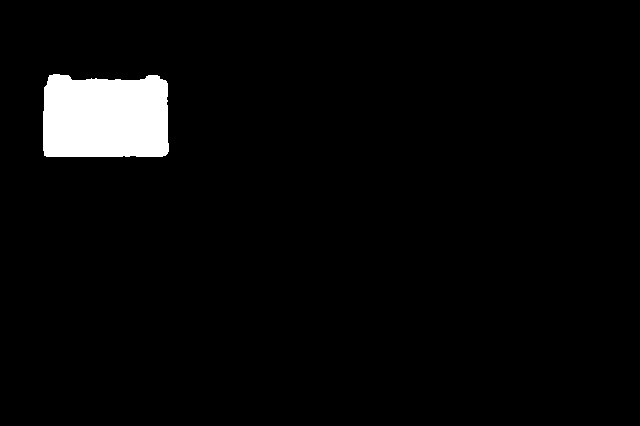}} 
		&
		\makecell[c]{\includegraphics[width=0.08\linewidth,height=0.06\linewidth]{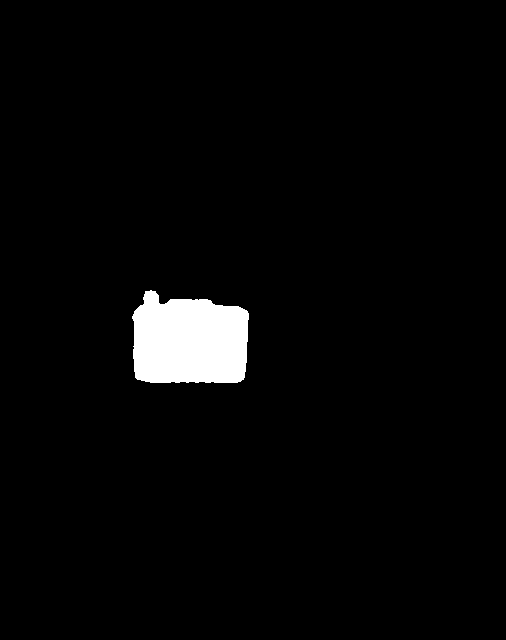}} 
		&
		\makecell[c]{\includegraphics[width=0.08\linewidth,height=0.06\linewidth]{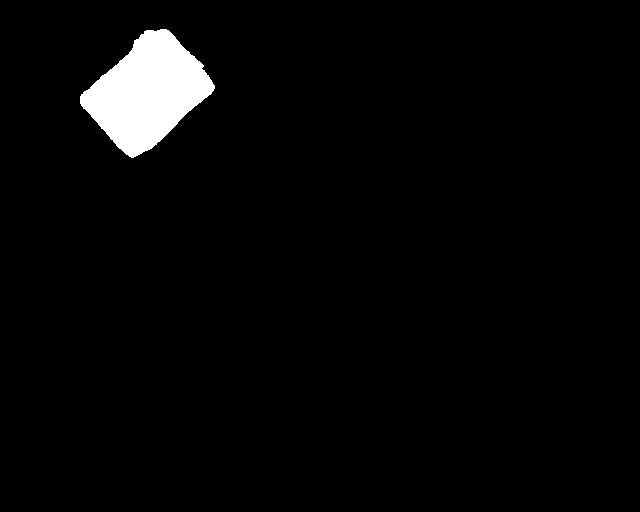}}
        &
		\makecell[c]{\includegraphics[width=0.08\linewidth,height=0.06\linewidth]{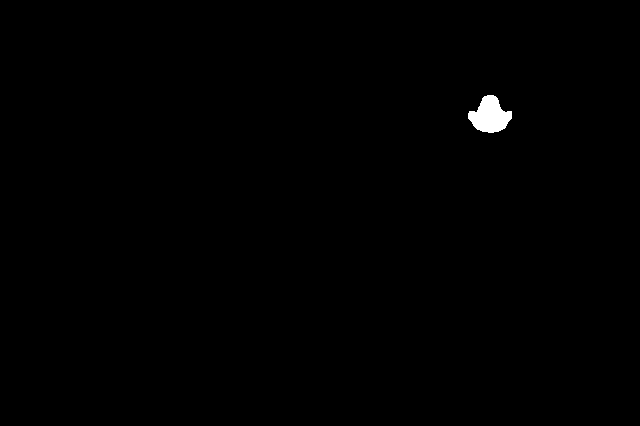}} 
		&
		\makecell[c]{\includegraphics[width=0.08\linewidth,height=0.06\linewidth]{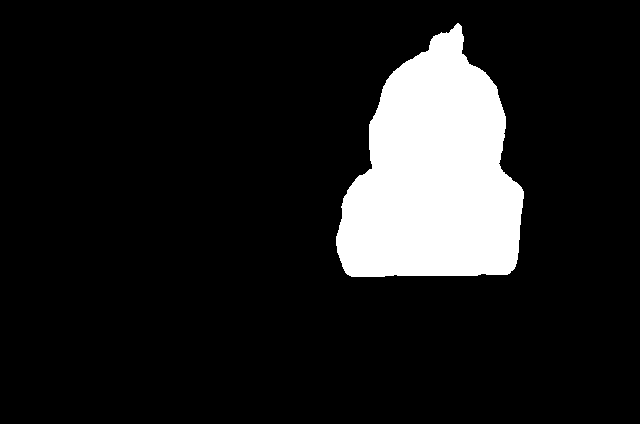}} 
		&
		\makecell[c]{\includegraphics[width=0.08\linewidth,height=0.06\linewidth]{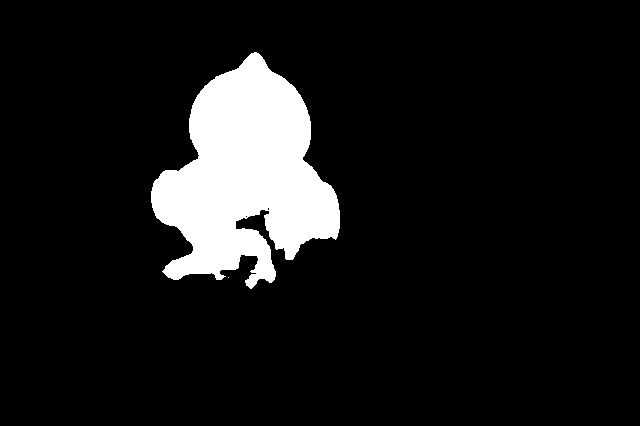}}
        &
		\makecell[c]{\includegraphics[width=0.08\linewidth,height=0.06\linewidth]{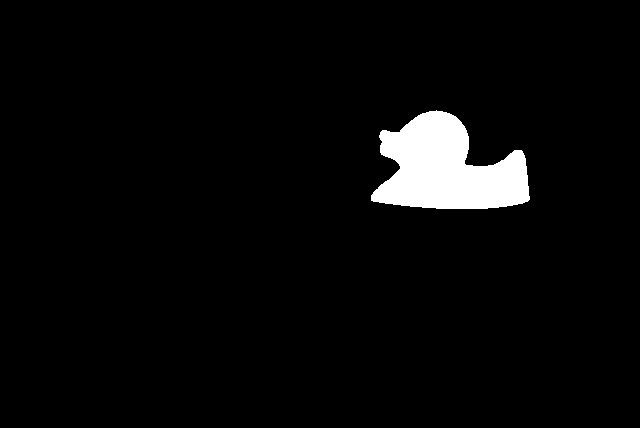}} 
            \vspace{-0.5mm}
		\\

            \rotatebox[origin=c]{90}{\small (d)}           
            &
		\makecell[c]{\includegraphics[width=0.08\linewidth,height=0.06\linewidth]{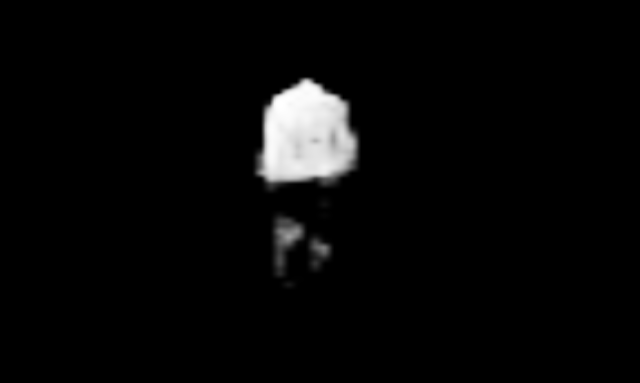}} 
		&
		\makecell[c]{\includegraphics[width=0.08\linewidth,height=0.06\linewidth]{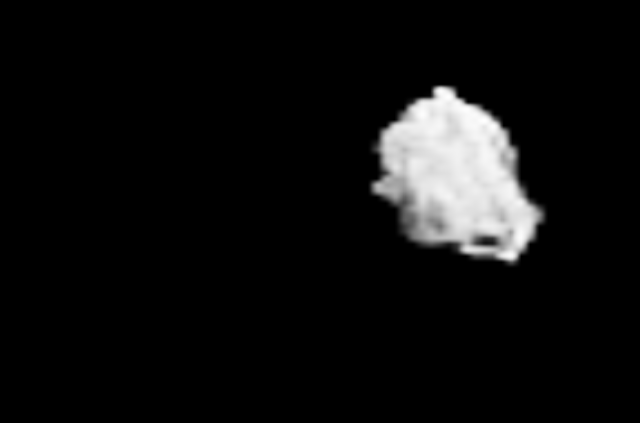}} 
		&
		\makecell[c]{\includegraphics[width=0.08\linewidth,height=0.06\linewidth]{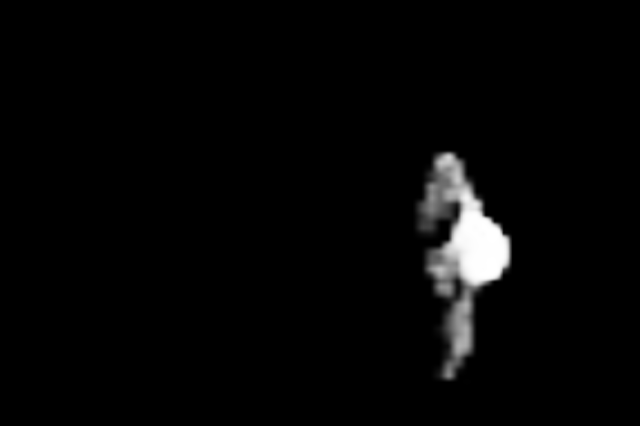}} 
		&
		\makecell[c]{\includegraphics[width=0.08\linewidth,height=0.06\linewidth]{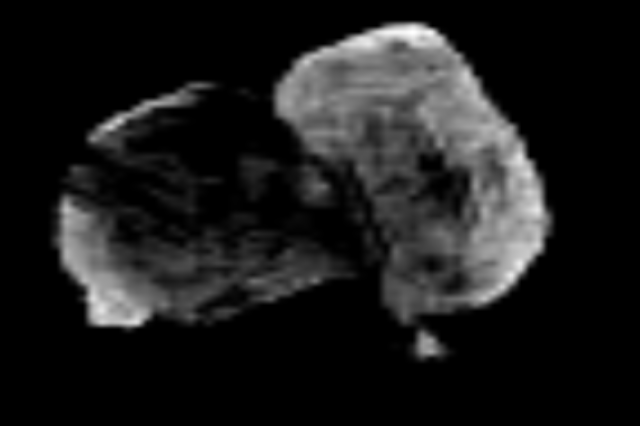}}     
		&
		\makecell[c]{\includegraphics[width=0.08\linewidth,height=0.06\linewidth]{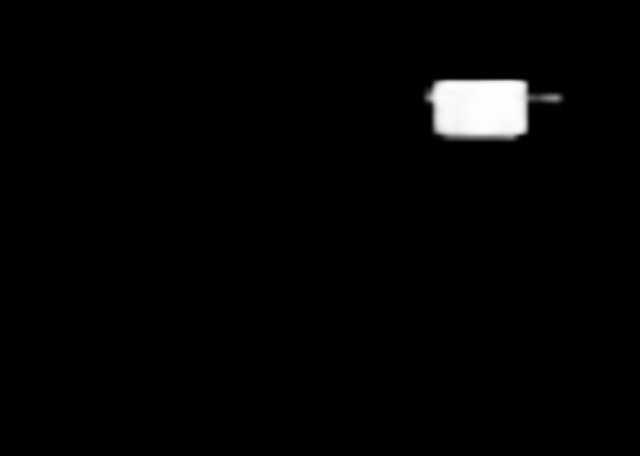}} 
		&
		\makecell[c]{\includegraphics[width=0.08\linewidth,height=0.06\linewidth]{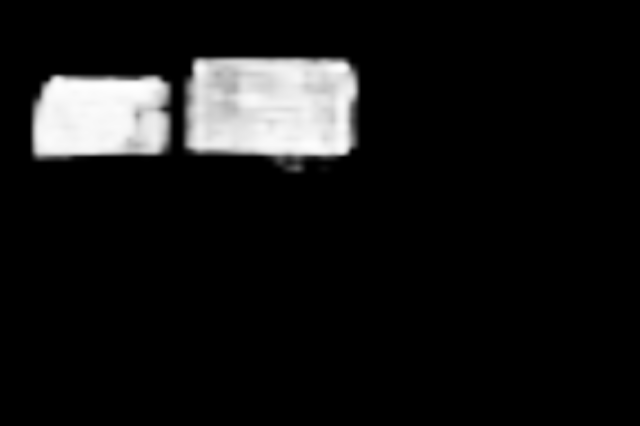}} 
		&
		\makecell[c]{\includegraphics[width=0.08\linewidth,height=0.06\linewidth]{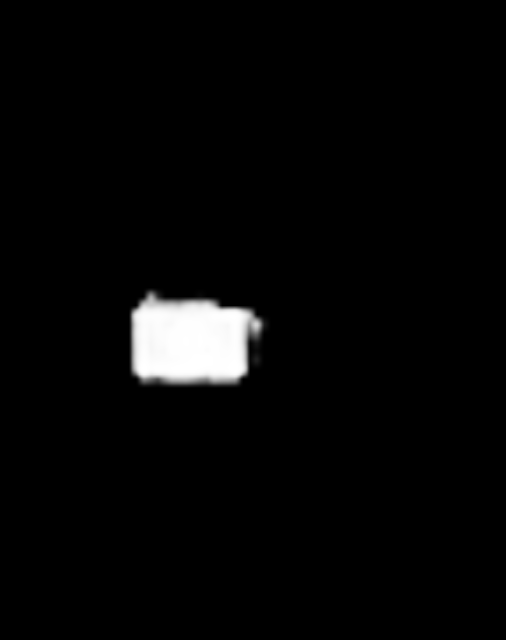}} 
		&
		\makecell[c]{\includegraphics[width=0.08\linewidth,height=0.06\linewidth]{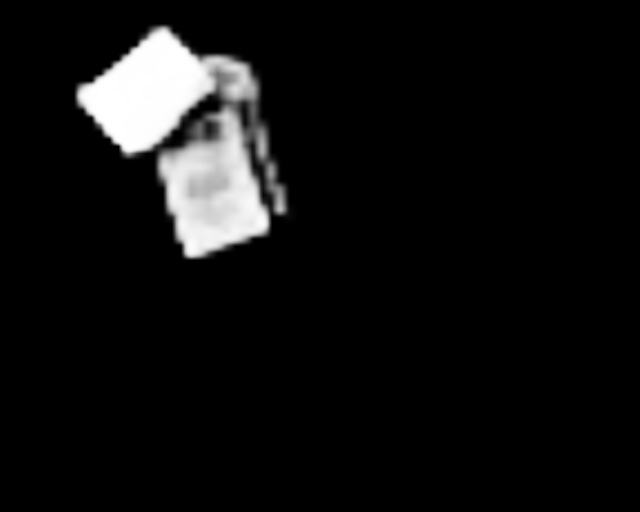}}
        &
		\makecell[c]{\includegraphics[width=0.08\linewidth,height=0.06\linewidth]{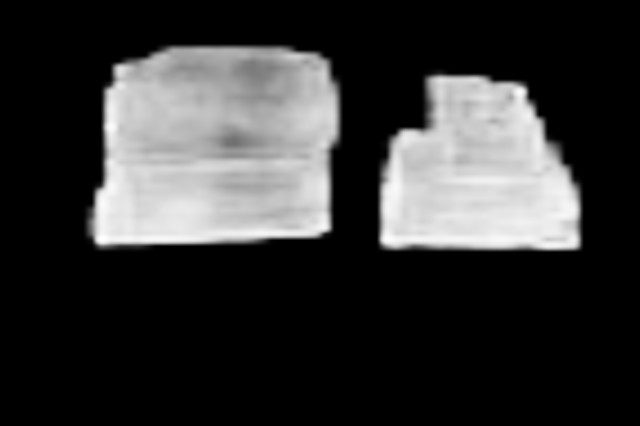}} 
        &
		\makecell[c]{\includegraphics[width=0.08\linewidth,height=0.06\linewidth]{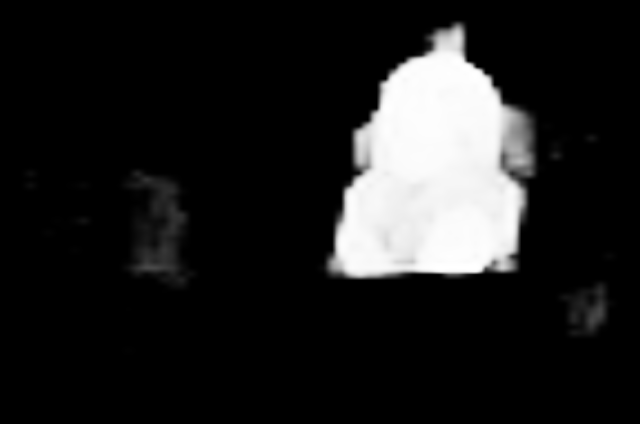}} 
		&
		\makecell[c]{\includegraphics[width=0.08\linewidth,height=0.06\linewidth]{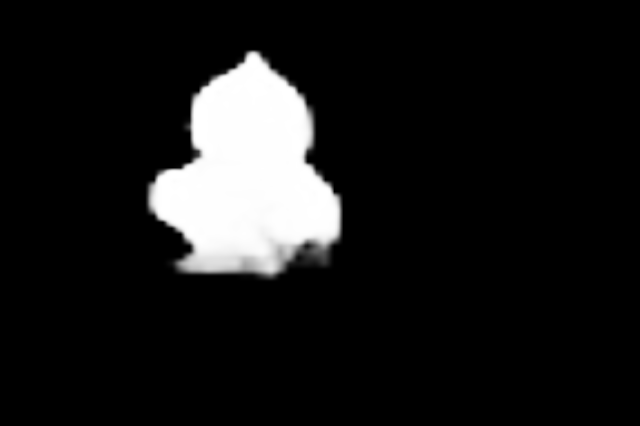}}
        &
		\makecell[c]{\includegraphics[width=0.08\linewidth,height=0.06\linewidth]{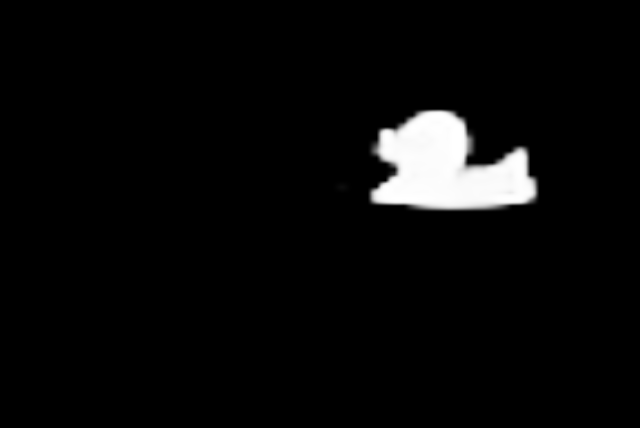}} 
            \vspace{-0.5mm}
		\\

             \rotatebox[origin=c]{90}{\small (e)}
            &
		\makecell[c]{\includegraphics[width=0.08\linewidth,height=0.06\linewidth]{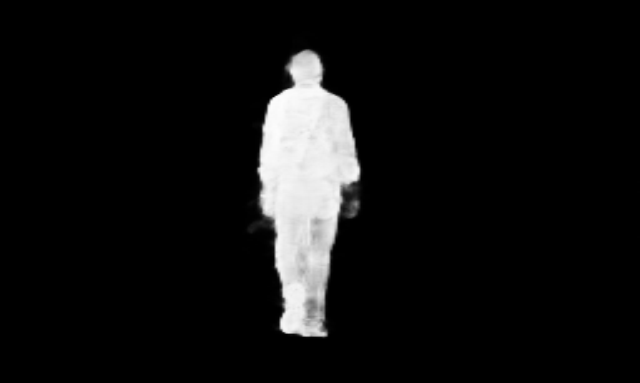}} 
		&
		\makecell[c]{\includegraphics[width=0.08\linewidth,height=0.06\linewidth]{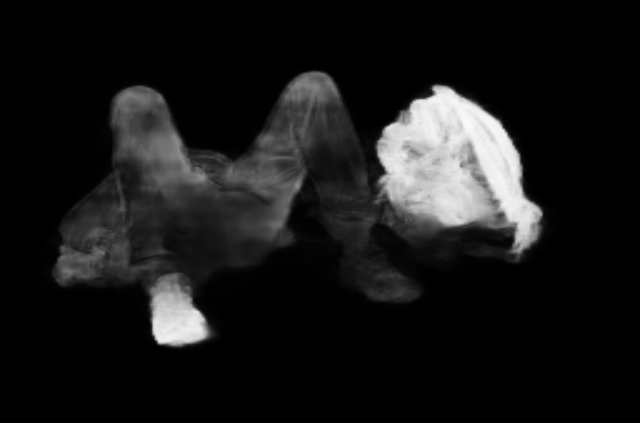}} 
		&
		\makecell[c]{\includegraphics[width=0.08\linewidth,height=0.06\linewidth]{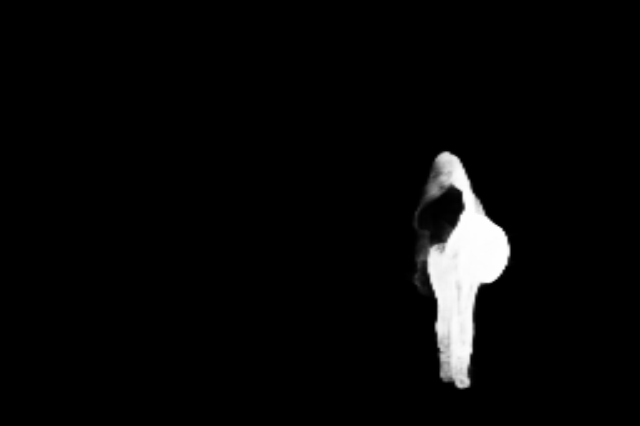}} 
		&
		\makecell[c]{\includegraphics[width=0.08\linewidth,height=0.06\linewidth]{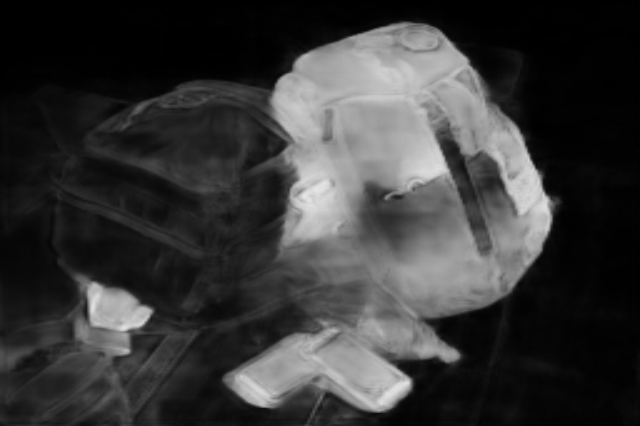}}     
		&
		\makecell[c]{\includegraphics[width=0.08\linewidth,height=0.06\linewidth]{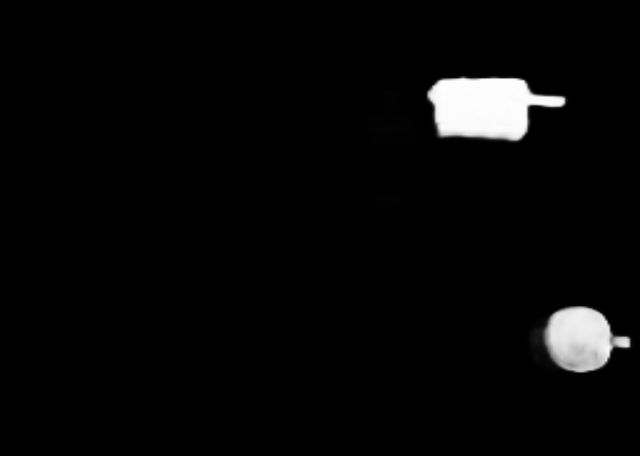}} 
		&
		\makecell[c]{\includegraphics[width=0.08\linewidth,height=0.06\linewidth]{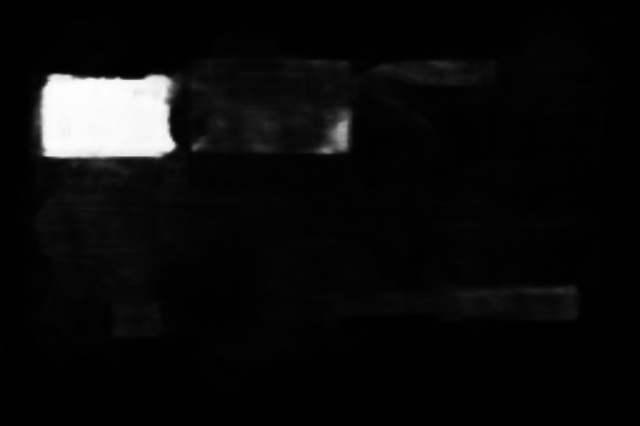}} 
		&
		\makecell[c]{\includegraphics[width=0.08\linewidth,height=0.06\linewidth]{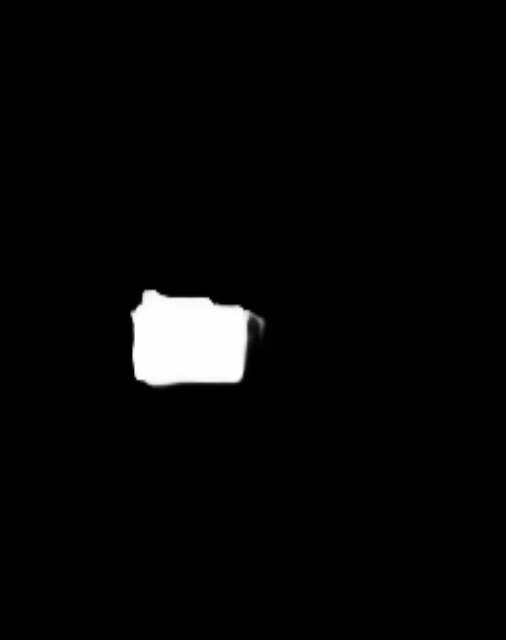}} 
		&
		\makecell[c]{\includegraphics[width=0.08\linewidth,height=0.06\linewidth]{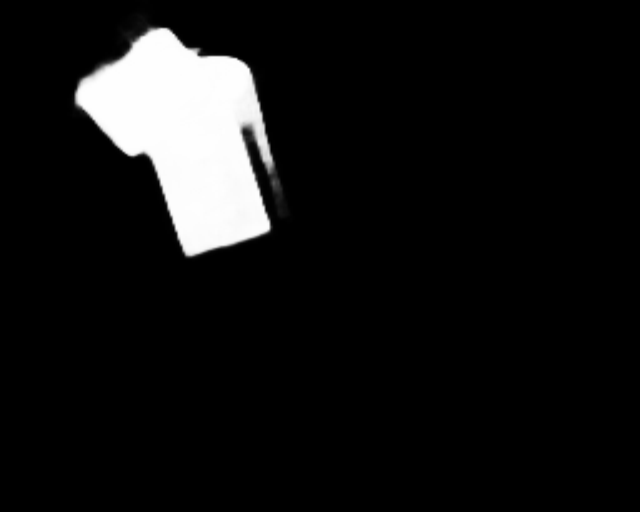}}
            &
		\makecell[c]{\includegraphics[width=0.08\linewidth,height=0.06\linewidth]{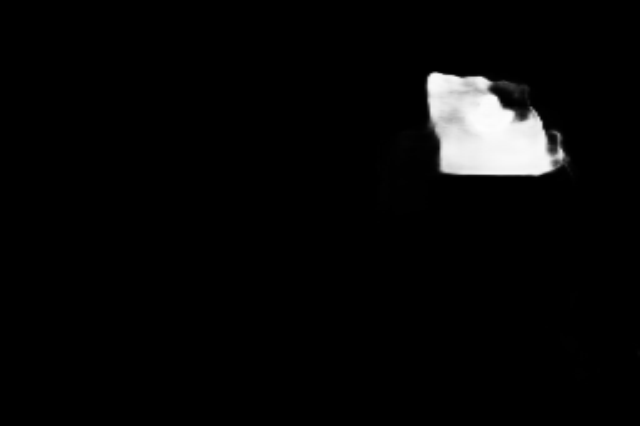}} 
		&
		\makecell[c]{\includegraphics[width=0.08\linewidth,height=0.06\linewidth]{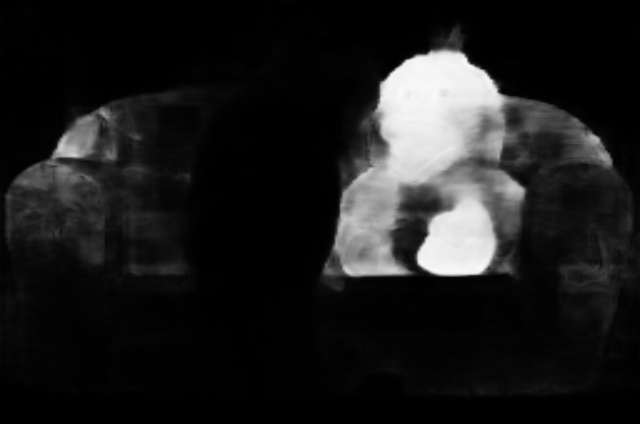}} 
		&
		\makecell[c]{\includegraphics[width=0.08\linewidth,height=0.06\linewidth]{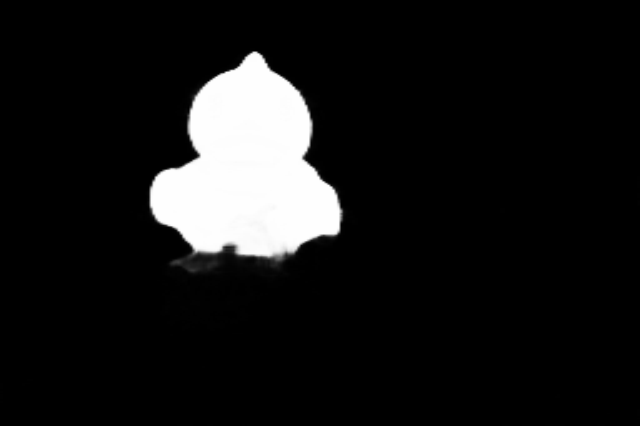}}
            &
		\makecell[c]{\includegraphics[width=0.08\linewidth,height=0.06\linewidth]{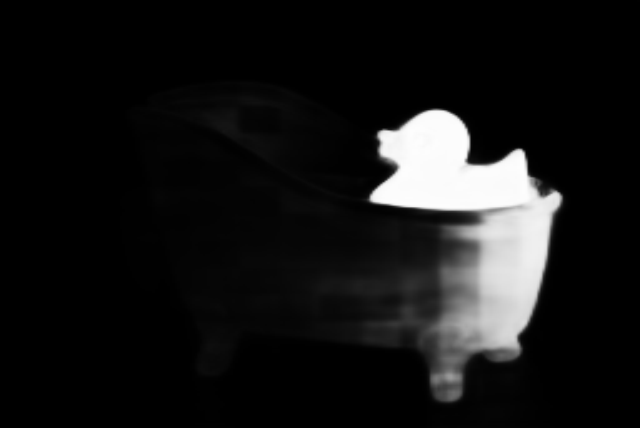}} 
            \vspace{-0.5mm}
		\\

            \rotatebox[origin=c]{90}{\small (f)}
        &
		\makecell[c]{\includegraphics[width=0.08\linewidth,height=0.06\linewidth]{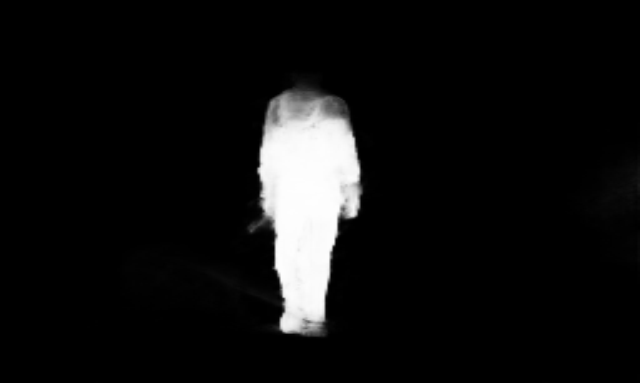}} 
		&
		\makecell[c]{\includegraphics[width=0.08\linewidth,height=0.06\linewidth]{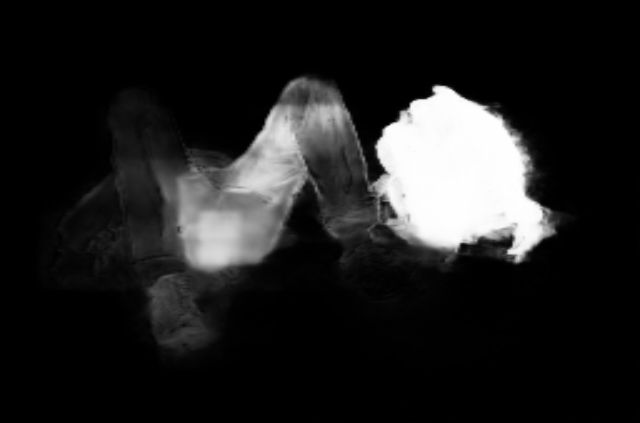}} 
		&
		\makecell[c]{\includegraphics[width=0.08\linewidth,height=0.06\linewidth]{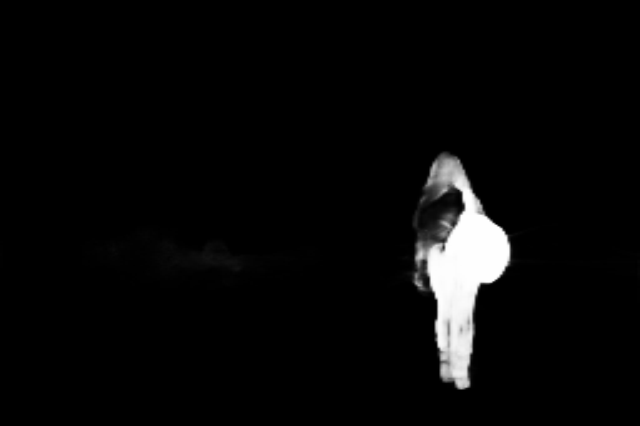}} 
		&
		\makecell[c]{\includegraphics[width=0.08\linewidth,height=0.06\linewidth]{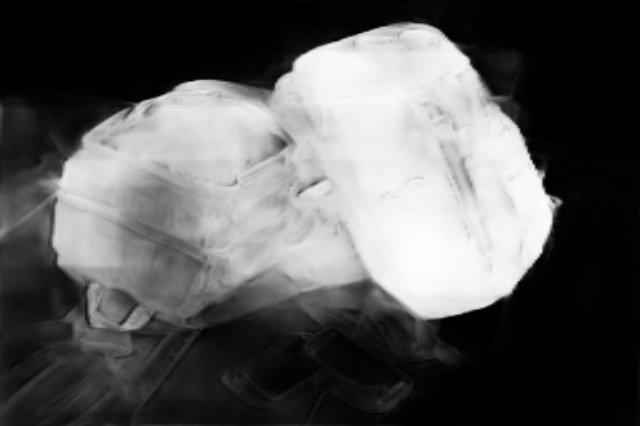}}     
		&
		\makecell[c]{\includegraphics[width=0.08\linewidth,height=0.06\linewidth]{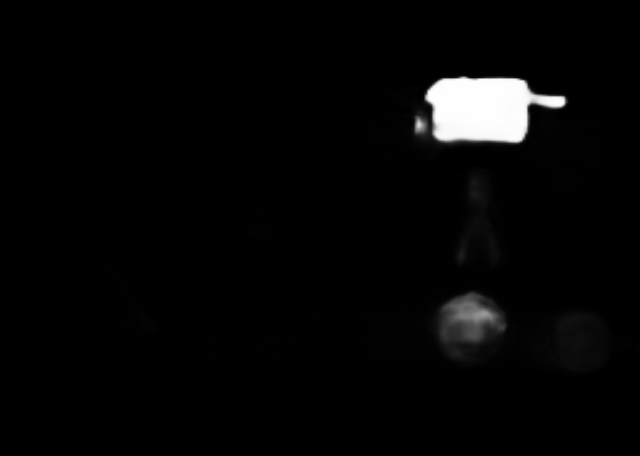}} 
		&
		\makecell[c]{\includegraphics[width=0.08\linewidth,height=0.06\linewidth]{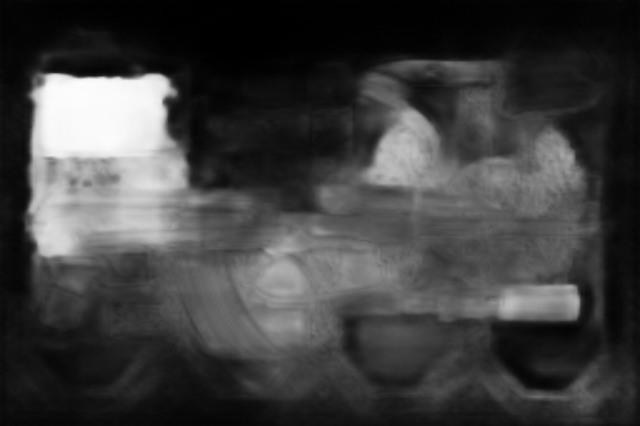}} 
		&
		\makecell[c]{\includegraphics[width=0.08\linewidth,height=0.06\linewidth]{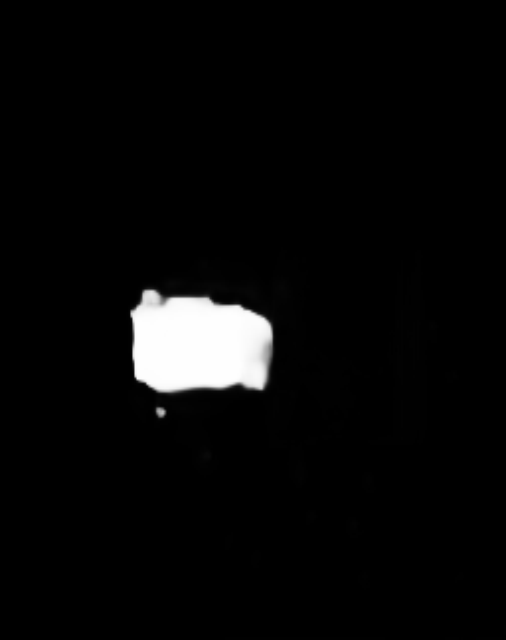}} 
		&
		\makecell[c]{\includegraphics[width=0.08\linewidth,height=0.06\linewidth]{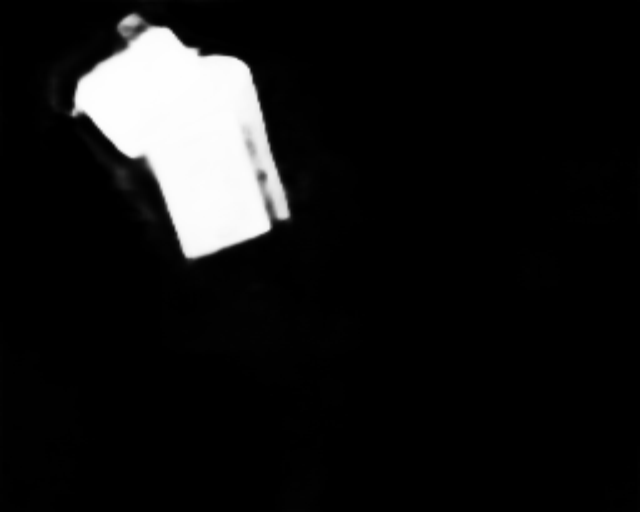}}
            &
		\makecell[c]{\includegraphics[width=0.08\linewidth,height=0.06\linewidth]{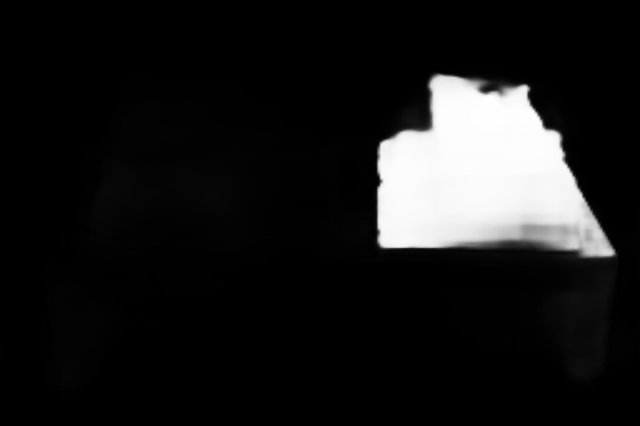}} 
		&
		\makecell[c]{\includegraphics[width=0.08\linewidth,height=0.06\linewidth]{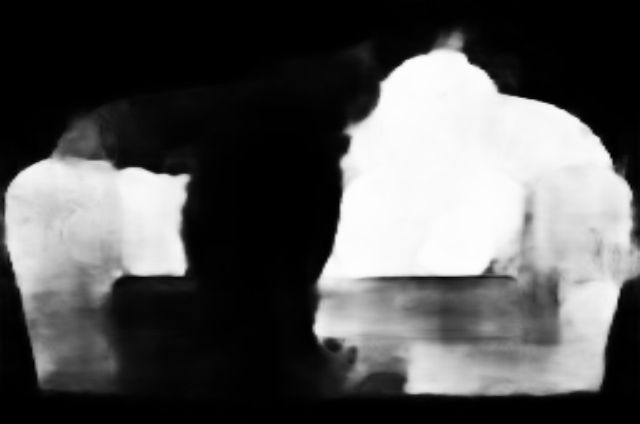}} 
		&
		\makecell[c]{\includegraphics[width=0.08\linewidth,height=0.06\linewidth]{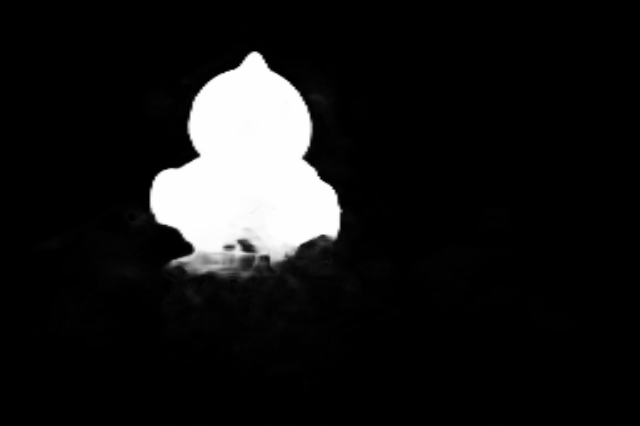}}
            &
		\makecell[c]{\includegraphics[width=0.08\linewidth,height=0.06\linewidth]{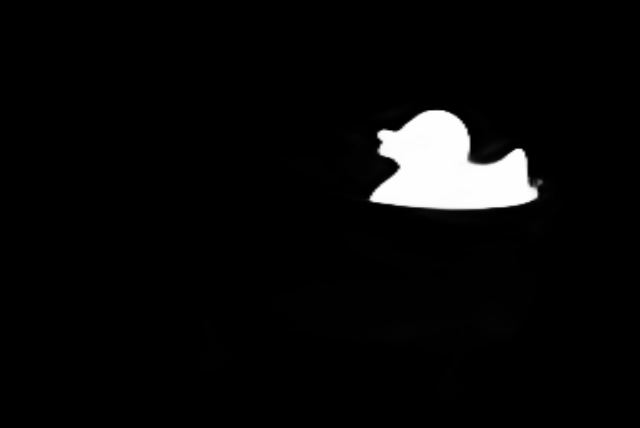}} 
            \vspace{-0.5mm}
		\\

            \rotatebox[origin=c]{90}{\small (g)}
        &
		\makecell[c]{\includegraphics[width=0.08\linewidth,height=0.06\linewidth]{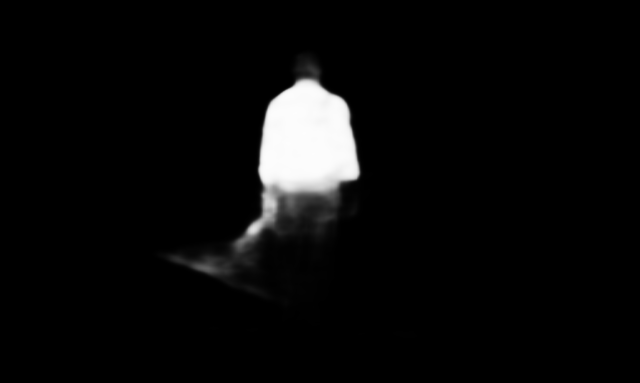}} 
		&
		\makecell[c]{\includegraphics[width=0.08\linewidth,height=0.06\linewidth]{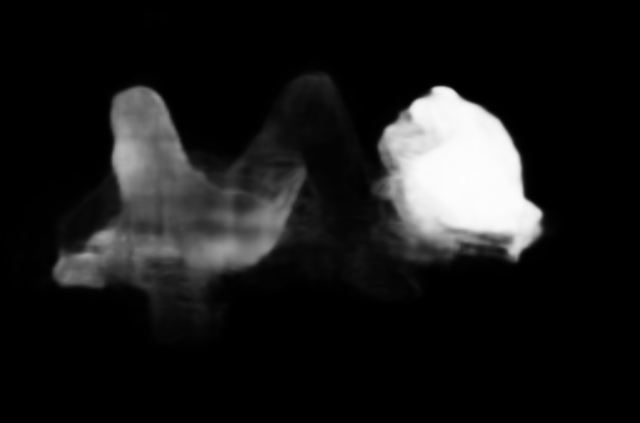}} 
		&
		\makecell[c]{\includegraphics[width=0.08\linewidth,height=0.06\linewidth]{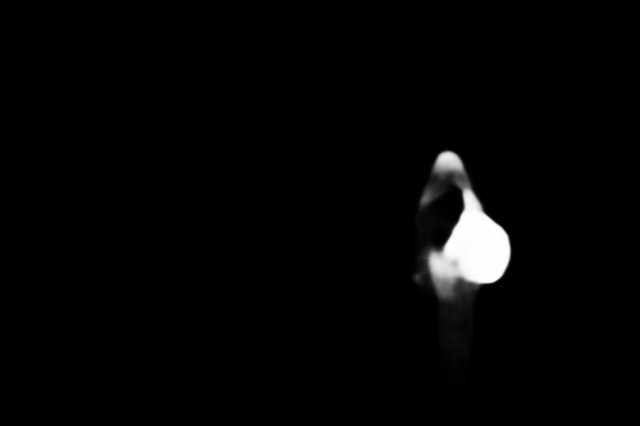}} 
		&
		\makecell[c]{\includegraphics[width=0.08\linewidth,height=0.06\linewidth]{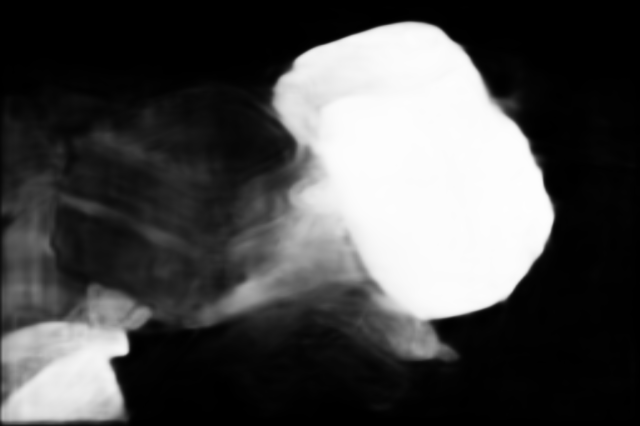}}     
		&
		\makecell[c]{\includegraphics[width=0.08\linewidth,height=0.06\linewidth]{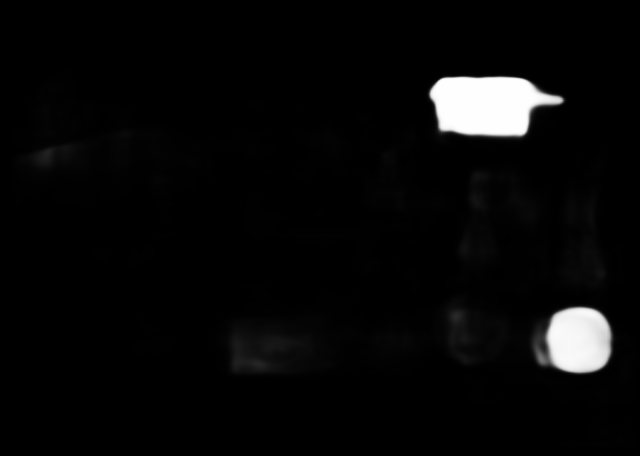}} 
		&
		\makecell[c]{\includegraphics[width=0.08\linewidth,height=0.06\linewidth]{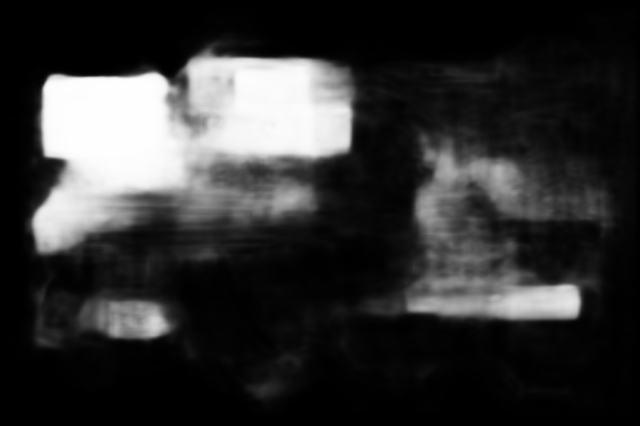}} 
		&
		\makecell[c]{\includegraphics[width=0.08\linewidth,height=0.06\linewidth]{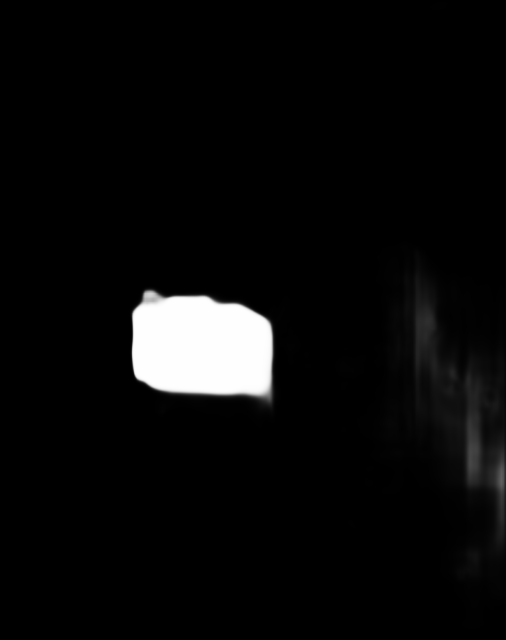}} 
		&
		\makecell[c]{\includegraphics[width=0.08\linewidth,height=0.06\linewidth]{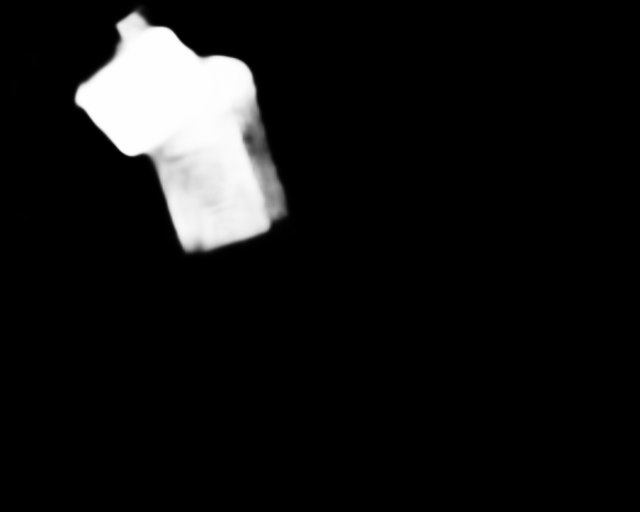}}
            &
		\makecell[c]{\includegraphics[width=0.08\linewidth,height=0.06\linewidth]{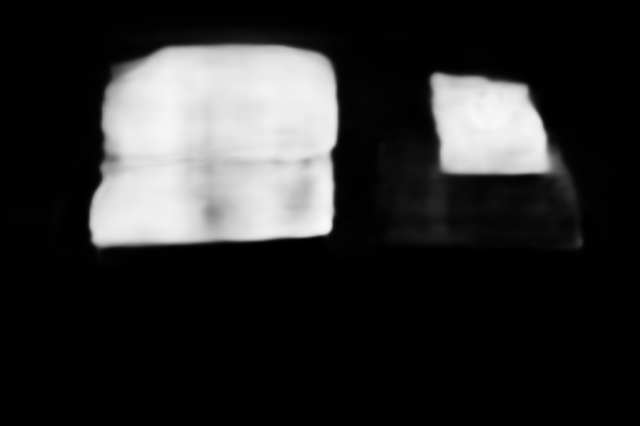}} 
		&
		\makecell[c]{\includegraphics[width=0.08\linewidth,height=0.06\linewidth]{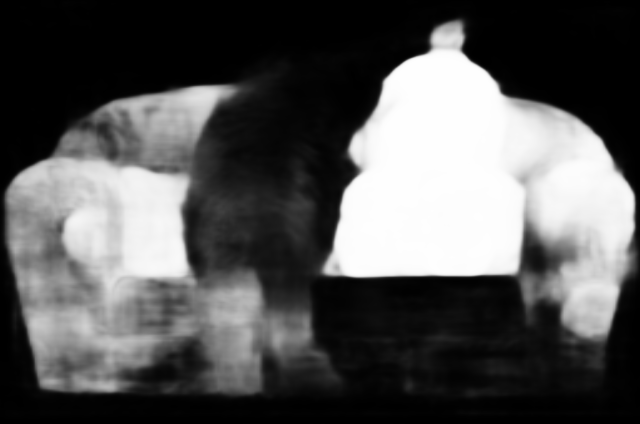}} 
		&
		\makecell[c]{\includegraphics[width=0.08\linewidth,height=0.06\linewidth]{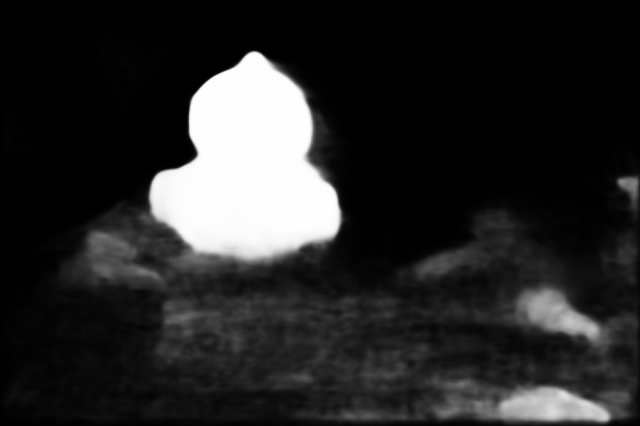}}
            &
		\makecell[c]{\includegraphics[width=0.08\linewidth,height=0.06\linewidth]{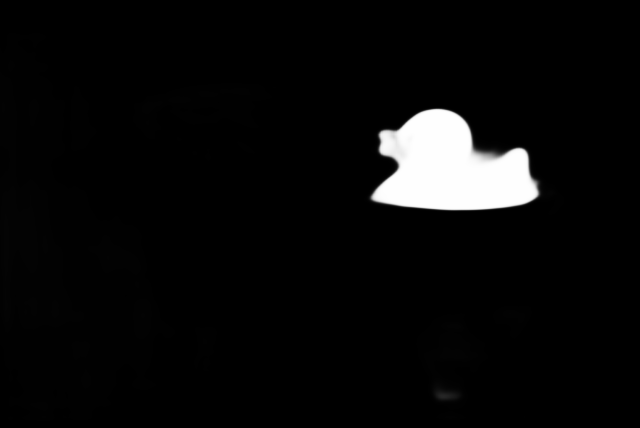}} 
            \vspace{-0.5mm}
		\\

		\end{tabular}
            \vspace{-2mm}
    \caption{\textbf{Qualitative comparison with specialized SOTA methods on CoSOD.} Here, \textbf{(a)} denotes the input image group, \textbf{(b)} the ground-truth co-salient masks, \textbf{(c)} our Saliency-R1, followed by: \textbf{(d)} VCP, \textbf{(e)} DMT+, \textbf{(f)} CONDA, and \textbf{(g)} GCoNet+.}
    \label{SOTA_CoSOD}
\end{minipage}
\end{figure*}

\begin{figure}[t] 
\begin{minipage}[c]{1.\columnwidth} 
    \tiny
	\renewcommand{\tabcolsep}{1.0pt} 
	\renewcommand{\arraystretch}{1.5} 
	\centering
        \begin{tabular}{ccccc}
	    \\

            \arrayrulecolor{red}
            \rotatebox[origin=c]{90}{\small (a)}
            &
		\makecell[c]{\includegraphics[width=0.226\linewidth,height=0.125\linewidth]{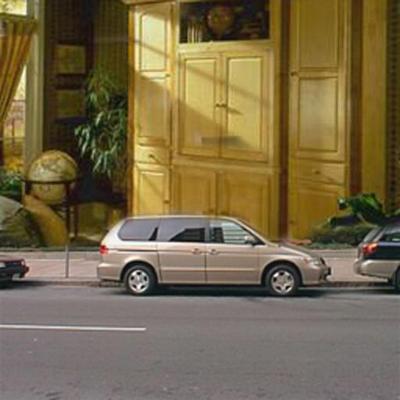}}
        &
		\makecell[c]{\includegraphics[width=0.226\linewidth,height=0.125\linewidth]{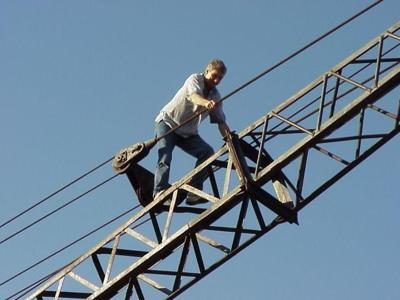}}
		&
		\makecell[c]{\includegraphics[width=0.226\linewidth,height=0.125\linewidth]{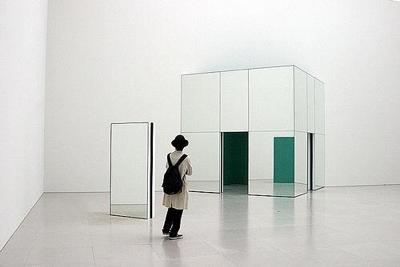}}
		&
		\makecell[c]{\includegraphics[width=0.226\linewidth,height=0.125\linewidth]{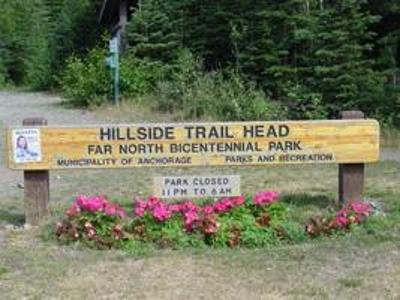}}
            \vspace{-0.5mm}
            \\

            \rotatebox[origin=c]{90}{\small (b)}
            &
		\makecell[c]{\includegraphics[width=0.226\linewidth,height=0.125\linewidth]{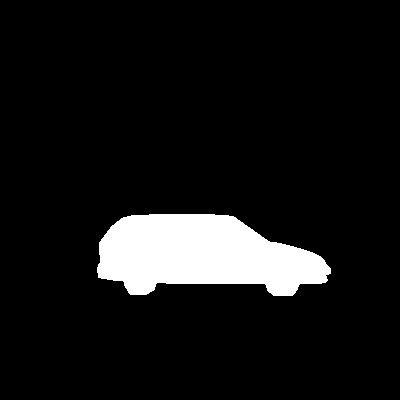}}
        &
		\makecell[c]{\includegraphics[width=0.226\linewidth,height=0.125\linewidth]{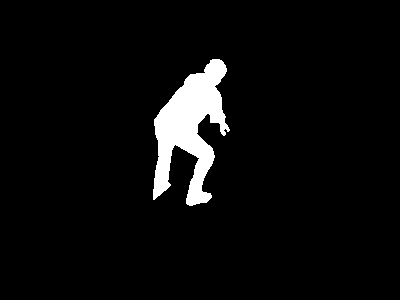}}
		&
		\makecell[c]{\includegraphics[width=0.226\linewidth,height=0.125\linewidth]{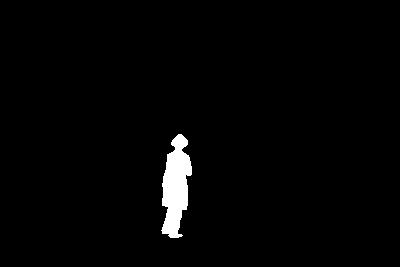}}
		&
		\makecell[c]{\includegraphics[width=0.226\linewidth,height=0.125\linewidth]{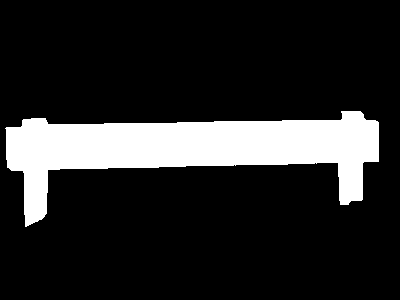}}
            \vspace{-0.5mm}
		\\

            \rotatebox[origin=c]{90}{\small (c)}  
            &
		\makecell[c]{\includegraphics[width=0.226\linewidth,height=0.125\linewidth]{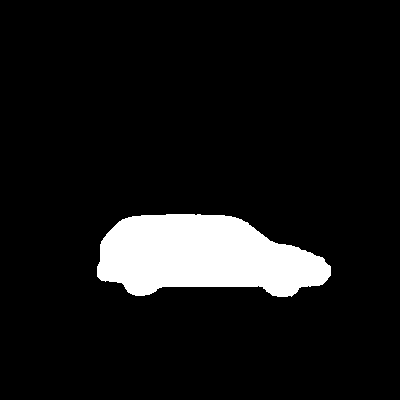}}
        &
		\makecell[c]{\includegraphics[width=0.226\linewidth,height=0.125\linewidth]{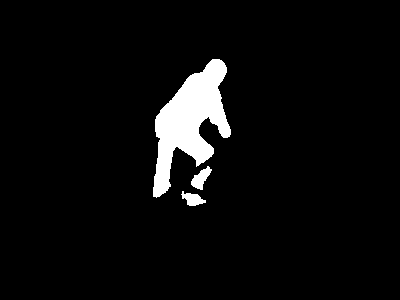}}
		&
		\makecell[c]{\includegraphics[width=0.226\linewidth,height=0.125\linewidth]{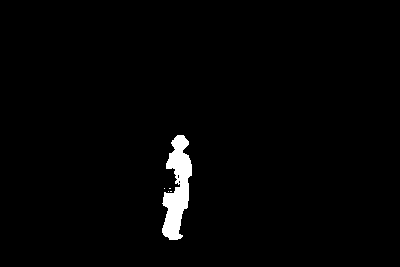}}
		&
		\makecell[c]{\includegraphics[width=0.226\linewidth,height=0.125\linewidth]{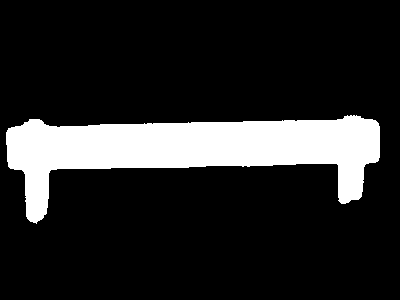}}      
            \vspace{-0.5mm}
		\\

            \rotatebox[origin=c]{90}{\small (d)}           
            &
		\makecell[c]{\includegraphics[width=0.226\linewidth,height=0.125\linewidth]{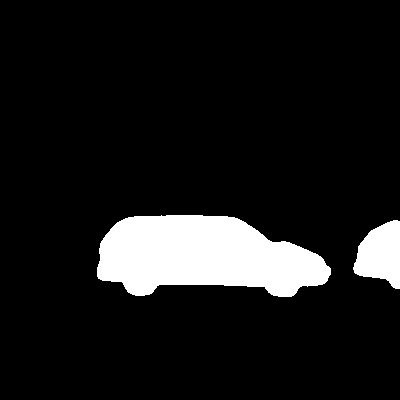}}
        &
		\makecell[c]{\includegraphics[width=0.226\linewidth,height=0.125\linewidth]{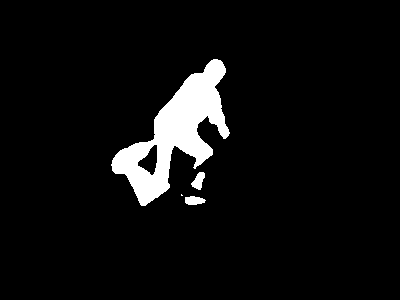}}
		&
		\makecell[c]{\includegraphics[width=0.226\linewidth,height=0.125\linewidth]{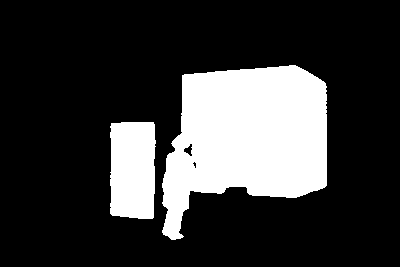}}
		&
		\makecell[c]{\includegraphics[width=0.226\linewidth,height=0.125\linewidth]{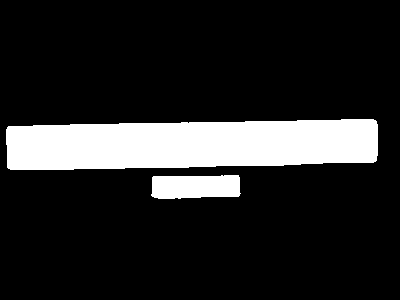}}  
            \vspace{-0.5mm}
		\\

             \rotatebox[origin=c]{90}{\small (e)}
            &
		\makecell[c]{\includegraphics[width=0.226\linewidth,height=0.125\linewidth]{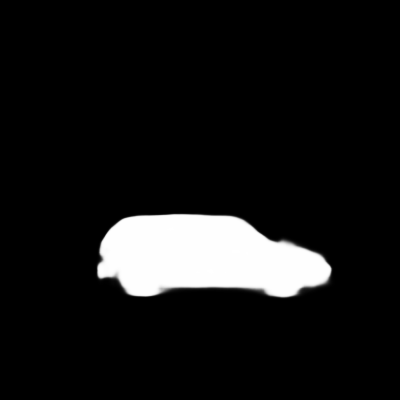}}
        &
		\makecell[c]{\includegraphics[width=0.226\linewidth,height=0.125\linewidth]{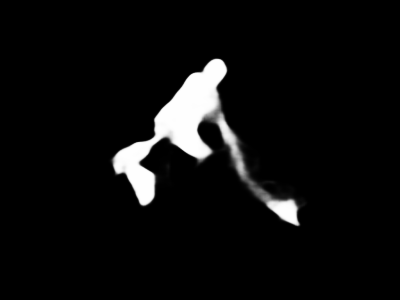}}
		&
		\makecell[c]{\includegraphics[width=0.226\linewidth,height=0.125\linewidth]{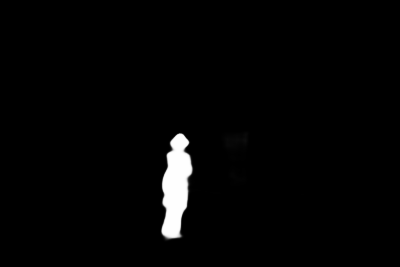}}
		&
		\makecell[c]{\includegraphics[width=0.226\linewidth,height=0.125\linewidth]{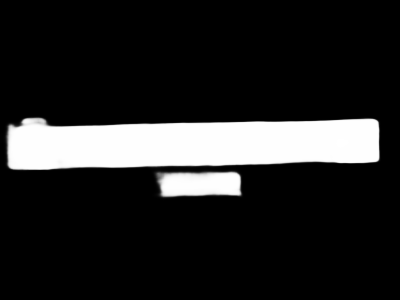}} 
            \vspace{-0.5mm}
		\\

            \rotatebox[origin=c]{90}{\small (f)}
            &
		\makecell[c]{\includegraphics[width=0.226\linewidth,height=0.125\linewidth]{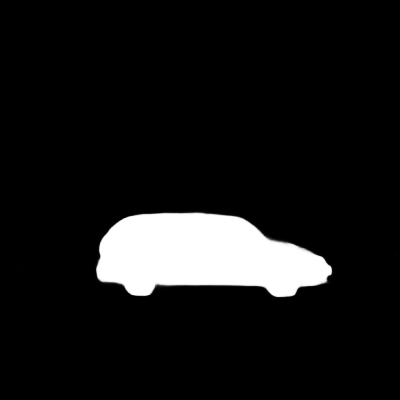}}
        &
		\makecell[c]{\includegraphics[width=0.226\linewidth,height=0.125\linewidth]{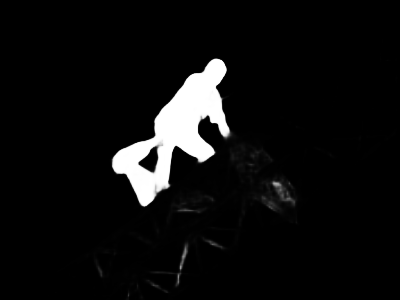}}
		&
		\makecell[c]{\includegraphics[width=0.226\linewidth,height=0.125\linewidth]{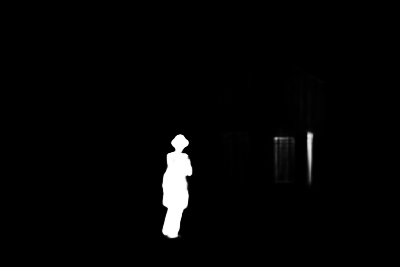}}
		&
		\makecell[c]{\includegraphics[width=0.226\linewidth,height=0.125\linewidth]{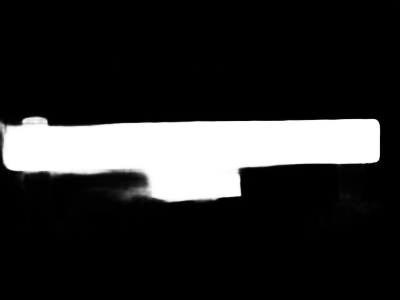}}   
            \vspace{-0.5mm}
		\\

            \rotatebox[origin=c]{90}{\small (g)}
            &
		\makecell[c]{\includegraphics[width=0.226\linewidth,height=0.125\linewidth]{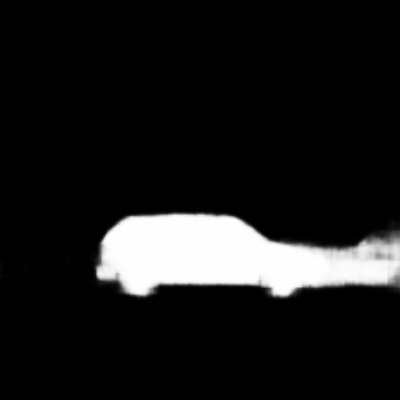}}
        &
		\makecell[c]{\includegraphics[width=0.226\linewidth,height=0.125\linewidth]{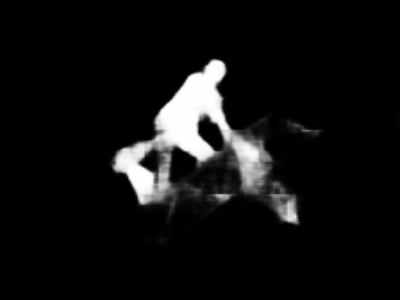}}
		&
		\makecell[c]{\includegraphics[width=0.226\linewidth,height=0.125\linewidth]{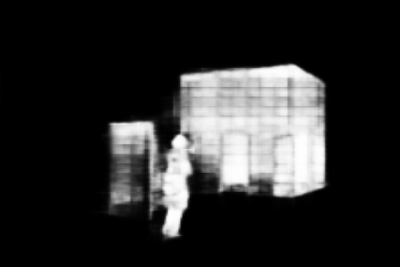}}
		&
		\makecell[c]{\includegraphics[width=0.226\linewidth,height=0.125\linewidth]{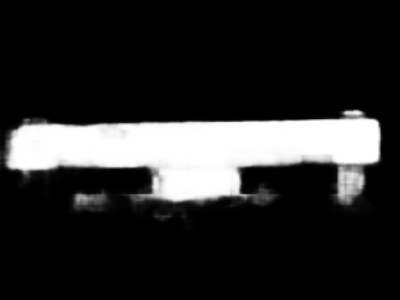}}     
            \vspace{-0.5mm}
		\\

		\end{tabular}
            \vspace{-2mm}
    \caption{\textbf{Qualitative comparison with specialized SOTA methods on SOD.} \textbf{(a)} denotes the input image, \textbf{(b)} the ground-truth mask, \textbf{(c)} our Saliency-R1, followed by: \textbf{(d)} FOCUS, \textbf{(e)} DMT+, \textbf{(f)} SelfReformer, \textbf{(g)} MENet, and \textbf{(h)} VST.}
    \label{SOTA_SOD}
\end{minipage}
\end{figure}

We further compare our method against specialized SOTA methods, \ie, VST \cite{liu2021visual}, MENet \cite{wang2023pixels}, SelFReformer \cite{yun2023towards} (SOD); S4Net~\cite{fan2019s4net}, RDPNet~\cite{wu2021regularized}, OQTR~\cite{pei2022transformer} (SIS); GCoNet+~\cite{zheng2023gconet+}, CONDA~\cite{li2024conda}, DMT+~\cite{li2023discriminative}, and VCP~\cite{wang2025visual} (CoSOD). As shown in Figure~\ref{SOTA_SOD}–\ref{SOTA_CoSOD}, Saliency-R1 matches or exceeds their performance, even though we use far less task-specific training data.

In SOD, as shown in Figure~\ref{SOTA_SOD}, CNN/Transformer-based models (\eg, MENet, SelfReformer) suffer from over-prediction of background regions (\eg, the maintenance worker and scaffolding are not separated in Column 2, and the sign beneath the bench is incorrectly included in the bench's mask in Column 4) and under-prediction of object parts (\eg, the legs of the bench are missed in Column 4). Our approach, guided by contrast-aware CoT, recovers sharp edges and complete object extent.

\begin{figure}[t] 
\begin{minipage}[c]{1.\columnwidth}
    \tiny
	\renewcommand{\tabcolsep}{0.5pt} 
	\renewcommand{\arraystretch}{1.2} 
	\centering
        \begin{tabular}{ccccc}
	    \\

            \arrayrulecolor{red}
            \rotatebox[origin=c]{90}{\small (a)}
            &
		\makecell[c]{\includegraphics[width=0.225\linewidth,height=0.125\linewidth]{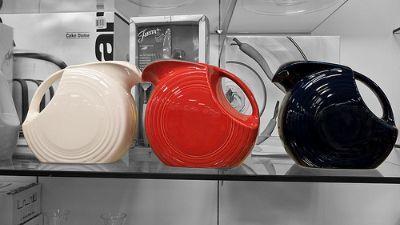}}
        &
		\makecell[c]{\includegraphics[width=0.225\linewidth,height=0.125\linewidth]{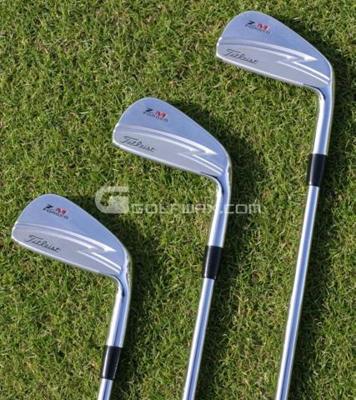}}
		&
		\makecell[c]{\includegraphics[width=0.225\linewidth,height=0.125\linewidth]{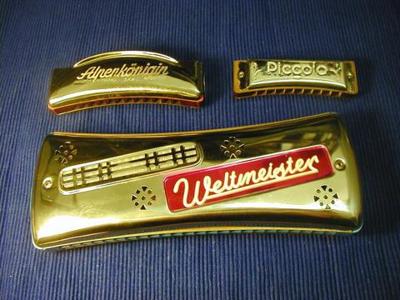}}
		&
		\makecell[c]{\includegraphics[width=0.225\linewidth,height=0.125\linewidth]{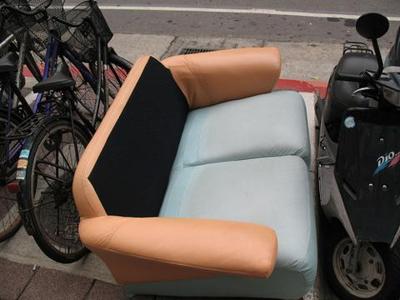}}
            \vspace{-0.5mm}
            \\

            \rotatebox[origin=c]{90}{\small (b)}
            &
		\makecell[c]{\includegraphics[width=0.225\linewidth,height=0.125\linewidth]{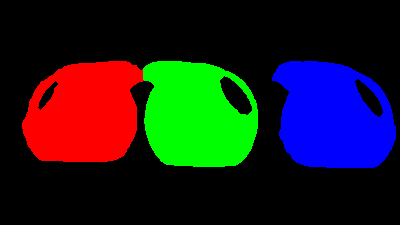}}
        &
		\makecell[c]{\includegraphics[width=0.225\linewidth,height=0.125\linewidth]{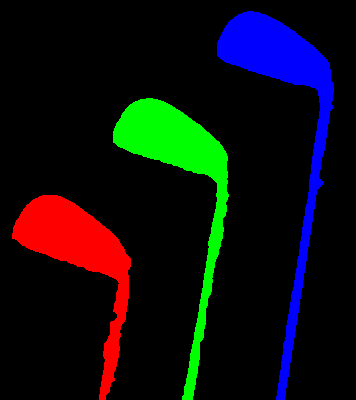}}
		&
		\makecell[c]{\includegraphics[width=0.225\linewidth,height=0.125\linewidth]{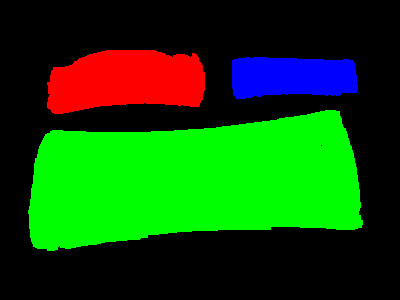}}
		&
		\makecell[c]{\includegraphics[width=0.225\linewidth,height=0.125\linewidth]{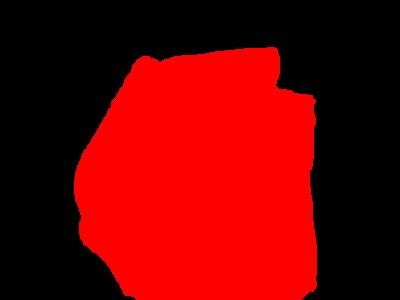}}
            \vspace{-0.5mm}
		\\

            \rotatebox[origin=c]{90}{\small (c)}  
            &
		\makecell[c]{\includegraphics[width=0.225\linewidth,height=0.125\linewidth]{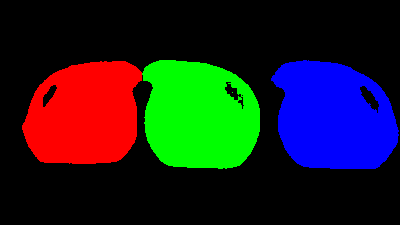}}
        &
		\makecell[c]{\includegraphics[width=0.225\linewidth,height=0.125\linewidth]{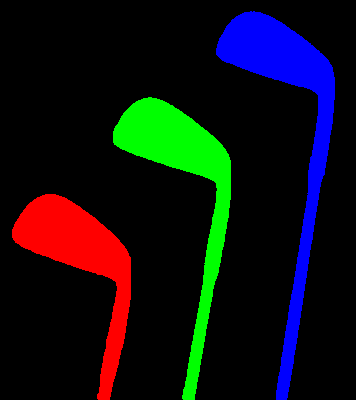}}
		&
		\makecell[c]{\includegraphics[width=0.225\linewidth,height=0.125\linewidth]{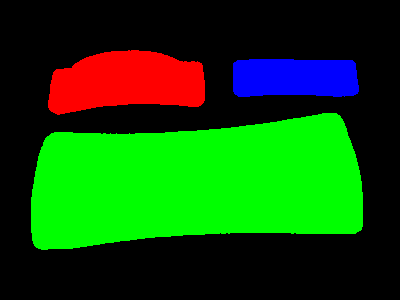}}
		&
		\makecell[c]{\includegraphics[width=0.225\linewidth,height=0.125\linewidth]{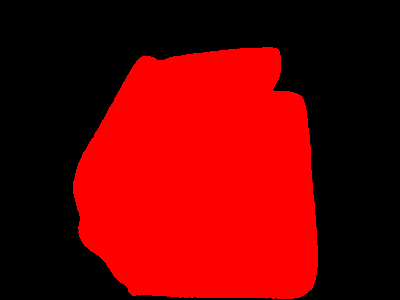}}   
            \vspace{-0.5mm}
		\\

            \rotatebox[origin=c]{90}{\small (d)}           
            &
		\makecell[c]{\includegraphics[width=0.225\linewidth,height=0.125\linewidth]{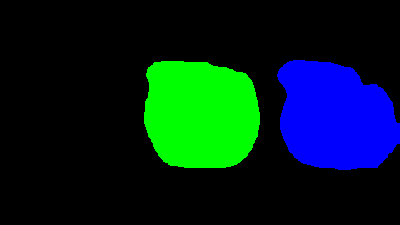}}
        &
		\makecell[c]{\includegraphics[width=0.225\linewidth,height=0.125\linewidth]{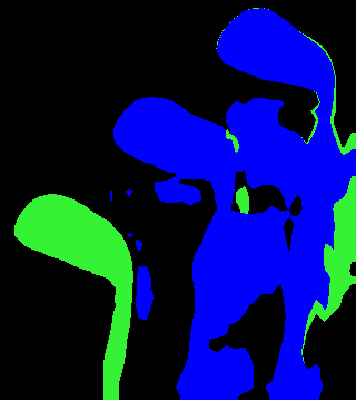}}
		&
		\makecell[c]{\includegraphics[width=0.225\linewidth,height=0.125\linewidth]{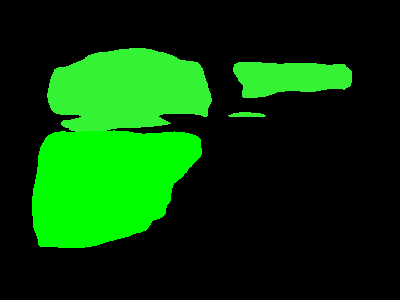}}
		&
		\makecell[c]{\includegraphics[width=0.225\linewidth,height=0.125\linewidth]{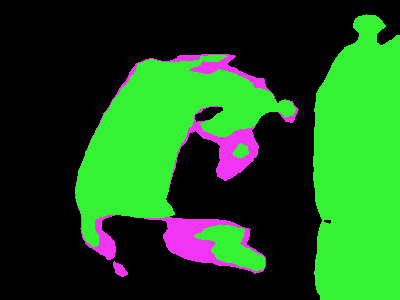}}
            \vspace{-0.5mm}
		\\

             \rotatebox[origin=c]{90}{\small (e)}
            &
		\makecell[c]{\includegraphics[width=0.225\linewidth,height=0.125\linewidth]{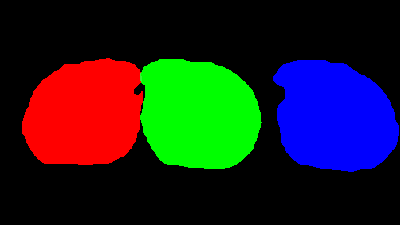}}
        &
		\makecell[c]{\includegraphics[width=0.225\linewidth,height=0.125\linewidth]{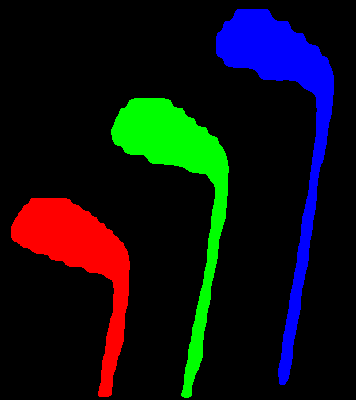}}
		&
		\makecell[c]{\includegraphics[width=0.225\linewidth,height=0.125\linewidth]{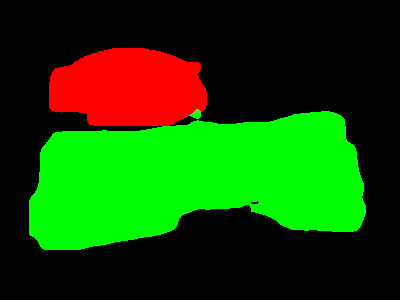}}
		&
		\makecell[c]{\includegraphics[width=0.225\linewidth,height=0.125\linewidth]{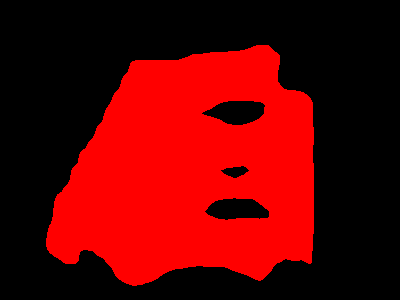}}
            \vspace{-0.5mm}
		\\

            \rotatebox[origin=c]{90}{\small (f)}
            &
		\makecell[c]{\includegraphics[width=0.225\linewidth,height=0.125\linewidth]{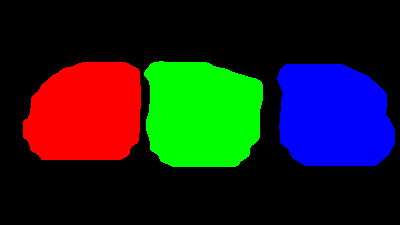}}
        &
		\makecell[c]{\includegraphics[width=0.225\linewidth,height=0.125\linewidth]{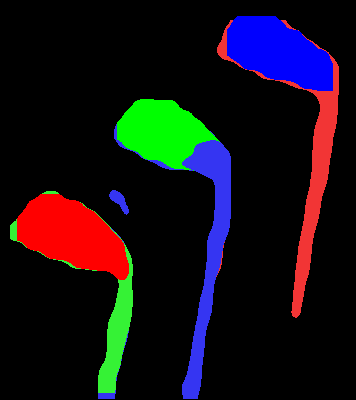}}
		&
		\makecell[c]{\includegraphics[width=0.225\linewidth,height=0.125\linewidth]{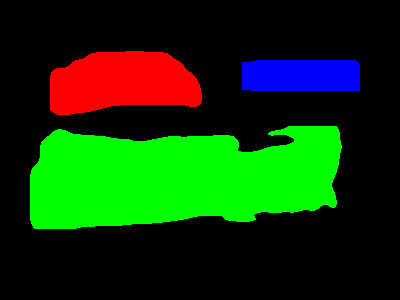}}
		&
		\makecell[c]{\includegraphics[width=0.225\linewidth,height=0.125\linewidth]{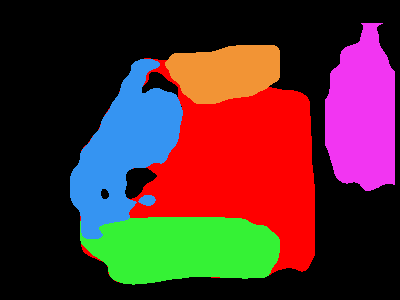}}
            \vspace{-0.5mm}
		\\

		\end{tabular}
            \vspace{-2mm}
    \caption{\textbf{Qualitative comparison with specialized SOTA methods on SIS.} \textbf{(a)} denotes the input image, \textbf{(b)} the ground-truth instance masks, \textbf{(c)} our Saliency-R1, followed by: \textbf{(d)} OQTR, \textbf{(e)} RDPNet, and \textbf{(f)} S4Net.}
    \label{SOTA_SIS}

\end{minipage}
\end{figure}

In SIS, as shown in Figure~\ref{SOTA_SIS}, specialized methods exhibit diverse failure modes: under-prediction of small instances (\eg, one water cup is missed in Column 1), failure to disambiguate overlapping instances (\eg, adjacent golf clubs are merged by OQTR in Column 2), confusion or omission of multiple objects (\eg, the three golden bars are incorrectly segmented as a single mask by OQTR in Column 3), and over-segmentation of a single instance into multiple parts (\eg, the chair is split into separate masks by OQTR in Column 4). Our method, guided by explicit CoT reasoning that generates distinct \texttt{<ins>}-tagged descriptions (\eg, ``a golf iron on the left, slightly tilted''), enables EVF-SAM to robustly assign precise, instance-level masks.

In CoSOD, as shown in Figure~\ref{SOTA_CoSOD}, prior methods often segment irrelevant co-occurring objects (\eg, the person beside the backpack in Column 1, or background items around the camera in Column 2). In contrast, our CoT’s explicit consensus deduction enables EVF-SAM to suppress distractors and produce cleaner, more selective masks, evident in the precise segmentation of cameras and yellow ducks.

Importantly, all improvements stem from the same unified pipeline, no per-task architecture changes, demonstrating the transferability of CoT-guided reasoning.

\begin{bibunit}
{
    \small
    \bibliographystyle{ieeenat_fullname}
    \bibliography{main}
}
\end{bibunit}



    




\end{document}